\documentclass[lettersize,journal,twoside]{IEEEtran}
\usepackage{amsmath,amsfonts,amssymb}
\usepackage{algorithmic}
\usepackage{algorithm}
\usepackage{array}
\usepackage[caption=false,font=normalsize,labelfont=sf,textfont=sf]{subfig}
\usepackage{textcomp}
\usepackage{stfloats}
\usepackage{url}
\usepackage{verbatim}
\usepackage{graphicx}
\usepackage{cite}
\usepackage[export]{adjustbox}
\usepackage[table,xcdraw]{xcolor}
\usepackage{color}
\usepackage[hidelinks]{hyperref}
\definecolor{linkcolor}{HTML}{ec008c}
\hypersetup{
	colorlinks=true,
	linkcolor=red,
	citecolor=red,
}
\usepackage{multirow}
\usepackage{booktabs}
\usepackage{orcidlink}
\usepackage{threeparttable}
\usepackage{pifont}

\newcommand\liehat[1]{\left[ #1 \right]_\times}

\newcommand\mlcomment[1]{\iffalse #1 \fi}
\newcommand\bsm[1]{\boldsymbol{\mathrm{#1}}}
\newcommand\transform[2]{{\bsm{T}_{#1}^{#2}}}

\newcommand\rotation[2]{{\bsm{R}_{#1}^{#2}}}
\newcommand\rotationhat[2]{{\hat{\bsm{R}}_{#1}^{#2}}}

\newcommand\timeoffset[2]{{\tau_{#1}^{#2}}}
\newcommand\timeoffsethat[2]{{\hat{\tau}_{#1}^{#2}}}

\newcommand\angvel[2]{{\bsm{\omega}_{#1}^{#2}}}

\newcommand\angacce[2]{{\bsm{\alpha}_{#1}^{#2}}}

\newcommand\translation[2]{{\bsm{p}_{#1}^{#2}}}
\newcommand\translationhat[2]{{\hat{\bsm{p}}_{#1}^{#2}}}
\newcommand\translationtilde[2]{{\tilde{\bsm{p}}_{#1}^{#2}}}
\newcommand\linvel[2]{{\bsm{v}_{#1}^{#2}}}
\newcommand\linvelhat[2]{{\hat{\bsm{v}}_{#1}^{#2}}}

\newcommand\linacce[2]{{\bsm{a}_{#1}^{#2}}}

\newcommand\gravity[1]{{\bsm{g}^{#1}}}
\newcommand\gravityhat[1]{{\hat{\bsm{g}}^{#1}}}

\newcommand\smallminus{{\text{-}}}
\newcommand\smallplus{{\text{+}}}
\newcommand\coordframe[1]{\underrightarrow{\mathcal{F}}_{#1}}
\newcommand{\tabheight}{0.6}
\newcommand{\tabwidth}{6pt}
\usepackage{siunitx}
\newcommand{\figscale}{0.9}
\begin{document}

\title{iKalibr: Unified Targetless Spatiotemporal Calibration for Resilient Integrated Inertial Systems}


\author{
Shuolong Chen \hspace{-1mm}$^{\orcidlink{0000-0002-5283-9057}}$, Xingxing Li \hspace{-1mm}$^{\orcidlink{0000-0002-6351-9702}}$, 
Shengyu Li \hspace{-1mm}$^{\orcidlink{0000-0003-4014-2524}}$, 
Yuxuan Zhou \hspace{-1mm}$^{\orcidlink{0000-0002-5261-0009}}$,
and Xiaoteng Yang \hspace{-1mm}$^{\orcidlink{0009-0008-7961-8749}}$

\thanks{
This work was supported by the National Key Research and Development Program of China under Grant 2023YFB3907100, the National Natural Science Foundation of China under Grant 423B240.
The authors are with the School of Geodesy and Geomatics (SGG), Wuhan University, Wuhan 430070, China.
Corresponding authors: Xingxing Li (\emph{xxli@sgg.whu.edu.cn}) and Shengyu Li (\emph{lishengyu@whu.edu.cn}).
}
\thanks{
The specific contributions of the authors to this work are listed in Section \emph{CRediT Authorship Contribution Statement} at the end of the article.
}
}
\markboth{Journal of \LaTeX\ Class Files,~Vol.~14, No.~8, August~2021}%
{Chen \MakeLowercase{\textit{et al.}}: iKalibr: Unified Targetless Spatiotemporal Calibration for Resilient Integrated Inertial Systems}


\maketitle

\begin{abstract}
The integrated inertial system, typically integrating an IMU and an exteroceptive sensor such as radar, LiDAR, and camera, has been widely accepted and applied in modern robotic applications for ego-motion estimation, motion control, or autonomous exploration.
To improve system accuracy, robustness, and further usability, both multiple and various sensors are generally resiliently integrated, which benefits the system performance regarding failure tolerance, perception capability, and environment compatibility.
For such systems, accurate and consistent spatiotemporal calibration is required to maintain a unique spatiotemporal framework for multi-sensor fusion.
Considering that most existing calibration methods ($i$) are generally oriented to specific integrated inertial systems, ($ii$) often focus on spatial-only determination, ($iii$) usually require artificial targets, lacking convenience and usability, we propose \emph{iKalibr}: a unified targetless spatiotemporal calibration framework for resilient integrated inertial systems, which overcomes the above issues, and enables both accurate and consistent calibration.
Altogether four commonly employed sensors are supported in \emph{iKalibr} currently, namely IMU, radar, LiDAR, and camera.
The proposed method starts with a rigorous and efficient dynamic initialization, where all parameters in the estimator would be accurately recovered.
Subsequently, several continuous-time batch optimizations are conducted to refine the initialized parameters toward better states.
Sufficient real-world experiments were conducted to verify the feasibility and evaluate the calibration performance of \emph{iKalibr}.
The results demonstrate that \emph{iKalibr} can achieve accurate resilient spatiotemporal calibration.
We open-source our implementations at (\url{https://github.com/Unsigned-Long/iKalibr}) to benefit the research community.
\end{abstract}


\section{Introduction}
\IEEEPARstart{I}{ntegrated} inertial systems, such as visual-inertial \cite{qin2018vins,geneva2020openvins,campos2021orb}, light detection and ranging (LiDAR)-inertial \cite{shan2020lio,xu2022fast,nguyen2023slict}, and radar-inertial \cite{kramer2020radar,ng2021continuous,michalczyk2022tightly} systems, have been widely applied in a variety of robotic tasks, especially in ego-motion estimation and autonomous exploration.
To further improve system performance in terms of failure tolerance, perception capability, and environment compatibility, multiple sensors, regarding both type \cite{zuo2020lic,lin2022r,zheng2022fast} and quantity \cite{eckenhoff2021mimc,doer2021x,nguyen2021miliom}, are commonly integrated into the system, which has been affordable due to the significantly reduced price and power consumption of sensors in recent years.
For such systems, accurate spatiotemporal calibration is needed to determine unique spatial and temporal frameworks for multi-sensor fusion, as inaccurate spatiotemporal parameters would largely impair system performance.

Overall, spatiotemporal calibration methods can be categorized as target-based and target-free ones, according to whether artificial targets are employed in calibration.
In target-based methods, artificial targets, such as chessboards \cite{zhang2000flexible} and AprilTags \cite{olson2011apriltag} oriented to cameras, easy-to-identify geometries \cite{liu2019error} oriented to LiDARs, or corner reflectors \cite{oh2018comparative} oriented to radars, are introduced in calibration for efficient and accurate data association.
Although precise calibration can be generally achieved in target-based methods, the requirement of additional artificial instruments may limit their convenience and further usability.
Differently, target-free methods automatically extract natural targets, such as visual landmarks \cite{yang2016monocular} or plane features \cite{lv2022observability}, from the environment to assist calibration.
Requiring no artificial target ($i$) enables convenient on-site calibration and ($ii$) significantly enhances the calibration usability, especially for complex multi-sensor setups that rely on intricate artificial targets.
Notably, the calibration performance of existing target-free methods still has considerable potential for improvement compared to target-based ones.

Spatiotemporal calibration methods can also be categorized as online and offline ones. In online calibration, spatiotemporal parameters are estimated alongside other relevant states in real-time estimators, such as in simultaneous localization and mapping (SLAMs) \cite{qin2018online} or odometries \cite{li2014online,huai2022observability}, where prior knowledge about initials of spatiotemporal parameters are generally required for system startup.
Analysis \cite{yang2019degenerate} and awareness \cite{lv2022observability} of spatiotemporal observability are also required to prevent potential rank deficiency in estimation, as the stable motion expected by online mapping-based estimators for continuous tracking generally lacks sufficient excitation, which could result in weak spatiotemporal observability.
In this context, online calibration is quite challenging, even with rigorous observability-aware modules integrated into estimators (it would increase system complexity, potentially compromising system stability and spatiotemporal convergence).
To achieve potential usability improvements, offline calibration decouples spatiotemporal determination from online state estimation, in contrast to online calibration.
While sufficiently excited motion is also necessary for offline calibration, particularly for IMU-related calibration \cite{yang2019degenerate}, it could be achieved more conveniently without concerns about challenges in real-time state estimation, such as continuous feature tracking and map updates.

To achieve accurate multi-sensor fusion, both spatial and temporal reference frameworks are required.
Most early calibration works \cite{fu2021high,mishra2021target,domhof2021joint} only focus on spatial determination (extrinsic determination), assuming sensors are temporally hardware-synchronized.
However, this could not hold for most low-cost smart devices, where only rough software-level time synchronization is conducted, leading to time offsets between sensors.
In such cases, ignoring the temporal calibration would introduce inevitable errors in spatial determination.
Considering this, most state-of-the-art methods \cite{furgale2013unified,lv2020targetless,huai2022continuous} support both spatial and temporal determination in calibration, thus are capable of higher usability.

In terms of spatiotemporal estimation, continuous-time estimation, relying on continuous-time representation, has been deeply studied in recent years, and tends to be more popular than traditional discrete-time estimation.
The continuous-time representation temporally decouples state estimates with sensor measurements, rather than inevitably couples them in the discrete-time representation \cite{cioffi2022continuous}.
Benefiting from this, continuous-time representation allows efficient state querying at arbitrary time instants by maintaining sparse but descriptive parameters \cite{sommer2020efficient,johnson2024continuous}, which is more suitable for temporal calibration.
Altogether two continuous-time representation methods are commonly utilized in the state estimation, namely nonparametric Gaussian processes (GPs) \cite{barfoot2014batch,anderson2015full} and parametric B-splines \cite{furgale2013unified,lv2020targetless}.
Compared with GPs where additional motion priors are generally required to guarantee sparsity for computation acceleration \cite{barfoot2014batch}, B-splines inherently possess sparsity because of their local controllability \cite{haarbach2018survey}.
Meanwhile, as piecewise polynomial functions, B-splines have closed-form expression, which allows efficient and accurate computation for analytical time derivatives.

\begin{figure}[t]
\centering
\includegraphics[width=\figscale\linewidth]{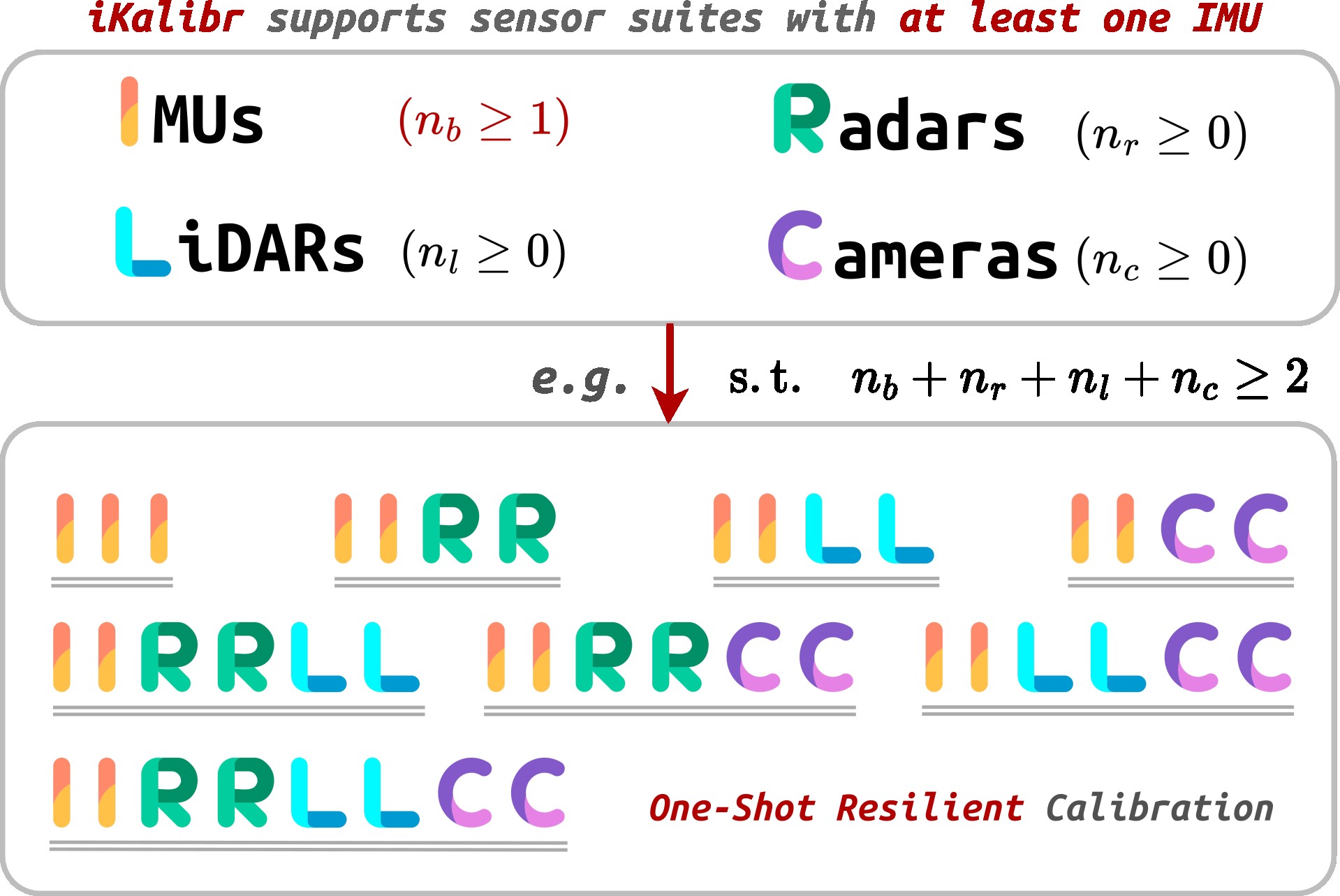}
\caption{Currently supported sensor suites in \emph{iKalibr} for one-shot calibration.}
\label{fig:ikalibr}
\end{figure}

Although numerous calibration methods oriented to specific minimum integrated inertial systems (integrating an IMU and an exteroceptive sensor) exist, varying from target-based to target-free, from online to offline, from spatial or temporal only to spatiotemporal, from discrete-time to continuous-time, multiple separate calibrations are generally required for more complex or resilient integrated inertial systems, which are labor-intensive and would lead to inconsistent spatiotemporal framework.
In this article, focusing on resilient integrated inertial systems and taking into account both calibration accuracy and usability, we present a unified spatiotemporal calibration framework, termed 
\emph{iKalibr}\footnote{
\textbf{The naming of \emph{iKalibr}} is with a little intention to follow the well-known target-based visual-inertial calibration toolkit \emph{Kalibr}\cite{furgale2013unified}. The lowercase $i$ at the beginning of \emph{iKalibr} highlights its focus on inertial systems, where at least one IMU is required. Note that while \emph{iKalibr} shares a similar name with \emph{Kalibr}, its calibration pipeline is fundamentally different from \emph{Kalibr}.
}, which is targetless, offline, and continuous-time-based.
\emph{iKalibr} is capable of 
one-shot resilient spatiotemporal determination (see footnotes for further explanation), 
and 
the key idea is to adaptively maintain continuous-time linear kinematic curves (such as acceleration, velocity, or translational ones) across different sensor configurations to ensure maximum usability.
This differs from traditional continuous-time-based methods, where a continuous-time trajectory is considered and maintained.
\emph{iKalibr} starts with a rigorous initialization to recover initials of all states in estimator, followed by several batch optimizations, where initialized states would be iteratively and steadily refined to better ones.
\emph{iKalibr} makes the following (potential) contributions:
\begin{enumerate}
\item As a continuous-time-based calibrator, \emph{iKalibr} is capable of accurate, consistent, and compact spatiotemporal determination, requiring \textbf{no} additional artificial targets or facilities, prior knowledge, and aiding sensors (thus is a compact calibrator).
Meanwhile, as a motion-based calibrator, \emph{iKalibr} does \textbf{not} require overlapping fields of view (FoVs) between sensors.

\item Due to the capability of \textbf{resilient calibration}\footnote{
\textbf{Resilient calibration} is inspired by the concept of resilient fusion \cite{fan2024autonomous}, which denotes a flexible data fusion approach that integrates data from diverse available sensors. Similarly, resilient calibration refers to the calibration of such resilient integrated multi-sensor systems without assumptions about the types or quantities of sensors involved. This differs significantly from traditional calibration methods, which are tailored to specific sensor configurations.
}, \emph{iKalibr} supports numerous integrated inertial systems, and is capable of \textbf{one-shot calibration}\footnote{
\textbf{One-shot calibration} refers to determining spatiotemporal parameters of all integrated sensors within a single, unified least-squares problem. This is distinguished from separate multi-step calibration, where parameters for each sensor pair are estimated separately using different calibration methods.
For example, calibrating an IMU-LiDAR-camera suite may involve combining methods like \emph{Kalibr} \cite{furgale2013unified} and \emph{LI-Calib} \cite{lv2020targetless}.
}, see Fig. \ref{fig:ikalibr}.
The IMUs, radars, LiDARs, and optical cameras are currently supported in \emph{iKalibr}. Other types of sensors, such as the popular event cameras, can be easily extended in \emph{iKalibr}.

\item A rigorous and efficient \textbf{dynamic initialization}\footnote{
\textbf{Dynamic initialization} refers to determining the gravity vector without relying on static assumptions about the system’s startup state in initialization procedure.
Static initialization, as opposed to dynamic initialization, is more commonly applied in INS or Aided INS systems. However, it requires the system to start in a stationary state, enabling direct determination of the gravity vector and even the global attitude.
}
procedure is designed in \emph{iKalibr} to recover initial guesses of \textbf{all parameters} in the state vector from unknown states, which potentially benefits initialization of continuous-time-based real-time estimators, such as visual-inertial, LiDAR-inertial, radar-inertial odometries, or more complex integrated inertial systems, since continuous-time curves and the gravity vector can be recovered alongside relevant spatiotemporal parameters during initialization.
Such an initialization would also benefit the online calibration methods by providing initials of spatiotemporal parameters for estimator startup.

\item 
Extensive real-world experiments were conducted to verify the feasibility and evaluate the calibration performance of \emph{iKalibr}.
Both the real-world datasets and code implementations have been open-sourced, to further support the robotics community if possible.
\end{enumerate}

The rest of this article is structured as follows:
Section \ref{sect:realted_works} reviews the related spatiotemporal calibration works.
Section \ref{sect:preliminaries} provides sensor models and continuous-time state representation.
Section \ref{sect:overview} gives an overview of \emph{iKalibr}.
Section \ref{sect:init} and \ref{sect:ct_ba} present the detailed calibration pipeline of initialization and batch optimization, respectively.
Section \ref{sect:experiments} reports the real-world experiments and results.
Finally, Section \ref{sect:conclusion} concludes our work and provides future directions.
Appendix \hyperref[sect:app_inertial_intri_calib]{A} and \hyperref[sect:app_alignment]{B} provide essential supplementary materials to facilitate a deep and comprehensive understanding of this work for the readers.

\begin{table*}[t]
\setlength{\tabcolsep}{\tabwidth}
\centering
\caption{\textbf{Chronological Summary of The Most Representative Spatiotemporal Calibration Methods}
\\The open-source iKalibr supports the widest range of sensor types and offers resilient calibration capability
}

\label{tab:calib_comp}
\begin{threeparttable}

\begin{tabular}{r|lcccccccc}
\toprule
\multicolumn{1}{c|}{\multirow{3}{*}{Method}}                    & \multicolumn{1}{c}{\multirow{3}{*}{Sensor}}                  & \multicolumn{2}{c}{Parameters}                              & \multirow{3}{*}{Target} & \multirow{3}{*}{Optimization} & \multirow{3}{*}{Estimator} & \multicolumn{2}{c}{Resilient Support}                       & \multirow{3}{*}{Open-Source} \\ \cmidrule{3-4} \cmidrule{8-9}
\multicolumn{1}{c|}{}                                           & \multicolumn{1}{c}{}                                         & Spatial                      & Temporal                     &                         &                               &                            & Quantity                     & Type                         &                              \\ \midrule\midrule
Mirzaei et al. \cite{mirzaei2008kalman}                         & \textbf{I}, \textbf{GS}                                      & \textcolor{green}{\ding{51}} & \textcolor{red}{\ding{55}}   & Based                   & Discrete                      & Offline                    & \textcolor{red}{\ding{55}}   & \textcolor{red}{\ding{55}}   & \textcolor{red}{\ding{55}}   \\
Kalibr \cite{furgale2013unified, huai2022continuous}            & \textbf{I}, \textbf{GS}, \textbf{RS}                         & \textcolor{green}{\ding{51}} & \textcolor{green}{\ding{51}} & Based                   & Continuous                    & Offline                    & \textcolor{green}{\ding{51}} & \textcolor{green}{\ding{51}} & \textcolor{green}{\ding{51}} \\
Li et al. \cite{li2014online}                                   & \textbf{I}, \textbf{GS}                                      & \textcolor{red}{\ding{55}}   & \textcolor{green}{\ding{51}} & Free                    & Discrete                      & Online                     & \textcolor{red}{\ding{55}}   & \textcolor{red}{\ding{55}}   & \textcolor{red}{\ding{55}}   \\
Natour et al. \cite{el2015radar}                                & \textbf{GS}, \textbf{R}                                      & \textcolor{green}{\ding{51}} & \textcolor{red}{\ding{55}}   & Based                   & Discrete                      & Offline                    & \textcolor{red}{\ding{55}}   & \textcolor{red}{\ding{55}}   & \textcolor{red}{\ding{55}}   \\
Nikolic et al. \cite{nikolic2016non}                            & \textbf{I}, \textbf{GS}                                      & \textcolor{green}{\ding{51}} & \textcolor{green}{\ding{51}} & Based                   & Continuous                    & Offline                    & \textcolor{red}{\ding{55}}   & \textcolor{red}{\ding{55}}   & \textcolor{red}{\ding{55}}   \\
Yang et al. \cite{yang2016monocular}                            & \textbf{I}, \textbf{GS}                                      & \textcolor{green}{\ding{51}} & \textcolor{red}{\ding{55}}   & Free                    & Discrete                      & Online                     & \textcolor{red}{\ding{55}}   & \textcolor{red}{\ding{55}}   & \textcolor{red}{\ding{55}}   \\
Per{\v{s}}i{\'c} et al. \cite{pervsic2017extrinsic}             & \textbf{L}, \textbf{R}                                       & \textcolor{green}{\ding{51}} & \textcolor{red}{\ding{55}}   & Based                   & Discrete                      & Offline                    & \textcolor{red}{\ding{55}}   & \textcolor{red}{\ding{55}}   & \textcolor{red}{\ding{55}}   \\
Le et al. \cite{le20183d}                                       & \textbf{I}, \textbf{L}                                       & \textcolor{green}{\ding{51}} & \textcolor{red}{\ding{55}}   & Free                    & Discrete                      & Offline                    & \textcolor{red}{\ding{55}}   & \textcolor{red}{\ding{55}}   & \textcolor{red}{\ding{55}}   \\
Liu et al. \cite{liu2019error}                                  & \textbf{I}, \textbf{L}                                       & \textcolor{green}{\ding{51}} & \textcolor{red}{\ding{55}}   & Based                   & Discrete                      & Offline                    & \textcolor{red}{\ding{55}}   & \textcolor{red}{\ding{55}}   & \textcolor{red}{\ding{55}}   \\
Mishra et al. \cite{mishra2021target}                           & \textbf{I}, \textbf{L}                                       & \textcolor{green}{\ding{51}} & \textcolor{red}{\ding{55}}   & Free                    & Discrete                      & Offline                    & \textcolor{red}{\ding{55}}   & \textcolor{red}{\ding{55}}   & \textcolor{green}{\ding{51}} \\
Domhof et al. \cite{domhof2021joint}                            & \textbf{GS}, \textbf{R}, \textbf{L}                          & \textcolor{green}{\ding{51}} & \textcolor{red}{\ding{55}}   & Based                   & Discrete                      & Offline                    & \textcolor{green}{\ding{51}} & \textcolor{green}{\ding{51}} & \textcolor{green}{\ding{51}} \\
Wise et al. \cite{wise2021continuous}                           & \textbf{GS}, \textbf{R}                                      & \textcolor{green}{\ding{51}} & \textcolor{red}{\ding{55}}   & Free                    & Continuous                    & Offline                    & \textcolor{red}{\ding{55}}   & \textcolor{red}{\ding{55}}   & \textcolor{red}{\ding{55}}   \\
Li et al. \cite{li20213d}                                       & \textbf{I}, \textbf{L}                                       & \textcolor{green}{\ding{51}} & \textcolor{red}{\ding{55}}   & Free                    & Continuous                    & Offline                    & \textcolor{red}{\ding{55}}   & \textcolor{red}{\ding{55}}   & \textcolor{green}{\ding{51}} \\
Calirad \cite{pervsic2021spatiotemporal}                        & \textbf{L}, \textbf{GS}, \textbf{R}                          & \textcolor{green}{\ding{51}} & \textcolor{green}{\ding{51}} & Based                   & Continuous                    & Offline                    & \textcolor{red}{\ding{55}}   & \textcolor{green}{\ding{51}} & \textcolor{green}{\ding{51}} \\
X-RIO \cite{doer2021x}                                          & \textbf{I}, \textbf{R}                                       & \textcolor{green}{\ding{51}} & \textcolor{red}{\ding{55}}   & Free                    & Discrete                      & Online                     & \textcolor{green}{\ding{51}} & \textcolor{red}{\ding{55}}   & \textcolor{green}{\ding{51}} \\
Mix-Cal \cite{lee2022extrinsic}                                 & \textbf{I}                                                   & \textcolor{green}{\ding{51}} & \textcolor{red}{\ding{55}}   & Free                    & Discrete                      & Offline                    & \textcolor{red}{\ding{55}}   & \textcolor{red}{\ding{55}}   & \textcolor{green}{\ding{51}} \\
Multical \cite{zhi2022multical}                                 & \textbf{I}, \textbf{L}, \textbf{GS}                          & \textcolor{green}{\ding{51}} & \textcolor{green}{\ding{51}} & Based                   & Continuous                    & Offline                    & \textcolor{green}{\ding{51}} & \textcolor{red}{\ding{55}}   & \textcolor{green}{\ding{51}} \\
Huai et al. \cite{huai2022observability}                        & \textbf{I}, \textbf{GS}, \textbf{RS}                         & \textcolor{green}{\ding{51}} & \textcolor{green}{\ding{51}} & Free                    & Discrete                      & Online                     & \textcolor{red}{\ding{55}}   & \textcolor{green}{\ding{51}} & \textcolor{red}{\ding{55}}   \\
OA-Calib \cite{lv2020targetless,lv2022observability} & \textbf{I}, \textbf{L}                                       & \textcolor{green}{\ding{51}} & \textcolor{green}{\ding{51}} & Free                    & Continuous                    & Offline                    & \textcolor{red}{\ding{55}}   & \textcolor{red}{\ding{55}}   & \textcolor{green}{\ding{51}} \\
Li et al. \cite{li2023two}                                      & \textbf{I}, \textbf{L}, \textbf{GS}                          & \textcolor{green}{\ding{51}} & \textcolor{green}{\ding{51}} & Free                    & Continuous                    & Offline                    & \textcolor{red}{\ding{55}}   & \textcolor{red}{\ding{55}}   & \textcolor{red}{\ding{55}}   \\
\textbf{iKalibr (ours)}                                                  & \textbf{I}, \textbf{L}, \textbf{GS}, \textbf{RS}, \textbf{R} & \textcolor{green}{\ding{51}} & \textcolor{green}{\ding{51}} & Free                    & Continuous                    & Offline                    & \textcolor{green}{\ding{51}} & \textcolor{green}{\ding{51}} & \textcolor{green}{\ding{51}} \\ \bottomrule
\end{tabular}

\begin{tablenotes} 
\item[*] In the \textbf{Sensor} column, \textbf{I}, \textbf{L}, \textbf{GS}, \textbf{RS}, and \textbf{R} are used to represent IMU, LiDAR, GS/RS Camera, and Radar, respectively.
\item[*] \textbf{Target}: whether artificial targets are required; \textbf{Optimization}: continuous-time-based or discrete-time-based;
\textbf{Estimator}: whether the calibration is online or offline.
\textbf{Resilient Support}: whether resilient calibration is supported in terms of sensor quantity or type.
\end{tablenotes}
\end{threeparttable}
\end{table*}

\section{Related Work}
\label{sect:realted_works}
There has been considerable research on the spatiotemporal calibration of the sensors addressed in this work.
Taking into account both readability and integrity, this section reviews the most relevant calibration works, organized according to sensor suite categories.

\subsection{Visual-Inertial Calibration}
It is no exaggeration to say that the visual-inertial suite is the most popular multi-sensor integration of the last decade.
To determine the extrinsics of a visual-inertial suite, Mirzaei et al. \cite{mirzaei2008kalman} presented an extended Kalman filter (EKF)-based and chessboard-aided spatial calibrator, where a detailed observability analysis is also conducted.
Similarly employing EKF but orienting to temporal calibration, Li et al. \cite{li2014online} proposed a visual-inertial system, supporting time offset determination.
Focusing on both spatial and temporal calibration, Furgale et al. \cite{furgale2013unified} presented the distinguished target-based \emph{Kalibr}, which first introduces B-spline-based continuous-time estimation to calibration.
\emph{Kalibr} is then extended by Huai et al. \cite{huai2022continuous} to support rolling shutter (RS) cameras.
Different from global shutter (GS) cameras exposing an entire image simultaneously, the RS cameras expose consecutive rows of an image with a constant line delay, which requires further calibration.
Utilizing a non-parametric trajectory representation rather than the B-spline-based parametric one, Nikolic et al. \cite{nikolic2016non} designed a spatiotemporal visual-inertial calibrator, where the inertial intrinsic determination is also considered.
In addition to offline calibration, online calibration is also deeply studied.
Yang et al. \cite{yang2016monocular} proposed a graph-optimization-based visual-inertial navigation system, which supports online extrinsic calibration.
Huai et al. \cite{huai2022observability} presented a visual-inertial odometry, which is capable of full calibration (extrinsic, temporal, and visual and inertial intrinsic).
Meanwhile, both GS cameras and RS cameras are supported in \cite{huai2022observability}.
Considering the weak observability of spatiotemporal parameters in online calibration, Yang et al. \cite{yang2019degenerate} performed an in-depth observability analysis for aided inertial navigation systems, where several degenerate motions are considered.
More directly, Xiao et al. \cite{xiao2022camera} designed a computationally efficient monitoring algorithm to explicitly determine when significant observability exists to support online visual-inertial recalibration.

\subsection{LiDAR-Inertial Calibration}
Compared to optical cameras, LiDARs are nearly unaffected by lighting conditions and can provide precise, dense point clouds with real-world geometric scale.
To calibrate extrinsics of LiDAR-inertial sensor suites, Liu et al. \cite{liu2019error} introduced artificial cone and cylinder features as calibration targets to aid geometric constraint construction.
Utilizing plane features extracted from structured environments, Le et al. \cite{le20183d} estimated LiDAR-inertial extrinsics based on a point-to-plane optimization framework.
Similar to \cite{le20183d} but considering more geometric features in environments, Liu et al. \cite{liu2019novel} proposed a multi-feature on-site extrinsic calibration method for LiDAR-inertial systems.
More weakly relying on structured environments, Mishra et al. \cite{mishra2021target} presented an EKF-based target-free extrinsic calibrator using the motion-based calibration constraint.
To determine both spatial and temporal parameters, Lv et al. \cite{lv2020targetless} proposed a target-free continuous-time-based spatiotemporal calibration method, i.e., the well-known \emph{LI-Calib}, which performs spatiotemporal optimization based on IMU-derived kinematic constraints and LiDAR-derived point-to-plane constraints.
In their further work, i.e., \emph{OA-Calib} \cite{lv2022observability}, an observability-aware module is designed for accurate observability-constrained spatiotemporal determination.
Similarly, as a continuous-time-based method but modeled by nonparametric GP, a targetless extrinsic calibration framework is proposed by Li et al. \cite{li20213d}. 
However, a priori dense point cloud map is required for plane feature extraction, thus potential inconvenience exists in application.
Differently, focusing on online calibration, Wu et al. \cite{wu2023afli} recently presented an adaptive frame length odometry with robust LiDAR-inertial extrinsic estimation, where IMU-derived pre-integration constraints and LiDAR-derived point-to-plane constraints are introduced in optimization.

\subsection{Radar-Related Calibration}
Different from LiDARs, radars provide sparse target clouds and additional radial velocities utilizing lower-frequency measuring signals, thus considered as weather-robust sensors.
Due to the shortage of radar-inertial calibration works, other radar-related works, such as radar-visual and radar-LiDAR, will also be discussed alongside radar-inertial ones in this section.
To calibrate a 2D-radar-LiDAR suite, Per{\v{s}}i{\'c} et al. \cite{pervsic2017extrinsic} designed a triangular calibration target with a trihedral reflector for correspondence registration, and presented a radar cross section (RCS)-based two-stage optimization method for extrinsic determination.
Focusing on 2D-radar-camera suites, Natour et al. \cite{el2015radar} proposed a radar-visual extrinsic calibration method, where painted canonical targets are employed for radar-visual data association.
To further jointly calibrate 2D-radar-camera-LiDAR suites, Domhof et al. \cite{domhof2021joint} designed a calibration target consisting of four round holes and a trihedral corner reflector, which supports efficient data association among multi-model sensor measurements.
Recently, 3D radars\footnote{
In some literature, the term \textbf{4D radars} is used to describe what we refer to here as \textbf{3D radars}, as they consider three-dimensional position combined with radial velocity as a four-dimensional observation.
} have become available and popular in robotic applications, which can provide additional elevation information of targets compared with 2D radars.
Wise et al. \cite{wise2021continuous} presented a continuous-time-based target-free 3D-radar-visual extrinsic calibrator, utilizing visual landmarks and radar-derived velocities.
More advancedly, Doer et al. \cite{doer2020radar} proposed an EKF-based 3D-radar-inertial odometry, supporting online extrinsic calibration.
However, initial guesses of extrinsics are required in this method.

\subsection{Inertial-Inertial (Multi-IMU) Calibration}
In practice, multi-IMU calibration could be solved through multiple separate sensor-IMU calibrations by introducing an additional aiding sensor, such as a camera, LiDAR, or radar.
However, this could introduce potential inconsistency, and is labor-intensive.
Considering the limitations, the authors of \emph{Kalibr} extended it to support single-camera-aided multi-IMU extrinsic calibration in \cite{rehder2016extending}, where the visual chessboards are required.
Similarly, Li et al. \cite{li2024targetless} extended the \emph{LI-Calib} to support multi-LiDAR multi-IMU spatiotemporal calibration method, which could be considered as a LiDAR-aided multi-IMU calibrator.
Differently, focusing on online multi-IMU calibration, Hartzer et al. \cite{hartzer2023online} proposed an EKF-based visual-inertial odometry, supporting extrinsic determination.
To ensure maximum usability, Lee et al. \cite{lee2022extrinsic} presented an IMU-only dual-IMU extrinsic calibration method, which requires no additional aiding sensors compared with the above ones.

\subsection{Calibration for Complex Integrated Inertial Systems}
While spatiotemporal parameters for more complex integrated inertial systems, like visual-LiDAR-inertial setups, can be determined through multiple separate calibration processes focused on individual subsystems, it's inconvenient and may introduce potential inconsistencies in the unified spatiotemporal framework to be determined.
Considering this, the authors of \emph{Kalibr} then extended it in \cite{rehder2016general} to support one-shot chessboard-based visual-inertial-laser spatiotemporal calibration.
To more effectively use artificial targets for constraint construction, Per\v{s}i\'c et al. \cite{pervsic2021spatiotemporal} introduced an isosceles-triangle integrated target, which could be simultaneously tracked by a camera, LiDAR, and radar for further spatiotemporal determination.
Similarly, Zhi et al. \cite{zhi2022multical} employed a chessboard-like integrated target compatible with multiple sensors for visual-LiDAR-inertial spatiotemporal calibration.
For such target-based methods orienting to complex integrated systems, although accurate calibration can be achieved, supporting additional new sensors may be challenging, as it requires redesigning the integrated target to ensure compatibility with the new sensors for one-shot calibration.
To address this issue, Liu et al. \cite{liu2022spatiotemporal} and Li et al. \cite{li2023two} successively proposed target-free spatiotemporal calibration methods for visual-LiDAR-inertial systems.
The unfortunate point is that, in both methods, the spatiotemporal parameters are determined in multiple separate least-squares problems, rather than through one-shot calibration.

\section{Preliminaries}
\label{sect:preliminaries}
This section presents mathematical notations and coordinate frame definitions used throughout this article, corresponding sensor models and B-spline-based continuous-time representation for time-varying states are also described in detail.

\subsection{Notations and Definitions}
Altogether four commonly employed sensor types, namely IMU, radar, LiDAR, and optical camera, are considered in this work for resilient spatiotemporal calibration.
Given an inertial system resiliently integrating a total of $n_b$ IMUs, $n_r$ radars, $n_l$ LiDARs, and $n_c$ cameras with
\begin{equation}
\label{equ:support_kit_type}
\small
\begin{gathered}
1\le n_b\in\mathbb{N},\;0\le n_r,n_l,n_c\in\mathbb{N},\\
n_b+n_r+n_l+n_c\ge 2,
\end{gathered}
\end{equation}
we use $\coordframe{b^i}$, $\coordframe{r^j}$, $\coordframe{l^k}$, and $\coordframe{c^m}$ to denote the coordinate frames of the $i$-th IMU, $j$-th radar, $k$-th LiDAR, and $m$-th camera, respectively.
We represent the six-degrees-of-freedom (DoF) rigid transform from $\coordframe{a}$ to $\coordframe{b}$ utilizing the Euclidean matrix $\transform{a}{b}$ in Lie group $\mathrm{SE}(3)$, which is defined as follows:
\begin{equation}
\small
\transform{a}{b}\triangleq\begin{bmatrix}
\rotation{a}{b}&\translation{a}{b}\\
\bsm{0}_{1\times 3}&1
\end{bmatrix}
\end{equation}
where $\rotation{a}{b}$  and $\translation{a}{b}$  denotes the rotation matrix in Lie group $\mathrm{SO}(3)$ and translation vector in vector space $\mathbb{R}^3$, respectively.
In terms of their high-order kinematics, we use $\angvel{a}{b}\in\mathbb{R}^3$ and $\angacce{a}{b}\in\mathbb{R}^3$ to express the angular velocity and acceleration of $\coordframe{a}$ with respect to and parameterized in $\coordframe{b}$.
Similarly, $\linvel{a}{b}\in\mathbb{R}^3$ and $\linacce{a}{b}\in\mathbb{R}^3$ denotes the corresponding linear velocity and acceleration, respectively.
$\timeoffset{a}{b}$ denotes time offset from $\coordframe{a}$ to $\coordframe{b}$, i.e., $\timeoffset{b}{}=\timeoffset{a}{}+\timeoffset{a}{b}$.
The gravity vector is expressed as $\bsm{g}\in\mathbb{R}^3$, which is considered as a two-DoF quantity with a constant Euclidean norm.
Finally, the noisy measurements and quantity estimates are denoted as $\tilde{(\cdot)}$ and $\hat{(\cdot)}$, respectively.

\subsection{Sensor Models}
\subsubsection{IMU Intrinsic Model}

\begin{figure}[t]
\centering
\includegraphics[width=\figscale\linewidth]{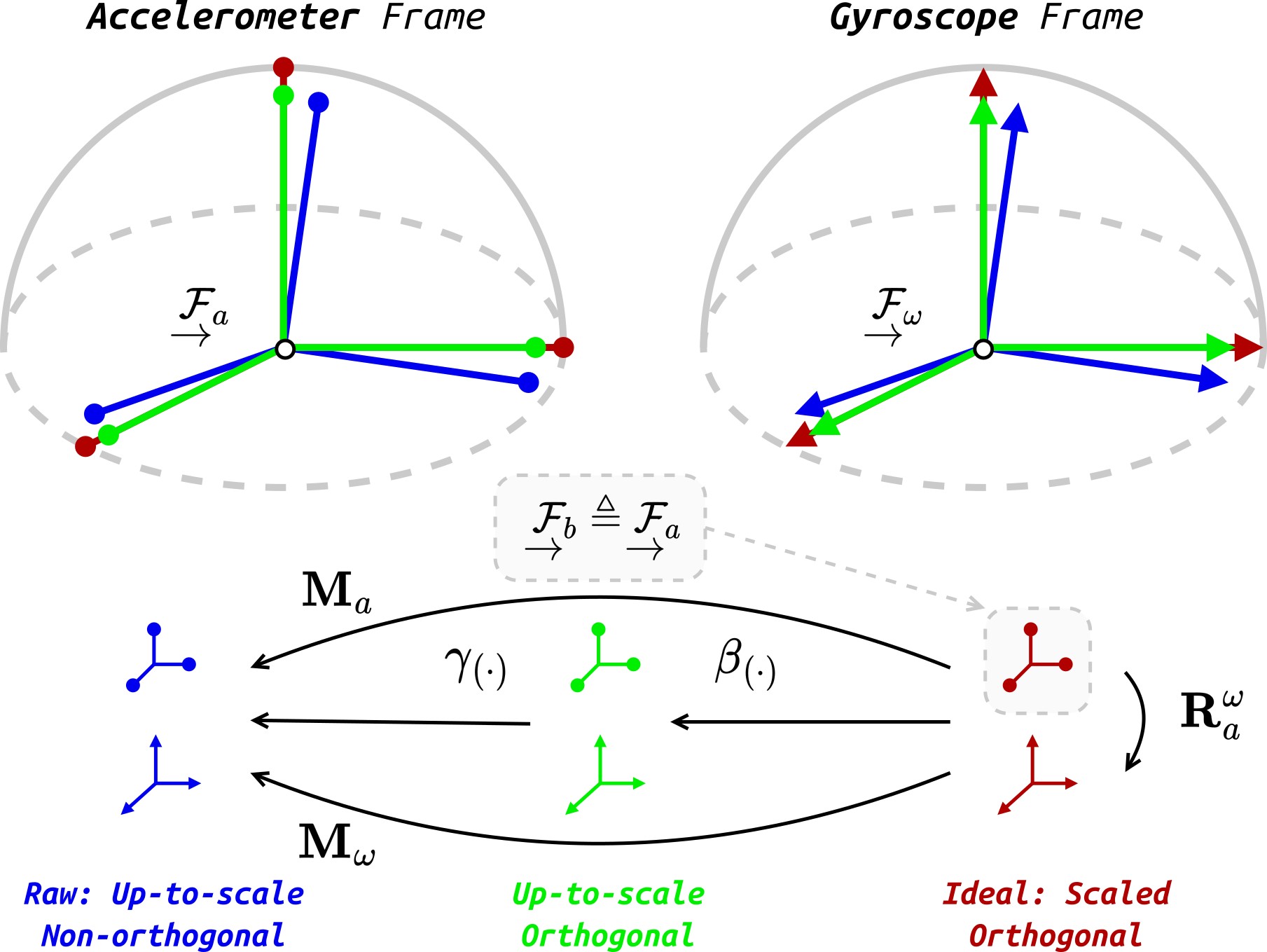}
\caption{Illustration of inertial measuring coordinates frames in an IMU. The rotational misalignment between accelerometer and gyroscope frames, non-orthogonality, and scale factors are considered here.}
\label{fig:imu_model}
\end{figure}

As a proprioceptive sensor, IMU perceives the kinematics of the body in inertial space, providing high-frequency body-frame angular velocity and specific force measurements.
Due to the potential inaccurate assembly and nominal scaling factors, and variable biases in output signals, IMU typically provides biased and noisy inertial measurements, especially the low-cost micro-electro-mechanical system (MEMS) IMUs.
Adhering to \cite{tedaldi2014robust}, the inertial measurements from the IMU are modeled as
\begin{equation}
\small
\begin{aligned}
\tilde{\bsm{a}}=
h_a\left(\bsm{a},\bsm{x}_{\mathrm{in}}^b \right)\;&\triangleq
\bsm{M}_a\cdot\bsm{a}+\bsm{b}_a+\bsm{\epsilon}_a
\\
\tilde{\bsm{\omega}}=
h_\omega\left( \bsm{\omega},\bsm{x}_{\mathrm{in}}^b \right) &\triangleq
\bsm{M}_\omega\cdot\rotation{a}{\omega}\cdot\bsm{\omega}+\bsm{b}_\omega+\bsm{\epsilon}_\omega
\end{aligned}
\end{equation}
where $\bsm{a}$ and $\bsm{\omega}$ are ideal specific force and angular velocity in scaled orthogonal frame $\coordframe{b}$, while $\tilde{\bsm{a}}$ and $\tilde{\bsm{\omega}}$ are noisy ones in update-to-scale non-orthogonal frames, see Fig. \ref{fig:imu_model}; $\coordframe{a}$ and $\coordframe{\omega}$ denote frames of the accelerometer and gyroscope respectively, which may not be coincident.
In this work, only the rotational misalignment in inconsistency is considered, which is parameterized as $\rotation{a}{\omega}$ by assuming that $\coordframe{b}$ coincides with $\coordframe{a}$, i.e., $\coordframe{b}\triangleq\coordframe{a}$.
$\bsm{M}_a$ and $\bsm{M}_\omega$ are upper triangular mapping matrices, which introduce the scale factors $\beta_{(\cdot)}$ and non-orthogonality factors $\gamma_{(\cdot)}$ to associate the ideal frames with up-to-scale non-orthogonal ones:
\begin{equation}
\small
\bsm{M}_a\triangleq\begin{bmatrix}
\beta_{a,1}&\gamma_{a,1}&\gamma_{a,2}\\
0&\beta_{a,2}&\gamma_{a,3}\\
0&0&\beta_{a,3}
\end{bmatrix}
,\;
\bsm{M}_\omega\triangleq\begin{bmatrix}
\beta_{\omega,1}&\gamma_{\omega,1}&\gamma_{\omega,2}\\
0&\beta_{\omega,2}&\gamma_{\omega,3}\\
0&0&\beta_{\omega,3}
\end{bmatrix}.
\end{equation}
$\bsm{b}_a$ and $\bsm{b}_\omega$ denote time-varying biases of the accelerometer and gyroscope respectively, and can be considered as constants in a short period, which holds in the proposed calibration framework.
The above intrinsic parameters, denoted as $\bsm{x}_{\mathrm{in}}^b$, can be pre-calibrated (see Appendix \hyperref[sect:app_inertial_intri_calib]{A}), or jointly optimized with other parameters in spatiotemporal calibration if strong observability\footnote{
\textbf{Inertial intrinsic observability}: when the platform undergoes fully excited six-DoF motion, the inertial intrinsics are observable \cite{yang2020online}.
However, in \emph{iKalibr}, only the IMU biases are estimated, while other intrinsic parameters are fixed as user inputs (acceptable to set them directly as identities or zeros, which are generally already compensated by the sensor manufacturer) in calibration. Such an approach is adopted because, under high measurement noise, directly estimating inertial intrinsics during spatiotemporal calibration could lead to these parameters absorbing other potential errors, resulting in estimation bias.
} exists.
$\bsm{\epsilon}_a$ and $\bsm{\epsilon}_\omega$ are corresponding zero-mean Gaussian white noises of sensors.

\subsubsection{Radar Doppler Velocity Model}
\begin{figure}[t]
\centering
\includegraphics[width=\figscale\linewidth]{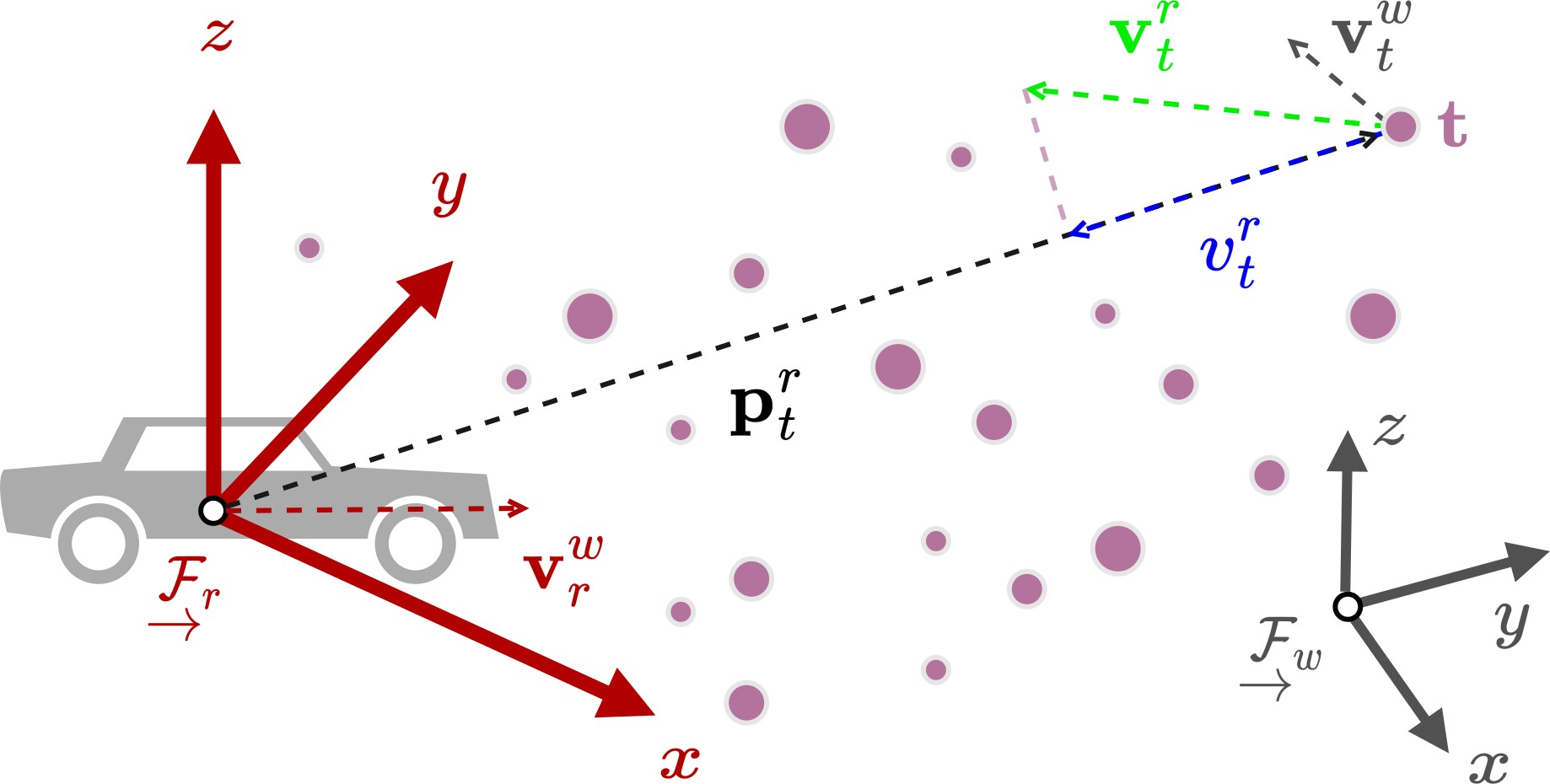}
\caption{Illustration of radar target measurements. Target position $\translation{t}{r^j}$ and its radial Doppler velocity $v^{r^j}_t$ can be obtained when tracked by the radar.}
\label{fig:radar_model}
\end{figure}
As an active exteroceptive sensor, radar measures the target position $\translation{t}{r}$ and its radial velocity $v_t^{r}$ in $\coordframe{r}$, i.e., the radial projection of the linear velocity of the target $\bsm{t}$ with respect to $\coordframe{r}$ (see Fig. \ref{fig:radar_model}), which can be expressed as
\begin{equation}
\small
\label{equ:radar_radial_vel}
\tilde{v}_t^{r}\triangleq\frac{\left( \translationtilde{t}{r}\right) ^\top\cdot\linvel{t}{r}}{\Vert\translationtilde{t}{r}\Vert}+\bsm{\epsilon}_r
=\frac
{\left( \translationtilde{t}{r}\right) ^\top\cdot\left( \rotation{r}{w}\right)^\top\cdot\left(  \linvel{t}{w}-\linvel{r}{w}\right)  }
{\Vert\translationtilde{t}{r}\Vert}+\bsm{\epsilon}_r
\end{equation}
with
\begin{equation}
\small
\begin{aligned}
\linvel{t}{r}=
\liehat{\translation{t}{r}}\cdot\left( \rotation{r}{w}\right)^\top \cdot\angvel{r}{w}
+\left( \rotation{r}{w}\right)^\top\cdot
\left(  \linvel{t}{w}-\linvel{r}{w}\right) 
\end{aligned}
\end{equation}
where $\coordframe{w}$ denotes the world frame; $\Vert\cdot\Vert$ and $\liehat{\cdot}$ represent the Euclidean norm and cross-product operation, respectively;
$\bsm{\epsilon}_r$ denotes the measuring noise.
When the target $\bsm{t}$ is stationary in $\coordframe{w}$, i.e., $\linvel{t}{w}\equiv\bsm{0}_{3\times 1}$, (\ref{equ:radar_radial_vel}) can be rewritten as
\begin{equation}
\small
\label{equ:radar_radial_static}
\tilde{v}_t^{r}\triangleq-\frac
{\left( \translationtilde{t}{r}\right) ^\top\cdot\left( \rotation{r}{w}\right)^\top\cdot\linvel{r}{w}  }
{\Vert\translationtilde{t}{r}\Vert}+\bsm{\epsilon}_r,
\end{equation}
which is assumed in the proposed spatiotemporal calibration for better usability and is generally easy to satisfy.

\subsubsection{Camera Model}
Functioning as an exteroceptive yet passive sensor, the optical camera captures scenes within its FoV and projects them onto the image plane, delivering informative images.
Optical images are generally distorted due to the deviation from rectilinear projection led by the manufacturing and assembly of the lens.
While multiple visual projection and distortion models exist, the commonly employed pinhole \cite{holmes1859stereoscope} projection model, and Brown \cite{duane1971close} and fish-eye \cite{kannala2006generic} distortion models are considered in this work, which could be described as
\begin{equation}
\small
\tilde{\bsm{f}}_l^c=
\begin{bmatrix}
\tilde{u}\\\tilde{v}
\end{bmatrix}=\pi\left( \bsm{p}^{c}_l,\bsm{x}^{c}_{\mathrm{in}}\right)+\bsm{\epsilon}_c
\end{equation}
with
\begin{equation}
\small
\label{equ:pinhole_visual_proj}
\begin{aligned}
\pi\left( \bsm{p}^{c}_l,\bsm{x}^{c}_{\mathrm{in}}\right)&\triangleq
\begin{bmatrix}
f_x&0&c_x\\
0&f_y&c_y
\end{bmatrix}\cdot
\begin{bmatrix}
d_x(x_n,y_n)
\\
d_y(x_n,y_n)
\\
1
\end{bmatrix}
\end{aligned}
\end{equation}
and
\begin{equation}
x_n=\frac{\bsm{p}^{c}_{l,x}}{\bsm{p}^{c}_{l,z}}
,\;
y_n=\frac{\bsm{p}^{c}_{l,y}}{\bsm{p}^{c}_{l,z}}
\end{equation}
where $\bsm{p}^{c}_l$ denotes the landmark position in $\coordframe{c}$, imaged as $\tilde{\bsm{f}}_l^c$ on the image plane;
$\bsm{x}^{c}_{\mathrm{in}}$ is the camera intrinsics, which contains principal point $(c_x,c_y)$, focal length $(f_x,f_y)$, and distortion coefficients, and needs to be pre-calibrated using target-based intrinsic calibration method such as \cite{zhang2000flexible};
$\pi(\cdot)$ represents the pinhole projection function;
$d_x(\cdot)$ and $d_y(\cdot)$ are Brown or fish-eye distortion functions in the $x$-axis and $y$-axis directions, respectively; $\bsm{\epsilon}_c$ denotes the imaging noise.

\begin{figure}[t]
\centering
\includegraphics[width=\figscale\linewidth]{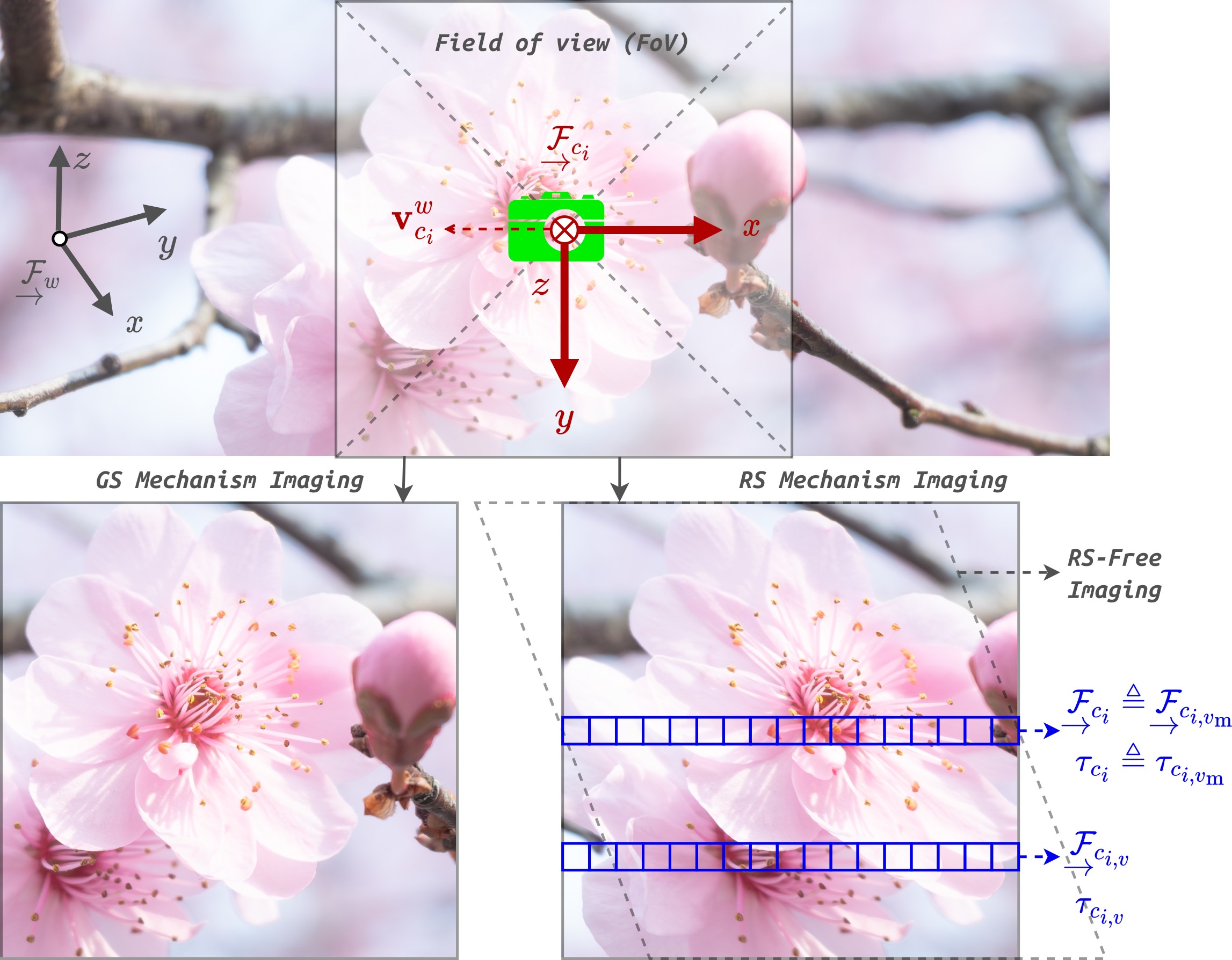}
\caption{Illustration of GS and RS mechanisms. All rows in an image are associated with a unique camera frame for GS mechanism, which does not hold for RS mechanism since rows are exposed sequentially in a readout timepiece. When relative motion exists between the RS camera and the scenario (movement to the left in this example), RS exposure mechanism leads to RS effect.}
\label{fig:gs_rs_model}
\end{figure}
In addition to imaging models, the visual geometric imaging process also varies depending on the exposure mode, which can be categorized as GS and RS mechanisms.
Consider that a landmark $\translation{l}{w}$ in $\coordframe{w}$ is imaged as feature $\tilde{\bsm{f}}_l^{c_i}$ in the $i$-th image frame, for the GS camera, such process could be described as
\begin{equation}
\small
\label{equ:gs_imaging}
\tilde{\bsm{f}}_l^{c_i}=\pi\left(
\translation{l}{c_i},\bsm{x}^{c}_{in}
\right)+\bsm{\epsilon}_{c_i}
\end{equation}
with
\begin{equation}
\small
\translation{l}{c_i}=\rotation{w}{c_i}\cdot\translation{l}{w}+\translation{w}{c_i}.
\end{equation}
However, for the RS camera, which exposes consecutive rows of an image with a constant line delay $\tau^c_\mathrm{ld}$ rather than exposing the entire image simultaneously like the GS camera, (\ref{equ:gs_imaging}) should be rewritten as
\begin{equation}
\small
\label{equ:rs_imaging}
\tilde{\bsm{f}}_l^{c_{i}}=\pi\left(
\translation{l}{c_{i,v}},\bsm{x}^{c}_{in}
\right)+\bsm{\epsilon}_{i,v}
\end{equation}
with
\begin{equation}
\small
\translation{l}{c_{i,v}}=\rotation{c_i}{c_{i,v}}\cdot\translation{l}{c_i}+\translation{c_i}{c_{i,v}}
,\quad
\translation{l}{c_{i}}=\rotation{w}{c_i}\cdot\translation{l}{w}+\translation{w}{c_i}
\end{equation}
where $\coordframe{c_{i,v}}$ denotes the camera frame when exposing the $v$-th row of the $i$-th image, which is generally not coincident with $\coordframe{c_i}$ in dynamic scenarios due to the RS exposure mechanism, see Fig. \ref{fig:gs_rs_model}.
By assuming that $\coordframe{c_i}$ coincides with the camera frame when exposing the \textbf{middle} row, i.e., $\coordframe{c_{i}}\triangleq\coordframe{c_{i,v_\mathrm{m}}}$, we have
\begin{equation}
\small
\tau_{c_{i,v}}=\tau_{c_i}+\left( \frac{v}{h}-\frac{1}{2}\right) \times \tau_\mathrm{red}
\end{equation}
where $\tau_{c_{i,v}}$ and $\tau_{c_i}\triangleq\tau_{c_{i,v_\mathrm{m}}}$ denote the exposure time associated with $\coordframe{c_{i,v}}$ and $\coordframe{c_i}$, respectively;
$h$ denotes the image height of the camera; $\tau_\mathrm{red}=h\times\tau_\mathrm{ld}$ is the readout time of a RS image frame, which would be calibrated simultaneously with other spatiotemporal parameters in the proposed method.
Note that (\ref{equ:rs_imaging}) is a more general geometric imaging process description, equally applicable for GS camera by considering $\tau_{\mathrm{red}}\equiv 0$, i.e., rows in a GS image frame are associated with a unique camera frame.
In addition, although this work focuses on the optical camera, it's suitable for more advanced cameras, such as the RGB-D camera and the bio-inspired event camera, since optical images are generally provided simultaneously by these cameras.

\subsubsection{LiDAR Model} 
\begin{figure}[t]
\centering
\includegraphics[width=\linewidth]{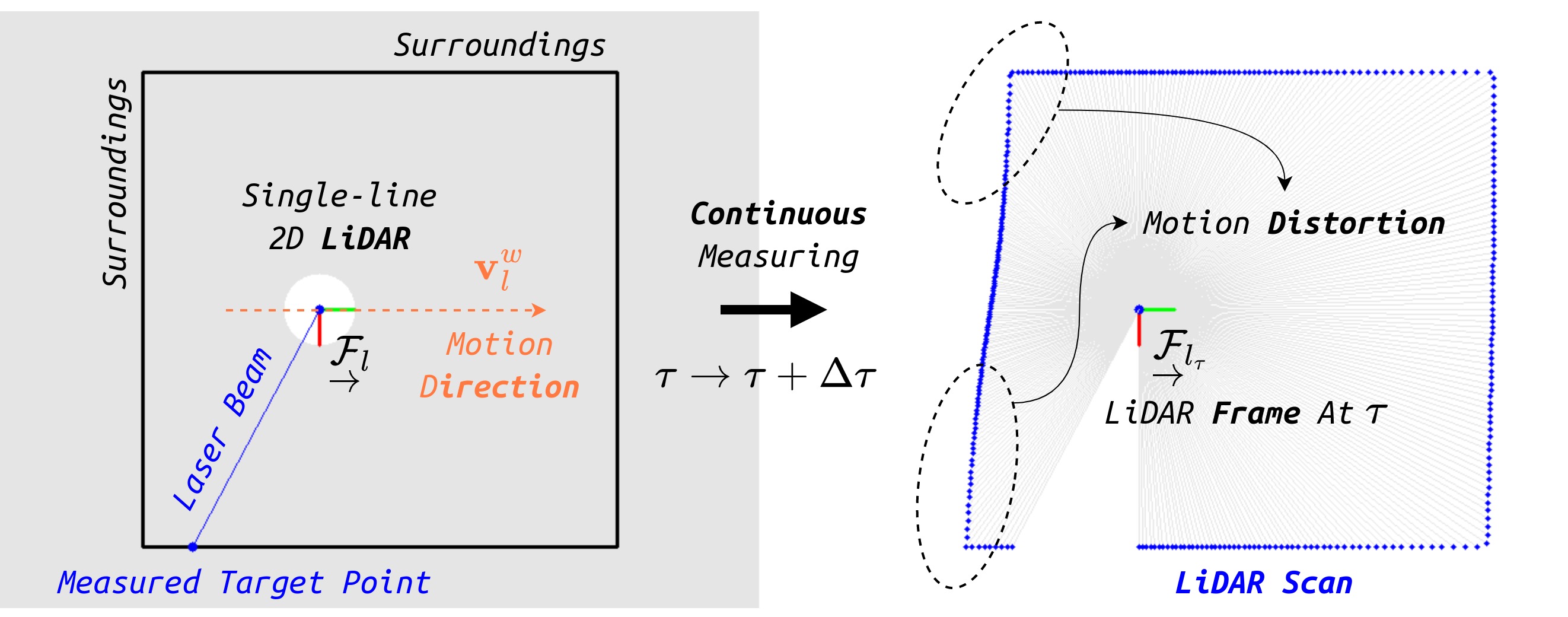}
\caption{Illustration of motion distortion in single-line 2D LiDAR scans. When relative motion exists between the LiDAR and its surroundings, the continuous measuring mode of the LiDAR results in distortions within the scans.}
\label{fig:lidar_motion_disto}
\end{figure}

As an exteroceptive sensor, LiDAR actively emits laser beams toward targets to measure the relative distances and further the relative positions of targets based on the orientations of laser beams.
LiDAR typically outputs data as scans of point clouds, each scan contains a substantial number of timestamped three-dimensional points measured within a time interval.
Due to LiDAR's continuous measuring mechanism\footnote{
\textbf{Continuous measuring mechanism} holds for mechanical rotating LiDARs and MEMS-based solid-state LiDARs this work focuses on, which is one of the primary causes of motion distortion in LiDAR scans.
For more advanced flash-based solid-state LiDARs, they illuminate the entire field of view in a single pulse, making them less prone to motion distortion \cite{zhou2023comparative}.
}, motion distortions similar to those in RS cameras could occur in scans when there is relative motion between the LiDAR and its surroundings, see Fig. \ref{fig:lidar_motion_disto}.

To achieve accurate LiDAR data fusion, distortions within LiDAR scans must be carefully considered.
In traditional discrete-time-based methods, such an issue is often addressed through motion model-based compensation \cite{zhang2014loam} or with the assistance of IMU data (e.g., through inertial pre-integration \cite{shan2020lio}).
For continuous-time-based methods, this process becomes significantly simpler. With states represented temporally continuously, the LiDAR pose for each point can be directly retrieved based on the point’s timestamp. Such an approach fundamentally resolves motion distortion by shifting from scan-based to point-based LiDAR data fusion.

\subsection{Continuous-Time State Representation}

\begin{figure}[t]
\centering
\includegraphics[width=0.7\linewidth]{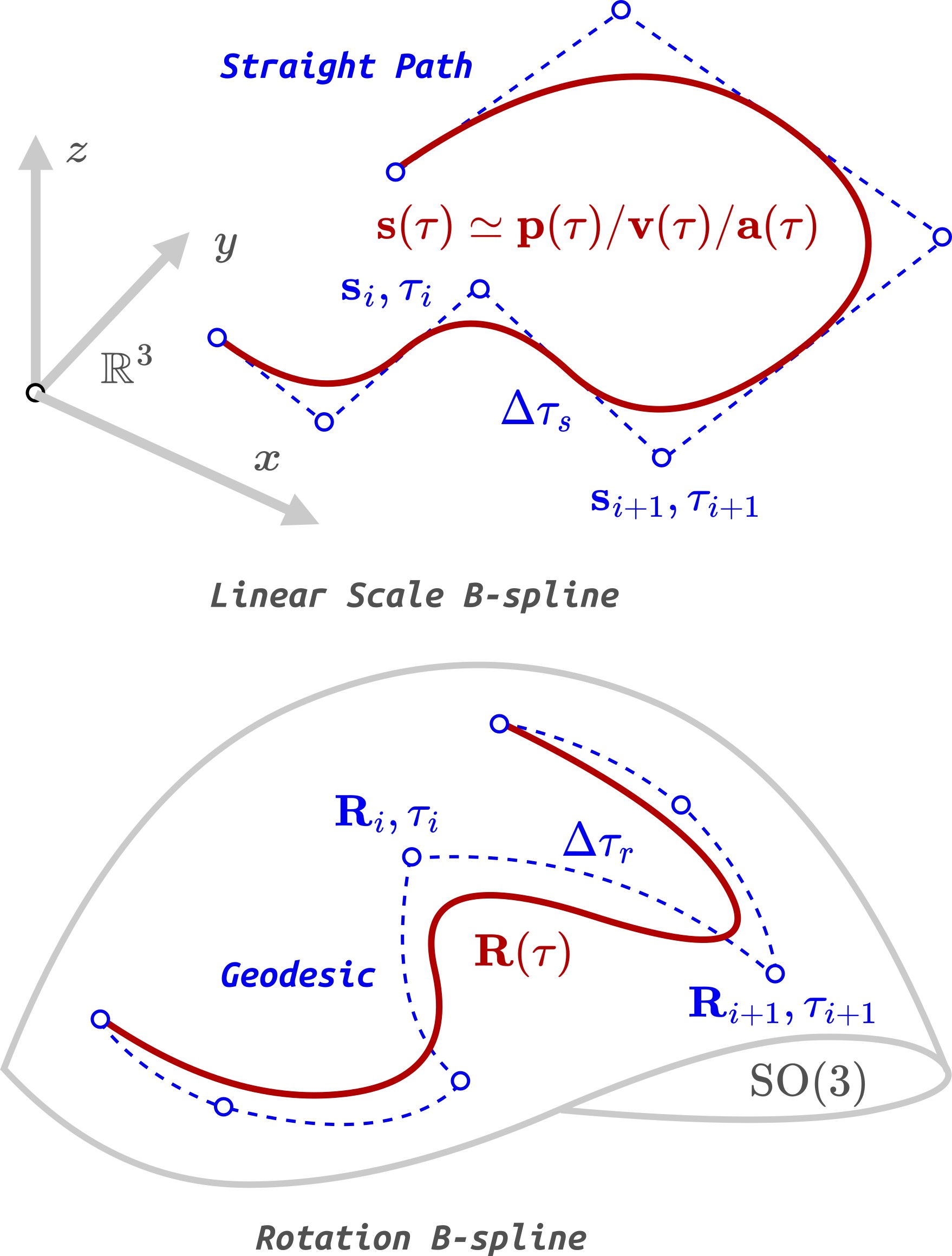}
\caption{Illustration of the linear scale B-spline living in $\mathbb{R}^3$ and the rotation B-spline living in $\mathrm{SO(3)}$, where red curves represent B-splines, blue circles are control points connected by dashed lines.
Different from the vector space where the distance between control points is measured by a straight line, the manifold of $\mathrm{SO(3)}$ measures distance by geodesic (also known as great arc).}
\label{fig:b-splines}
\end{figure}

\begin{figure*}[t]
\centering
\includegraphics[width=\linewidth]{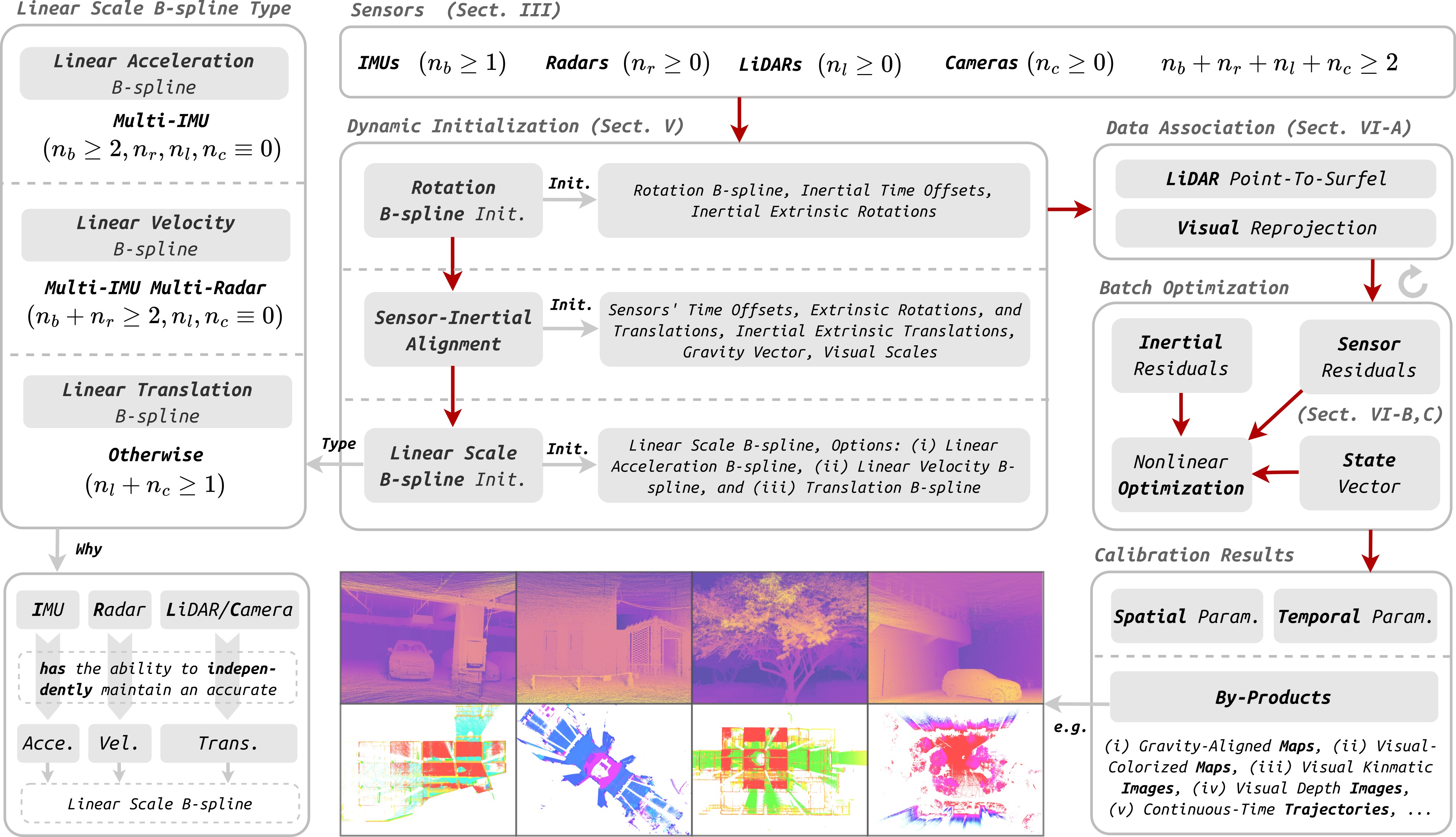}
\caption{Illustration of the full pipeline of the proposed targetless spatiotemporal calibration.
The system starts with a dynamic initialization procedure, where initial guesses of spatiotemporal parameters and B-splines are recovered (see Section \ref{sect:init}).
Subsequently, data association is performed to construct LiDAR point-to-surfel and visual reprojection correspondences if required (see Section \ref{sect:data_associate}).
Finally, a global nonlinear continuous-time-based batch optimization (see Section \ref{sect:batch_opt}) is carried out to iteratively refine (see Section \ref{sect:refine}) all initial states to better ones.
Though only four commonly employed sensor types, i.e., IMU, radar, LiDAR, and camera, are considered in this work currently, other types of sensors can be introduced conveniently.}
\label{fig:overview}
\end{figure*}

To efficiently fuse the high-frequency and asynchronous multi-sensor measurements in calibration, the B-spline-based continuous-time representation is employed in this work to encode time-varying states in the estimator.
Specifically, as an inertial-related spatiotemporal calibration framework, dynamic data acquisition is required, resulting in time-varying ($i$) rotational kinematics, associated with rotation-related spatiotemporal parameters such as extrinsic rotations, and ($ii$) translational kinematics, associated with linear-scale-related ones such as monocular visual scales and extrinsic translations.
In this work, the rotation B-spline living in $\mathrm{SO}(3)$ is employed for rotational kinematic representation.
In terms of the linear scale B-spline living in $\mathbb{R}^3$ for translational kinematic representation, it varies adaptively (it can be translation, linear velocity, or linear acceleration) based on the involved resilient integrated inertial sensor suites in calibration to ensure usability and flexibility, see Fig. \ref{fig:b-splines} and Fig. \ref{fig:overview}.

Specifically, given a series of linear scale control points (CPs, they could be translation, linear velocity, or acceleration ones) that are temporally uniformly distributed:
\begin{equation}
\small
\begin{gathered}
	\mathcal{S}\triangleq\left\lbrace
	 \bsm{s}_i,\tau_i\mid
	\bsm{s}_i\in\mathbb{R}^3,\tau_i\in\mathbb{R},i\in\mathbb{N}
	\right\rbrace 
	\\
	\mathrm{s.t.}\quad
	\tau_{i\smallplus 1}-\tau_{i}\equiv\Delta\tau_s,\;
	\bsm{s}_i\simeq\bsm{p}_i/\bsm{v}_i/\bsm{a}_i,
\end{gathered}
\end{equation}
the linear scale quantity $\bsm{s}(\tau)$ at time $\tau\in\left[ \tau_{i},\tau_{i\smallplus 1}\right) $ in a $k$-order uniform B-spline can be obtained by involving finite $k$ control points:
\begin{equation}
\small
\label{equ:b-spline-scale}
\bsm{s}(\tau)=\bsm{s}_i+\sum_{j=1}^{k\smallminus 1}
\lambda_j(u)
\cdot\left(\bsm{s}_{i\smallplus j}-\bsm{s}_{i\smallplus j\smallminus 1} \right) 
\end{equation}
where $u=\left( \tau-\tau_{i}\right)/\Delta\tau_{s}$, and $\lambda_j(u)$ is the $j$-th element of vector $\bsm{\lambda}(u)$, obtained from the order-determined cumulative matrix ${\bsm{N}}^{(k)}$ and normalized time vector $\bsm{u}$ \cite{sommer2020efficient}:
\begin{equation}
\small
\bsm{\lambda}(u)={\bsm{N}}^{(k)}\cdot \bsm{u}
\quad\mathrm{s.t.}\quad
\bsm{u}\triangleq\begin{bmatrix}
u^0&u^1&\cdots&u^{k\smallminus 1}
\end{bmatrix}^\top.
\end{equation}
In this work, the cubic ($k=4$) B-spline is employed, which stays $\mathcal{C}^2$ continuous and thus ensures continuity of zero-order, first-order, and second-order kinematics, such as continuous translation, velocity, and acceleration if a translational B-spline is utilized.
In such case, the cumulative matrix ${\bsm{N}}^{(4)}$ could be expressed as
\begin{equation}
\small
{\bsm{N}}^{(4)}\triangleq\frac{1}{6}
\begin{bmatrix}
6&0&0&0\\
5&3&-3&1\\
1&3&3&-2\\
0&0&0&1
\end{bmatrix}.
\end{equation}

As for the rotation B-spline, due to the non-closedness of scalar multiplication on the manifold of $\mathrm{SO}(3)$, the rotational control points should be first mapped to the tangent space of $\mathrm{SO}(3)$, i.e., the vector space of Lie algebra $\mathfrak{so}(3)$, for scaling, see Fig. \ref{fig:b-splines}.
Specifically, given a series of rotational control points that are temporally uniformly distributed:
\begin{equation}
\small
\begin{gathered}
	\mathcal{R}\triangleq\left\lbrace
	 \bsm{R}_i,\tau_i\mid
	\bsm{R}_i\in\mathrm{SO(3)},\tau_i\in\mathbb{R},i\in\mathbb{N}
	\right\rbrace 
	\\
	\mathrm{s.t.}\quad
	\tau_{i\smallplus 1}-\tau_{i}\equiv\Delta\tau_r
\end{gathered}
\end{equation}
the rotation $\bsm{R}(\tau)$ at time $\tau\in\left[ \tau_{i},\tau_{i\smallplus 1}\right) $ in a $k$-order uniform rotation B-spline can be obtained by involving finite $k$ control points:
\begin{equation}
\small
\label{equ:b-spline-rot}
\bsm{R}(\tau)=\bsm{R}_i\cdot\prod_{j=1}^{k\smallminus 1}
\mathrm{Exp}\left( \lambda_j(u)
\cdot\mathrm{Log}\left( \bsm{R}_{i\smallplus j\smallminus 1}^\top \cdot\bsm{R}_{i\smallplus j}\right) \right) 
\end{equation}
where the capitalized exponential mapping function $\mathrm{Exp}(\cdot)$ maps elements in the vector space of $\mathfrak{so}(3)$ to the manifold of $\mathrm{SO}(3)$, while the capitalized logarithmic mapping function $\mathrm{Log}(\cdot)$ is its inverse one.
Note that the time distances of linear scale and rotation B-splines, i.e., $\Delta\tau_s$ and $\Delta\tau_r$, should be carefully considered in practice\footnote{
\textbf{The time distance of a uniform B-spline}, as a critical hyperparameter, dictates the number of control points and overall representational fidelity of states. A smaller one yields a denser set of control points, enabling a more granular state representation but increasing the computational burden if control points are incorporated into the optimization. In \emph{iKalibr}, following empirical recommendations, a time distance in the range of $0.01\;sec$ to $0.1\;sec$ is typically appropriate for balanced accuracy and efficiency.
}.

\section{System Overview}
\label{sect:overview}
The structure of the proposed target-free spatiotemporal calibration method for resilient integrated inertial systems is shown in Fig. \ref{fig:overview}.
Given the raw measurements from integrated sensors that undergo sufficiently excited motion, the dynamic initialization procedure is first performed, which sequentially recovers initial guesses of the rotation B-spline, spatiotemporal parameters of involved sensors alongside the world-frame gravity vector, and the linear scale B-spline, see Section \ref{sect:init}.
The recovered linear scale B-spline type varies depending on the integrated sensor type, which can be linear acceleration B-spline, linear velocity B-spline, or translation B-spline, to guarantee both the usability and accuracy of the linear scale B-spline in calibration.
After the initialization, data association is carried out to construct LiDAR point-to-surfel and visual reprojection correspondences, if corresponding sensors are involved in calibration, see Section \ref{sect:data_associate}.
As for IMU and radar, no data association is required as the raw measurements can be tightly coupled in the continuous-time-based estimator.
Finally, a continuous-time-based nonlinear graph optimization is conducted to refine all initialized parameters to better states, see Section \ref{sect:batch_opt}.
Note that the final batch optimization is performed several times, using results from each iteration as initial estimates for the next, until convergence is achieved. For improved convergence performance, the spatiotemporal parameters are grouped and optimized sequentially rather than simultaneously, see Section \ref{sect:refine}.
To guarantee the consistency of the estimated spatiotemporal framework, an IMU is selected as the reference IMU (any one of
multiple IMUs, its frame is denoted as $\coordframe{b^r}$ in this article), whose kinematics coincide with the ones of maintained rotation and linear scale B-splines.
Note that the spatiotemporal parameters of the reference IMU are set as identities and keep constant in optimization to ensure a unique least-squares solution.

The total state vector in the estimator contains spatiotemporal parameters, inertial biases, gravity vector, and control points of B-splines, which are expressed as follows:
\begin{equation}
\small
\mathcal{X}\triangleq\left\lbrace
\bsm{x}_{\mathrm{rot}},\bsm{x}_{\mathrm{pos}},\bsm{x}_{\mathrm{tm}},\bsm{x}_{\mathrm{bias}},\gravity{w}, \mathcal{S}, \mathcal{R}
\right\rbrace 
\end{equation}
with
\begin{equation}
\small
\begin{gathered}
\bsm{x}_{\mathrm{rot}/\mathrm{pos}/\mathrm{tm}}\triangleq
\left\lbrace
\bsm{x}_{(\cdot)}^b,
\bsm{x}_{(\cdot)}^r,
\bsm{x}_{(\cdot)}^l,
\bsm{x}_{(\cdot)}^c
\right\rbrace
\quad\bsm{x}_{\mathrm{bias}}\triangleq
\left\lbrace
\bsm{b}_{a}^{b^i},\bsm{b}_{\omega}^{b^i}
\right\rbrace
\end{gathered}
\end{equation}
and
\begin{small}
\begin{alignat}{5}
\bsm{x}^{b}_{\mathrm{rot}}&\triangleq
\left\lbrace
\rotation{b^i}{b^r}
\right\rbrace
\quad
&&\bsm{x}^{b}_{\mathrm{pos}}&&\triangleq
\left\lbrace
\translation{b^i}{b^r}
\right\rbrace
\quad
&&\bsm{x}^{b}_{\mathrm{tm}}&&\triangleq
\left\lbrace
\timeoffset{b^i}{b^r}
\right\rbrace
\nonumber
\\
\bsm{x}^{r}_{\mathrm{rot}}&\triangleq
\left\lbrace
\rotation{r^j}{b^r}
\right\rbrace
\quad
&&\bsm{x}^{r}_{\mathrm{pos}}&&\triangleq
\left\lbrace
\translation{r^j}{b^r}
\right\rbrace
\quad
&&\bsm{x}^{r}_{\mathrm{tm}}&&\triangleq
\left\lbrace
\timeoffset{r^j}{b^r}
\right\rbrace
\nonumber
\\
\bsm{x}^{l}_{\mathrm{rot}}&\triangleq
\left\lbrace
\rotation{l^k}{b^r}
\right\rbrace
\quad
&&\bsm{x}^{l}_{\mathrm{pos}}&&\triangleq
\left\lbrace
\translation{l^k}{b^r}
\right\rbrace
\quad
&&\bsm{x}^{l}_{\mathrm{tm}}&&\triangleq
\left\lbrace
\timeoffset{l^k}{b^r}
\right\rbrace
\\
\bsm{x}^{c}_{\mathrm{rot}}&\triangleq
\left\lbrace
\rotation{c^m}{b^r}
\right\rbrace
\quad
&&\bsm{x}^{c}_{\mathrm{pos}}&&\triangleq
\left\lbrace
\translation{c^m}{b^r}
\right\rbrace
\quad
&&\bsm{x}^{c}_{\mathrm{tm}}&&\triangleq
\left\lbrace
\timeoffset{c^m}{b^r},\tau_{\mathrm{red}}^{c^m}
\right\rbrace
\nonumber
\end{alignat}
\end{small}
\begin{equation*}
\small
\begin{gathered}
\mathrm{s.t.}
\quad 
0\le i< n_b,
\quad
0\le j< n_r,
\quad
0\le k< n_l,
\\
\quad
0\le m< n_c,
\quad
i,j,k,m\in\mathbb{N}
\end{gathered}
\end{equation*}
where $\bsm{x}_{\mathrm{rot}}$, $\bsm{x}_{\mathrm{pos}}$, and $\bsm{x}_{\mathrm{tm}}$ denote extrinsic rotations, extrinsic translations, and temporal parameters respectively, all of them are parameterized with respect to the reference IMU $\coordframe{b^r}$;
Particularly, for RS camera, an additional temporal parameter, i.e., the readout time $\tau_{\mathrm{red}}^{c^m}$, is considered in calibration;
$\bsm{x}_{\mathrm{bias}}$ denotes the inertial biases of IMUs.
Note that the gravity vector and all control points in $\mathcal{S}$ and $\mathcal{R}$ are parameterized in $\coordframe{w}$, which is defined as the first frame of the reference IMU, i.e., $\coordframe{w}\triangleq\coordframe{b^r_0}$.

\section{Dynamic Initialization}
\label{sect:init}
As a highly nonlinear system, the continuous-time tightly-coupled state estimator requires reasonable initial guesses of involved parameters to pursue a global optimal solution and better convergence performance in the final batch optimization.
Considering this, a rigorous and efficient multi-stage initialization procedure is carried out, which sequentially recovers the rotation B-splines, the spatiotemporal parameters with the gravity vector, and the linear scale B-spline.
To avoid introducing unnecessary stationary prior and be consistent with the requirement of dynamic data acquisition, the initialization is performed dynamically (mainly on gravity determination), which significantly enhances both usability and flexibility.

\subsection{Rotation B-spline Recovery}
\label{sect:init_rot_spline_recovery}
Given the raw body-frame angular velocity measurements from multiple gyroscopes ($n_b\ge 1$), the rotation B-spline could be first recovered, alongside additional inertial extrinsic rotations and time offsets if $n_b> 1$, which is achieved by solving the following nonlinear least-squares problem:
\begin{equation}
\small
\label{equ:ls_rot_spline_recovery}
\left\lbrace  \hat{\mathcal{R}},\hat{\bsm{x}}_{\mathrm{rot}}^b,\hat{\bsm{x}}_{\mathrm{tm}}^b
\right\rbrace 
=\arg\min
\sum_{i}^{n_b}\sum_{n}^{\mathcal{D}^i_{\omega}}\left\| 
r_{\omega}\left( \tilde{\bsm{\omega}}^i_n\right) 
\right\| ^2_{\bsm{Q}_{\omega,n}^i}
\end{equation}
with
\begin{equation}
\small
\label{equ:gyro_residual}
\begin{aligned}
r_{\omega}\left(\tilde{\bsm{\omega}}^i_n\right)    &\triangleq
h_\omega\left( {\bsm{\omega}}^{i}(\tau_n+\timeoffsethat{b^i}{b^r}),\bsm{x}_{\mathrm{in}}^{b^i}\right) 
-\tilde{\bsm{\omega}}^i_n
\\
{\bsm{\omega}}^{i}(\tau)&=
\left( \rotation{b^r}{w}(\tau)
\cdot\rotationhat{b^i}{b^r} \right)^\top\cdot\angvel{b^r}{w}(\tau)	
\end{aligned}
\end{equation}
where $\mathcal{D}^i_{\omega}$ denotes the noisy angular velocity data sequence from the $i$-th gyroscope, in which $\tilde{\bsm{\omega}}^i_n$ is the $n$-th measurement at time $\tau_n$ stamped by $\coordframe{b^i}$;
$r_{\omega}\left( \cdot\right)$ is the gyroscope residual with information matrix ${\bsm{Q}_{\omega,n}^i}$ determined by measurement noise level; $\rotation{b^r}{w}(\tau)$ and $\angvel{b^r}{w}(\tau)$ are the ideal rotation and angular velocity at time $\tau$, analytically obtained from the rotation B-spline based on (\ref{equ:b-spline-rot}), which exactly involves the rotation control points into the optimization.

\begin{figure}[t]
\centering
\includegraphics[width=\figscale\linewidth]{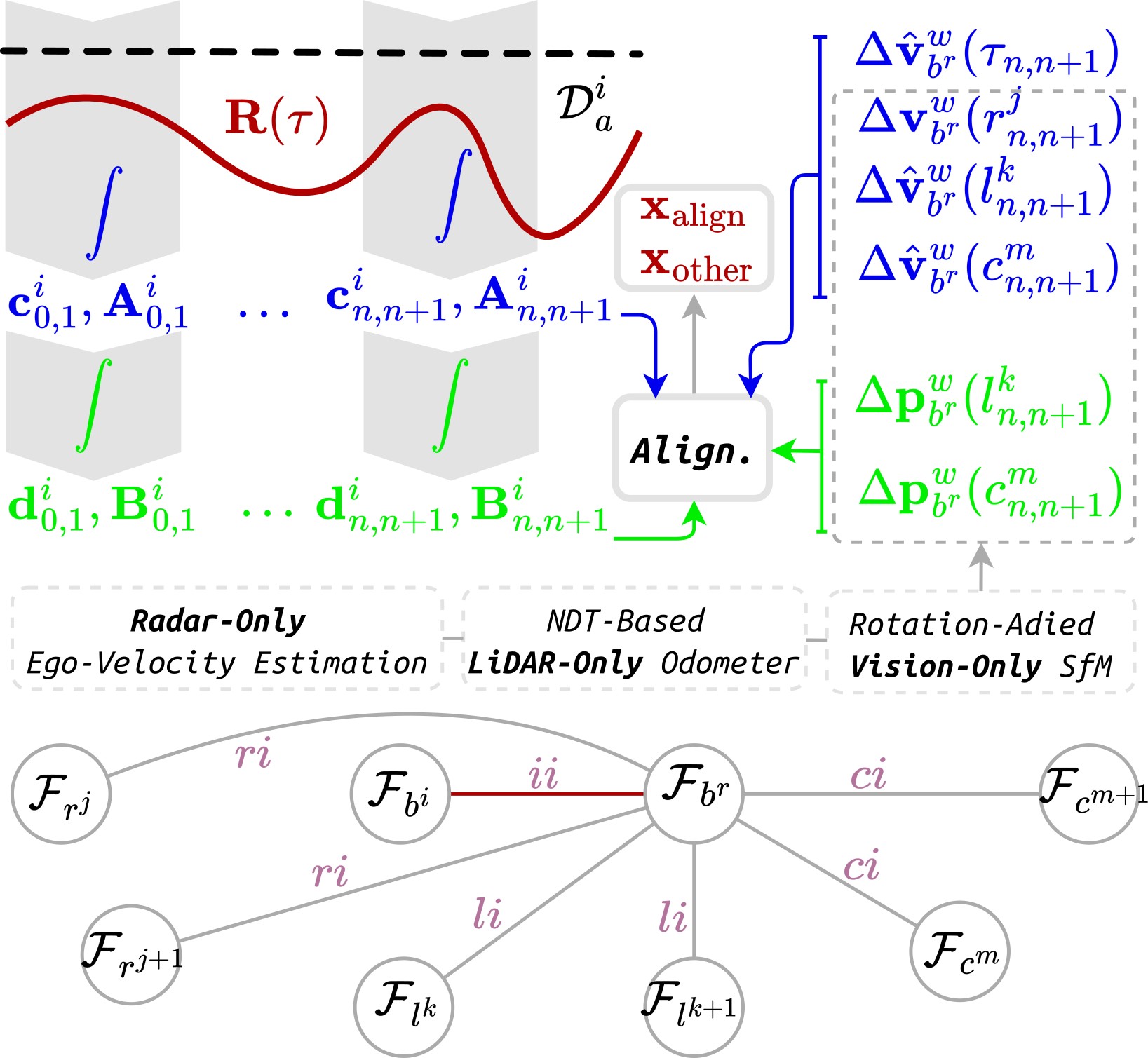}
\caption{Illustration of the sensor-inertial alignment. Sensor-inertial alignment constraints can be constructed between inertial sensors and other ones based on the initialized rotation B-spline $\bsm{R}(\tau)$, inertial specific force measurements $\mathcal{D}_a^i$, and sensor-derived kinematics. $ii$, $ri$, $li$, and $ci$ denote the constraints from inertial-inertial, radar-inertial, LiDAR-inertial, and visual-inertial alignments, respectively.} 
\label{fig:sensor-inertial-align}
\end{figure}

\subsection{Sensor-Inertial Alignment}
\label{sect:init_sen_inertial_align}
Based on the fitted rotation B-spline and recovered inertial extrinsic rotation and time offsets, the uninitialized spatiotemporal parameters and world-frame gravity vector can be determined by aligning the inertial measurements with sensor-derived kinematics, namely IMU-derived linear accelerations, radar-derived linear velocities, and LiDAR-derived and visual-derived poses, see Fig. \ref{fig:sensor-inertial-align}.
Note that the alignment constraints described in the following subsections would be introduced independently into a unique least-squared problem for one-shot alignment, see Section \ref{sect:one_shot_align}.
More mathematical details about alignment constraints can be found in Appendix
\hyperref[sect:app_alignment]{B}.

\subsubsection{Inertial-Inertial Alignment}
When multiple IMUs ($n_b>1$) are involved in calibration, inertial-inertial alignment constraints would be constructed between $\coordframe{b^i}$ and $\coordframe{b^r}$, which introduces the estimation of the inertial extrinsic translations and world-frame gravity vector.

Specifically, consider a timepiece $[\tau_n,\tau_{n\smallplus 1})$, the linear velocity variation of $\coordframe{b^r}$ can be expressed as:
\begin{equation}
\small
\label{equ:ii_align}
\begin{aligned}
\linvel{b^r}{w}(\tau_{n\smallplus 1})-\linvel{b^r}{w}(\tau_n)
=\bsm{c}^i_{n,n\smallplus 1}-\bsm{A}^i_{n,n\smallplus 1}
\cdot\translation{b^i}{b^r}+\gravity{w}\cdot\Delta\tau_{n,n\smallplus 1}
\end{aligned}
\end{equation}
with
\begin{equation}
\small
\begin{aligned}
\bsm{c}^i_{n,n\smallplus 1}&\triangleq
\int_{\tau_n}^{\tau_{n\smallplus 1}}\rotation{b^r}{w}(t)\cdot\rotation{b^i}{b^r} \cdot\bsm{a}^{i}(s) \cdot \mathrm{d}t,
\quad
s=t-\timeoffset{b^i}{b^r}
\\
\bsm{A}^i_{n,n\smallplus 1}&\triangleq
\int_{\tau_n}^{\tau_{n\smallplus 1}}\left(\liehat{\angacce{b^r}{w}(t)}+ \liehat{\angvel{b^r}{w}(t)}^2\right) \cdot\rotation{b^r}{w}(t)\cdot \mathrm{d}t
\end{aligned}
\end{equation}
where integration terms $\bsm{c}^i_{n,n\smallplus 1}$ and $\bsm{A}^i_{n,n\smallplus 1}$ can be obtained numerically based on raw specific force measurements and recovered states in (\ref{equ:ls_rot_spline_recovery}) ; $s$ and $t$ in integration are times stamped by $\coordframe{b^i}$ and $\coordframe{b^r}$, respectively.
By stacking multiple timepieces, the inertial-inertial alignment constraints would be introduced based on (\ref{equ:ii_align}), where world-frame linear velocities, i.e., $\linvel{b^r}{w}(\cdot)$, are treated as estimates alongside other interested quantities in the problem.

\subsubsection{Radar-Inertial Alignment}
When multiple radars ($n_r>0$) are involved in calibration, radar-inertial alignment constraints would be constructed between $\coordframe{r^j}$ and $\coordframe{b^r}$, which introduces the estimation of radar extrinsics (both rotations and translations) and world-frame gravity vector.
The radar-inertial alignment constraint in timepiece $[\tau_n,\tau_{n\smallplus 1})$
is similar to the one in (\ref{equ:ii_align}).
The difference is that the world-frame linear velocities of $\coordframe{b^r}$, i.e., $\linvel{b^r}{w}(\cdot)$, are derived by the radar, rather than treated as estimates in inertial-inertial alignment.

Specifically, consider the $k$-th scan $\mathcal{S}_r^j(\tau_k)$ from the $j$-th radar, which wraps multiple targets measurements, the radar-frame linear velocity of the $j$-th radar at time $\tau_k$, i.e., $\bsm{v}^{r^j}_{\tau_k}\triangleq\left( \rotation{r^j}{w}(\tau_k)\right)^\top \cdot\linvel{r^j}{w}(\tau_k)$, is estimated first based on (\ref{equ:radar_radial_static}) as follows:
\begin{equation}
\small
\label{equ:radar_velocity}
\hat{\bsm{v}}^{r^j}_{\tau_k}=\arg\min\sum^{\mathcal{S}_r^j(\tau_k)}_n
\left\| 
\left( \translationtilde{k,n}{r^j}\right) ^\top\cdot\hat{\bsm{v}}^{r^j}_{\tau_k}
+\tilde{v}^{r^j}_{k,n}\cdot\Vert\translationtilde{k,n}{r^j}\Vert
\right\| ^2_{\bsm{Q}_{r,k,n}^j}
\end{equation}
where $\left( \translationtilde{k,n}{r^j},\tilde{v}^{r^j}_{k,n}\right) \in\mathcal{S}_r^j(\tau_k)$ are position and radial velocity of the $n$-th target in $k$-th scan of the $j$-th radar.
The random sample consensus (RANSAC)-based outlier rejection is employed in problem (\ref{equ:radar_velocity}) to reject model-inconsistent dynamic targets.
Based on the estimated radar-frame linear velocity, the world-frame linear velocity of $\coordframe{b^r}$ can be expressed as:
\begin{equation}
\small
\label{equ:ri_align}
\linvel{b^r}{w}(r^j_\tau)\triangleq
\linvel{b^r}{w}(\tau)
=
\rotation{b^r}{w}(\tau)\cdot\rotation{r^j}{b^r}\cdot\bsm{v}^{r^j}_{\tau}-
\liehat{\angvel{b^r}{w}(\tau)}\cdot\rotation{b^r}{w}(\tau)\cdot\translation{r^j}{b^r}
\end{equation}
which involves the extrinsics of radars.
(\ref{equ:ii_align}) and (\ref{equ:ri_align}) would be organized together as radar-inertial alignment constraints.

\subsubsection{LiDAR-Inertial Alignment}
When LiDARs ($n_l>0$) are involved in calibration, LiDAR-inertial alignment constraints would be constructed between $\coordframe{l^k}$ and $\coordframe{b^r}$, which introduces the estimation of LiDAR extrinsics and gravity vector.

The normal distributions transform (NDT)-based LiDAR-only odometry \cite{biber2003normal} is first performed for each LiDAR independently to obtain rough poses.
Considering that the NDT algorithm relies on good initials when performing scan registration, we first ($i$) try to recover rough rotations using the non-prior NDT, then ($ii$) employ the rotation-only hand-eye alignment to recover the extrinsic rotation using the rough rotations of $\coordframe{l^k}$ and the rotation B-spline of $\coordframe{b^r}$, and finally ($iii$) perform a rotation-aided NDT to obtain accurate LiDAR poses.
Such a strategy is reasonable and effective, as sufficiently excited motion required by spatiotemporal determination causes rapid changes in the LiDAR FoV between consecutive scans, creating challenges for accurate registration when using non-prior NDT.
Specifically, given a rotation sequence $\mathcal{S}_{l}^k$ containing multiple NDT-derived rough rotations of the $\coordframe{l^k}$ from ($i$), the rotation-only hand-eye alignment described in ($ii$) can be performed by solving the following least-squares problem:
\begin{equation}
\small
\label{equ:lidar_rot_hand_eye}
\left\lbrace \hat{\bsm{R}}_{l^k}^{b^r},\timeoffsethat{l^k}{b^r}\right\rbrace =\arg\min\sum^{\mathcal{S}_{l}^k}_n
\left\| 
\rotationhat{l^k}{b^r}\cdot\rotation{l^k_{n\smallplus 1}}{l^k_{n}}
\left( \rotation{b^r_{n\smallplus 1}}{b^r_{n}}
\rotationhat{l^k}{b^r}\right) ^\top
\right\| ^2_{\bsm{Q}_{l,n}^k}
\end{equation}
with
\begin{equation}
\small
\begin{aligned}
\rotation{l^k_{n\smallplus 1}}{l^k_{n}}&\triangleq
\left( \rotation{l^k}{m^k}(\tau_{n})\right) ^\top
\cdot\rotation{l^k}{m^k}(\tau_{n\smallplus 1})
\\
\rotation{b^r_{n\smallplus 1}}{b^r_{n}}&\triangleq
\left( \rotation{b^r}{w}(\tau_{n}+\timeoffsethat{l^k}{b^r})\right) ^\top
\cdot\rotation{b^r}{w}(\tau_{n\smallplus 1}+\timeoffsethat{l^k}{b^r})
\end{aligned}
\end{equation}
where $\rotation{l^k}{m^k}(\tau_{n})\in\mathcal{S}_{l}^k$ is the NDT-derived rotation of $n$-th scan, which is expressed in the map frame $\coordframe{m^k}$; $\rotation{b^r}{w}(\cdot)$ is the rotation of the reference IMU obtained from the fitted rotation B-spline.
Note that both extrinsic rotation and time offset can be roughly determined for each LiDAR based on (\ref{equ:lidar_rot_hand_eye}).

To construct LiDAR-inertial alignment constraints, similarly to radar, the linear velocity of $\coordframe{b^r}$ in (\ref{equ:ii_align}) would be rewritten as:
\begin{equation}
\small
\label{equ:li_align_vel}
\linvel{b^r}{w}(l^k_\tau)\triangleq
\linvel{b^r}{w}(\tau)
=\linvel{l^k}{w}(\tau)-
\liehat{\angvel{b^r}{w}(\tau)}\cdot\rotation{b^r}{w}(\tau)\cdot\translation{l^k}{b^r}
\end{equation}
which is similar to (\ref{equ:ri_align}). The difference is that linear velocities of the LiDAR, i.e., $\linvel{l^k}{w}(\cdot)$, are treated as estimates here.
By continuing to perform time integration on (\ref{equ:ii_align}), NDT-derived positions can be involved in alignment constraints:
\begin{equation}
\small
\label{equ:li_align_pos}
\begin{aligned}
\translation{b^r}{w}(\tau_{n\smallplus 1})-\translation{b^r}{w}(\tau_n)
&=
\bsm{d}^i_{n,n\smallplus 1}-\bsm{B}^i_{n,n\smallplus 1}
\cdot\translation{b^i}{b^r}
\\
+&\linvel{b^r}{w}(\tau_n)\cdot\Delta\tau_{n,n\smallplus 1}+
\frac{1}{2}\cdot\gravity{w}\cdot\Delta^2\tau_{n,n\smallplus 1}
\end{aligned}
\end{equation}
with
\begin{equation}
\small
\begin{aligned}
\translation{b^r}{w}(l^k_\tau)&\triangleq
\translation{b^r}{w}(\tau)=
\rotation{m^k}{w}\cdot\translation{l^k}{m^k}(\tau)+\translation{m^k}{w}-\rotation{b^r}{w}(\tau)\cdot\translation{l^k}{b^r}
\\
\bsm{d}^i_{n,n\smallplus 1}&\triangleq
\iint_{\tau_n}^{\tau_{n\smallplus 1}}\rotation{b^r}{w}(t)\cdot\rotation{b^i}{b^r} \cdot\bsm{a}^{i}(s) \cdot \mathrm{d}t^2,
\quad
s=t-\timeoffset{b^i}{b^r}
\\
\bsm{B}^i_{n,n\smallplus 1}&\triangleq
\iint_{\tau_n}^{\tau_{n\smallplus 1}}\left(\liehat{\angacce{b^r}{w}(t)}+ \liehat{\angvel{b^r}{w}(t)}^2\right) \cdot\rotation{b^r}{w}(t)\cdot \mathrm{d}t^2
\end{aligned}
\end{equation}
where $\rotation{m^k}{w}=\rotation{b^r}{w}(\tau_m^k+\timeoffset{l^k}{b^r})\cdot\rotation{l^k}{b^r}$ denotes the rotation bias between $\coordframe{m^k}$ and $\coordframe{w}$, $\tau_m^k$ is the map time, which is usually set to the time of the first registered scan of NDT; $\translation{m^k}{w}$ is the position bias, which keeps unknown but would be eliminated in (\ref{equ:li_align_pos}) by subtraction; integration terms $\bsm{d}^i_{n,n\smallplus 1}$ and $\bsm{B}^i_{n,n\smallplus 1}$, similar to $\bsm{c}^i_{n,n\smallplus 1}$ and $\bsm{A}^i_{n,n\smallplus 1}$, can be obtained numerically.
(\ref{equ:ii_align}), (\ref{equ:li_align_vel}), and (\ref{equ:li_align_pos}) would be organized together as LiDAR-inertial alignment constraints.

\subsubsection{Visual-Inertial Alignment}
When cameras ($n_c>0$) are involved in calibration, visual-inertial alignment constraints would be constructed between $\coordframe{c^m}$ and $\coordframe{b^r}$, which introduces the estimation of visual extrinsics and gravity vector.

\begin{figure}[t]
\centering
\includegraphics[width=\figscale\linewidth]{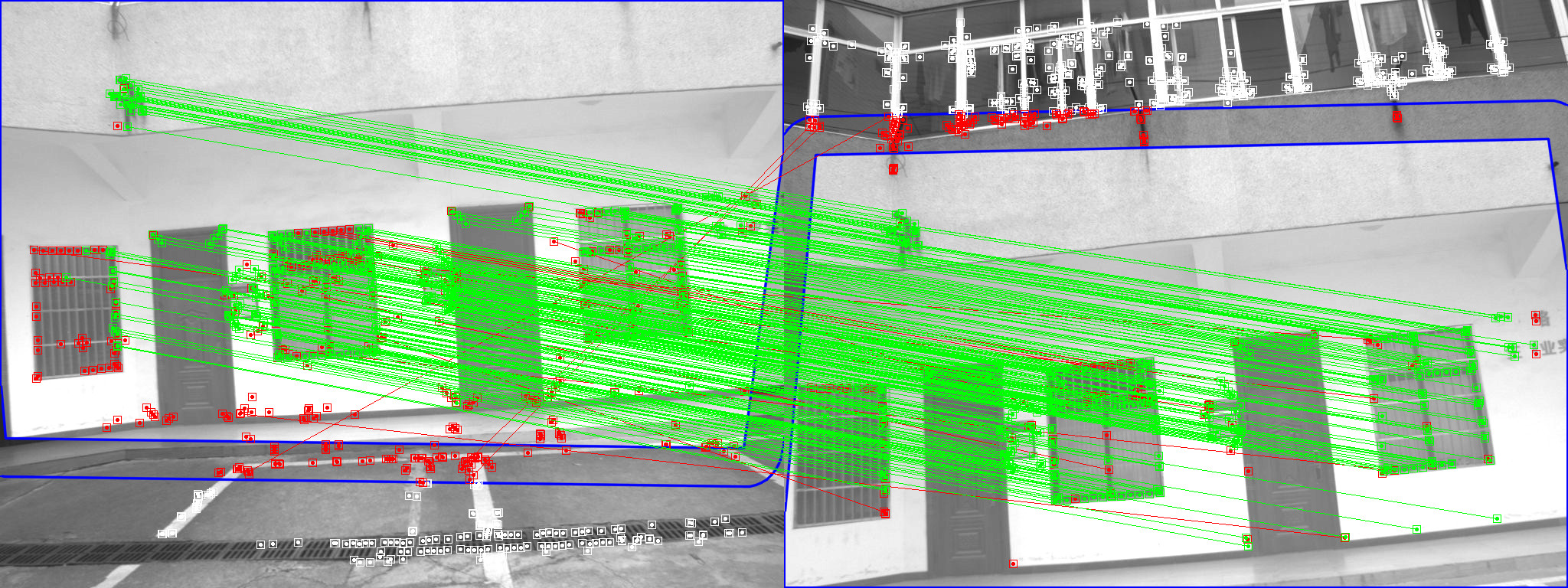}
		
\vspace{1mm}
		
\includegraphics[width=\figscale\linewidth]{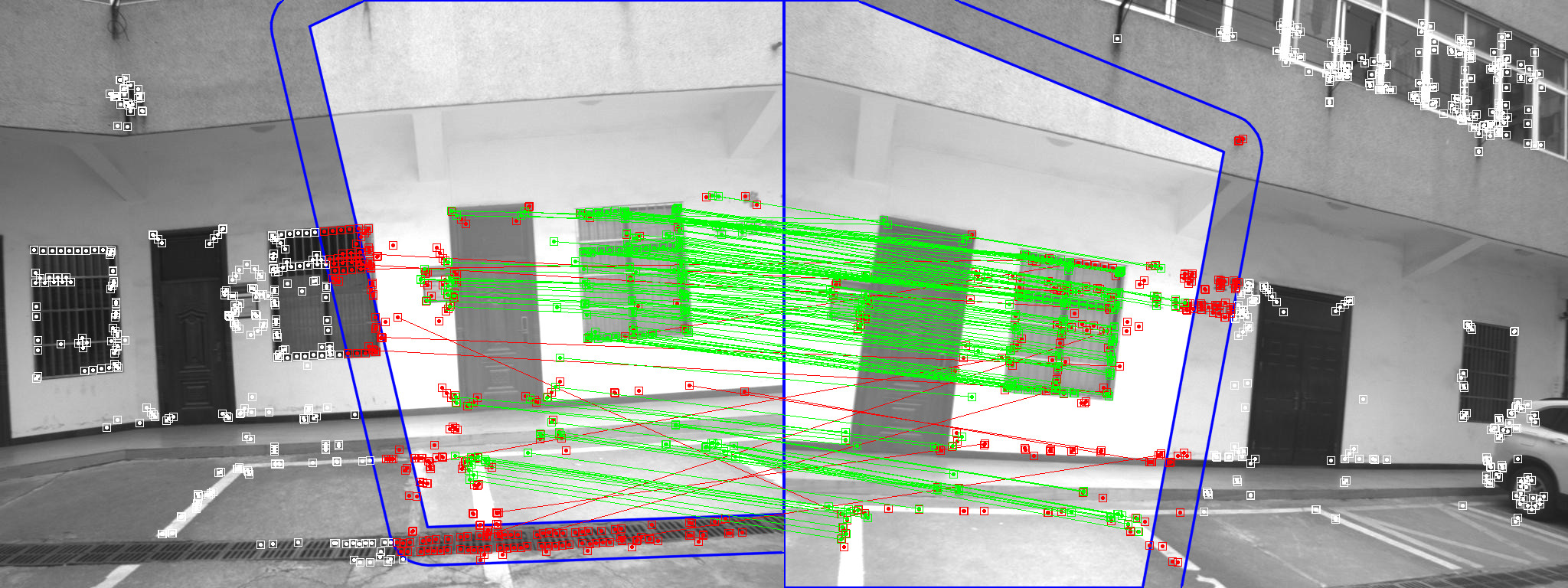}
\caption{Schematic of feature matching in the covisibility region found using the relative rotation by assuming a pure-rotation motion. White features are out-region ones and not involved in matching, red features are in-region (blue polygon) but matching-failed ones (outlier rejection using \cite{kneip2013direct}), and green features are in-region and matched ones. Considering the existence of translation motion, an expanded buffer (blue rounded polygon) is constructed to relax the assumption of the pure-rotation motion.}
\label{fig:sfm_covisibility}
\end{figure}

The structure from motion (SfM) is first performed for each monocular camera to recover visual structure, namely up-to-scale poses and landmarks.
To accelerate feature matching and visual reconstruction in SfM, we first ($i$) conduct a rotation-only visual odometry to obtain rotation sequence, and similar to LiDAR, ($ii$) employ the rotation-only hand-eye alignment to recover the visual extrinsic rotation, finally ($iii$) perform a rotation-aided vision-only SfM.
Specifically, in the rotation-only visual odometry, feature pairs of two consecutive frames tracked by KLT sparse optical flow algorithm \cite{lucas1981iterative} would be undistorted and involved in a frame-to-frame direct rotation optimization \cite{kneip2013direct}.
Different from pose recovery using fundamental or essential matrix models, the direct rotation optimization method decoupled the rotation estimation with translation, thus relative rotation can be solved accurately even for structural degeneracy or zero-translation motion.
After rotations are recovered, a least-squares problem similar to (\ref{equ:lidar_rot_hand_eye}) would be organized and solved to obtain extrinsic rotation and time offset.
At this point, the accurate world-frame rotation of each image can be computed based on the rotation B-spline, extrinsic rotation, and time offset.
In SfM, by assuming a pure-rotation motion, SIFT \cite{lowe2004distinctive}-based feature matching between \textbf{arbitrary} two images is performed in the covisibility region (if it exists), which can be found by relative rotation based on priori world-frame rotations, see Fig. \ref{fig:sfm_covisibility}.
The final incremental SfM would be performed in COLMAP \cite{schoenberger2016sfm}.

Similar to LiDAR-inertial alignment, equations similar to (\ref{equ:ii_align}), (\ref{equ:li_align_vel}), and (\ref{equ:li_align_pos}) would be organized together as visual-inertial alignment constraints.
The difference is that SfM-derived visual poses are up-to-scale ones since monocular camera lacks observability for metric scale.
By introducing explicit global visual scale, the position $\translation{b^r}{w}(\tau)$ in (\ref{equ:li_align_pos}) can be rewritten as:
\begin{equation}
\small
\label{equ:vi_align_pos}
\translation{b^r}{w}(c^k_{\tau})\triangleq
\translation{b^r}{w}(\tau)=
\beta^k\cdot\rotation{m^k}{w}\cdot\translation{c^k}{m^k}(\tau)+\translation{m^k}{w}-\rotation{b^r}{w}(\tau)\cdot\translation{c^k}{b^r}
\end{equation}
where $\translation{c^k}{m^k}(\tau)$ denotes the SfM-derived up-to-scale position at time $\tau$; $\beta^k$ denotes the global scale of the visual structure, which would be estimated in alignment.

\subsubsection{One-Shot Sensor-Inertial Alignment Optimization}
\label{sect:one_shot_align}
Finally, the least-squares problem of the sensor-inertial alignment can be described as follows:
\begin{equation}
\small
\label{equ:ls_one_shot_align}
\begin{gathered}
\hat{\bsm{x}}_{\mathrm{align}} \cup
\left\lbrace \hat{\bsm{v}}_{b^r}^{w}(\tau)\right\rbrace \cup
\left\lbrace \hat{\bsm{v}}_{l^k}^{w}(\tau)\right\rbrace \cup
\left\lbrace \hat{\bsm{v}}_{c^m}^{w}(\tau)\right\rbrace \cup
\left\lbrace \hat{\beta}^{m}\right\rbrace
=
\\
\begin{aligned}
\arg\min\sum_n^{\mathcal{S_\mathrm{tp}}}\bigg(&\sum_i^{n_b}
\left\| 
r_{v}(b^i_{n},b^i_{{n\smallplus 1}})
\right\| ^2
+
\sum_j^{n_r}
\left\| 
r_{v}(r^j_{n},r^j_{{n\smallplus 1}})
\right\| ^2
\\
+&
\sum_k^{n_l}
\left\| 
r_{v}(l^k_{n},l^k_{{n\smallplus 1}})
\right\| ^2
\;+
\sum_m^{n_c}\left\| 
r_{p}(l^k_{n},l^k_{{n\smallplus 1}})
\right\| ^2
\\
+&
\sum_m^{n_c}
 \left\| 
r_{v}(c^m_{n},c^m_{{n\smallplus 1}})
\right\| ^2
+
\sum_m^{n_c}\left\| 
r_{p}(c^m_{n},c^m_{{n\smallplus 1}})
\right\| ^2
\bigg)
\end{aligned}
\end{gathered}
\end{equation} 
with
\begin{small}
\begin{alignat}{4}
&r_{v}(b^i_{n},b^i_{{n\smallplus 1}})&&\triangleq
\linvelhat{b^r}{w}(\tau_{n\smallplus 1})&&-
\linvelhat{b^r}{w}(\tau_n)&&-
\Delta\bsm{v}_{n,n\smallplus 1}^{r,i}
\nonumber
\\
&r_{v}(r^j_{n},r^j_{{n\smallplus 1}})&&\triangleq
\linvel{b^r}{w}(r^j_{{n\smallplus 1}})&&-
\linvel{b^r}{w}(r^j_{{n}})&&-
\Delta\bsm{v}_{n,n\smallplus 1}^{r,r}
\nonumber
\\
&r_{v}(l^k_{n},l^k_{{n\smallplus 1}})&&\triangleq
\linvel{b^r}{w}(l^k_{{n\smallplus 1}})&&-
\linvel{b^r}{w}(l^k_{{n}})&&-
\Delta\bsm{v}_{n,n\smallplus 1}^{r,r}
\\
&r_{p}(l^k_{n},l^k_{{n\smallplus 1}})&&\triangleq
\translation{b^r}{w}(l^k_{{n\smallplus 1}})&&-
\translation{b^r}{w}(l^k_{{n}})&&-
\Delta\bsm{p}_{n,n\smallplus 1}^{r,r}
\nonumber
\\
&r_{v}(c^m_{n},c^m_{{n\smallplus 1}})&&\triangleq
\linvel{b^r}{w}(c^m_{{n\smallplus 1}})&&-
\linvel{b^r}{w}(c^m_{{n}})&&-
\Delta\bsm{v}_{n,n\smallplus 1}^{r,r}
\nonumber
\\
&r_{p}(c^m_{n},c^m_{{n\smallplus 1}})&&\triangleq
\translation{b^r}{w}(c^m_{{n\smallplus 1}})&&-
\translation{b^r}{w}(c^m_{{n}})&&-
\Delta\bsm{p}_{n,n\smallplus 1}^{r,r}
\nonumber
\end{alignat}
\end{small}
and
\begin{equation}
\small
\begin{aligned}
\hat{\bsm{x}}_{\mathrm{align}}&\triangleq
\hat{\bsm{x}}_{\mathrm{rot}}^r \cup
\hat{\bsm{x}}_{\mathrm{pos}}\cup
\gravityhat{w}
\\
\Delta\bsm{v}_{n,n\smallplus 1}^{r,i}&\triangleq
\tilde{\bsm{c}}^i_{n,n\smallplus 1}-\tilde{\bsm{A}}^i_{n,n\smallplus 1}
\cdot\translationhat{b^i}{b^r}+\gravityhat{w}\cdot\Delta\tau_{n,n\smallplus 1}
\\
\Delta\bsm{p}_{n,n\smallplus 1}^{r,i}&\triangleq
\tilde{\bsm{d}}^i_{n,n\smallplus 1}-\tilde{\bsm{B}}^i_{n,n\smallplus 1}
\cdot\translationhat{b^i}{b^r}
\\
&+\linvel{b^r}{w}(l^k_{n}\mid c^m_n)\cdot\Delta\tau_{n,n\smallplus 1}+
\frac{1}{2}\cdot\gravityhat{w}\cdot\Delta^2\tau_{n,n\smallplus 1}
\end{aligned}
\end{equation}
where $\linvel{b^r}{w}(r^j_{n})$ from (\ref{equ:ri_align}), $\linvel{b^r}{w}(l^k_{n}\mid c^m_n)$ from (\ref{equ:li_align_vel}),  $\translation{b^r}{w}(l^k_{n})$ from (\ref{equ:li_align_pos}), and $\translation{b^r}{w}(c^m_{n})$ from (\ref{equ:vi_align_pos}) are sensor-derived kinematics; 
$\Delta\bsm{v}_{n,n\smallplus 1}^{r,i}$ and $\Delta\bsm{p}_{n,n\smallplus 1}^{r,i}$ are velocity and position variations of $\coordframe{b^r}$ in time piece $[\tau_n,\tau_{n\smallplus 1})$, computed based on inertial measurements of $\coordframe{b^i}$;
$\mathcal{S}_\mathrm{tp}$ denotes time piece sequence.
Note that the world-frame gravity vector is solved on the sphere manifold to ensure a constant magnitude.
After alignment optimization, the SfM-derived up-to-scale visual structures (landmarks and poses) would be scaled using the estimated global scale factors, if cameras are involved in calibration.

\subsection{Linear Scale B-spline Recovery}
\label{sect:init_lin_scale_spline_recovery}
Based on the initialized quantities from the previous two stages, the linear scale B-spline living in $\mathbb{R}^3$ would be recovered, whose type is resiliently selected from ($i$) linear acceleration B-spline, ($ii$) linear velocity B-spline, and ($iii$) linear translation B-spline, based on the sensor suite to be calibrated.
Note that such resilient modeling aims to maximize calibration usability in terms of low computation complexity and high calibration accuracy.

\subsubsection{Linear Acceleration B-spline Recovery}
When only multiple IMUs are involved in spatiotemporal calibration, i.e., 
\begin{equation}
\small
\begin{gathered}
n_b\ge 2,\;n_r,n_l,n_c\equiv 0,
\end{gathered}
\end{equation}
the linear acceleration B-spline would be treated as the linear scale B-spline, and recovered using raw specific force measurements from multiple accelerometers, which could be conducted by solving the following least-squares problem:
\begin{equation}
\small
\label{equ:ls_lin_acce_spline_recovery}
\hat{\mathcal{S}}
=\arg\min
\sum_{i}^{n_b}\sum_{n}^{\mathcal{D}^i_{a}}\left\| 
r_{a}\left( \tilde{\bsm{a}}^i_n\right) 
\right\| ^2_{\bsm{Q}_{a,n}^i}
\end{equation}
with
\begin{equation}
\small
\label{equ:acce_residual}
\begin{aligned}
&r_{a}\left( \tilde{\bsm{a}}^i_n\right)     \triangleq
h_a\left( {\bsm{a}}^{i}(\tau_n+\timeoffsethat{b^i}{b^r}),\bsm{x}_{\mathrm{in}}^{b^i}\right) 
-\tilde{\bsm{a}}^i_n
\\
&{\bsm{a}}^{i}(\tau)=
\left( \rotation{b^r}{w}(\tau)
\cdot\rotationhat{b^i}{b^r} \right)^\top\cdot
\left(\linacce{b^i}{w}(\tau) -\gravity{w}\right) 
\\
&	\linacce{b^i}{w}(\tau)=
\linacce{b^r}{w}(\tau)
+\left( \liehat{\angacce{b^r}{w}(\tau)}+\liehat{\angvel{b^r}{w}(\tau)}^2
\right) 
\cdot\rotation{b^r}{w}(\tau)\cdot\translation{b^i}{b^r}
\end{aligned}
\end{equation}
where $\mathcal{D}^i_{a}$ denotes the noisy specific force data sequence from the $i$-th accelerometer, in which $\tilde{\bsm{a}}^i_n$ is the $n$-th measurement at time $\tau_n$ stamped by $\coordframe{b^i}$;
$r_{a}\left( \cdot\right)$ is the accelerometer residual with information matrix ${\bsm{Q}_{a,n}^i}$; Similar as $\rotation{b^r}{w}(\tau)$, $\angvel{b^r}{w}(\tau)$, and $\angacce{b^r}{w}(\tau)$, the linear acceleration $\linacce{b^r}{w}(\tau)$ could be analytically obtained from the linear scale B-spline based on (\ref{equ:b-spline-scale}), which exactly involves the linear scale control points into the optimization.

\subsubsection{Linear Velocity B-spline Recovery}
When additional radars are involved in spatiotemporal calibration, i.e., 
\begin{equation}
\small
\begin{gathered}
n_b+n_r\ge 2,\;n_l,n_c\equiv 0,
\end{gathered}
\end{equation}
the linear velocity B-spline is treated as the linear scale B-spline, and recovered using raw target measurements from multiple radars and the specific force data from the reference IMU, which could be conducted by solving the following least-squares problem:
\begin{equation}
\small
\label{equ:ls_lin_vel_spline_recovery}
\begin{aligned}
\left\lbrace \hat{\mathcal{S}},\hat{\bsm{x}}_{\mathrm{tm}}^r\right\rbrace 
=\arg\min
\sum_{j}^{n_r}&\sum_{n}^{\mathcal{D}^j_{r}}\left\| 
r_{r}\left(  \translationtilde{n}{r^j},\tilde{v}^{r^j}_{n}\right) 
\right\| ^2_{\bsm{Q}_{r,n}^j}
\\
+
&\sum_{n}^{\mathcal{D}^r_{a}}\left\| 
r_{a}\left( \tilde{\bsm{a}}^r_n\right) 
\right\| ^2_{\bsm{Q}_{a,n}^r}
\end{aligned}
\end{equation}
with
\begin{equation}
\small
\label{equ:radar_residual}
\begin{aligned}
r_{r}\left(  \translationtilde{n}{r^j},\tilde{v}^{r^j}_{n}\right) &\triangleq
\tilde{v}^{r^j}_{n}+\frac
{\left( \translationtilde{n}{r^j}\right) ^\top\left( \rotation{r^j}{w}(t_n)\right)^\top\cdot\linvel{r^j}{w}(t_n)  }
{\Vert\translationtilde{n}{r^j}\Vert}
\\
\rotation{r^j}{w}(\tau)&=\rotation{b^r}{w}(\tau)\cdot\rotation{r^j}{b^r},
\qquad
t_n\triangleq\tau_n+\timeoffsethat{r^j}{b^r}
\\
\linvel{r^j}{w}(\tau)&=\linvel{b^r}{w}(\tau)+
\liehat{\angvel{b^r}{w}(\tau)}\cdot\rotation{b^r}{w}(\tau)\cdot\translation{r^j}{b^r}
\end{aligned}
\end{equation}
where $\mathcal{D}^j_{r}$ is the noisy target measurements from the $j$-th radar;
$r_{r}\left( \cdot\right)$ denotes the Doppler velocity residual with information matrix ${\bsm{Q}_{r,n}^j}$.
Note that the uninitialized time offsets of radars would be also recovered in this problem by introducing the accelerometer residual of the reference IMU, i.e., $r_{a}\left( \cdot\right)$.

\subsubsection{Linear Translation B-spline Recovery}
When LiDARs or cameras exist in spatiotemporal calibration, i.e., 
\begin{equation}
\small
\begin{gathered}
n_l+n_c\ge 1,
\end{gathered}
\end{equation}
the linear translation B-spline would be treated as the linear scale B-spline and recovered using world-frame NDT-derived and scaled SfM-derived positions, which could be achieved by solving the following least-squares problem:
\begin{equation}
\small
\label{equ:ls_lin_trans_spline_recovery}
\begin{aligned}
\hat{\mathcal{S}}
=\arg\min
&\sum_{k}^{n_l}\sum_{n}^{\mathcal{P}_{l^k}}
\left\| 
r_{p}\left( \translationtilde{l^k}{w}(\tau_n)\right) 
\right\| ^2_{\bsm{Q}_{l,n}^k}
\\
+&\sum_{m}^{n_c}\sum_{n}^{\mathcal{P}_{c^m}}
\left\| 
r_{p}\left( \translationtilde{c^m}{w}(\tau_n)\right) 
\right\| ^2_{\bsm{Q}_{c,n}^m}
\end{aligned}
\end{equation}
with
\begin{equation}
\small
\begin{gathered}
r_{p}\left( \translationtilde{\mathrm{sor}}{w}(\tau_n)\right)
\triangleq
\rotation{b^r}{w}(t_n)\cdot\translation{\mathrm{sor}}{b^r}+\translation{b^r}{w}(t_n)-\translationtilde{\mathrm{sor}}{w}(\tau_n)
\\
t_n\triangleq \tau_n+\timeoffset{\mathrm{sor}}{b^r},
\quad
\mathcal{P}_\mathrm{sor}\triangleq\left\lbrace \translationtilde{\mathrm{sor}}{w}(\tau_n)\right\rbrace,
\quad
\mathrm{sor}\simeq l^k/c^m
\end{gathered}
\end{equation}
where $\mathcal{P}_{l^k}$ and $\mathcal{P}_{c^m}$ denote the position sequence of $\coordframe{l^k}$ and $\coordframe{c^m}$, respectively.
If additional radars are also involved in this case, their time offsets would be recovered subsequently by solving the least-squares problem described in (\ref{equ:ls_lin_vel_spline_recovery}), where linear scale control points would be fixed and not optimized.
The only difference is that the linear velocities of the reference IMU, required in (\ref{equ:ls_lin_vel_spline_recovery}), would be obtained as the first-order kinematics of the linear translation B-spline here, rather than as the zero-order kinematics of linear velocity B-spline.

At this point, \textbf{all parameters} in the state vector have been initialized rigorously, which benefits the following continuous-time-based batch optimization significantly in terms of convergence performance and global optimal solution.
For better readers' understanding of the total initialization process, we provide Table \ref{tab:recovered_states} to show when and where states are recovered.

\begin{table}[t]
\centering
\caption{\textbf{Recovered States In Dynamic Initialization}
\\All parameters are systematically and rigorously initialized
}
\label{tab:recovered_states}
\begin{tabular}{c|lcc}
\toprule
State & Meaning & Recovered In & LSQ Prob. \\ \midrule\midrule
$\mathcal{R}$                           & 
CPs of the rotation B-spline          &
Section \ref{sect:init_rot_spline_recovery}  &              
(\ref{equ:ls_rot_spline_recovery})     \\
$\bsm{x}_{\mathrm{rot}}^b$              & 
extrinsic rotations of IMUs             &
Section \ref{sect:init_rot_spline_recovery}  &              
(\ref{equ:ls_rot_spline_recovery})     \\
$\bsm{x}_{\mathrm{tm}}^b$               & 
time offsets of IMUs                    &
Section \ref{sect:init_rot_spline_recovery}  &              
(\ref{equ:ls_rot_spline_recovery})     \\ \midrule
$\bsm{x}_{\mathrm{rot}}^l$              & 
extrinsic rotations of LiDARs           &
Section \ref{sect:init_sen_inertial_align}   &              
(\ref{equ:lidar_rot_hand_eye})         \\
$\bsm{x}_{\mathrm{tm}}^l$               & 
time offsets of LiDARs                  &
Section \ref{sect:init_sen_inertial_align}   &              
(\ref{equ:lidar_rot_hand_eye})         \\
$\bsm{x}_{\mathrm{rot}}^c$              & 
extrinsic rotations of cameras          &
Section \ref{sect:init_sen_inertial_align}   &              
(\ref{equ:lidar_rot_hand_eye})         \\
$\bsm{x}_{\mathrm{tm}}^c$               & 
time offsets of cameras                 &
Section \ref{sect:init_sen_inertial_align}   &              
(\ref{equ:lidar_rot_hand_eye})         \\
$\bsm{x}_{\mathrm{pos}}$                & 
extrinsic translations                  &
Section \ref{sect:init_sen_inertial_align}   &              
(\ref{equ:ls_one_shot_align})          \\
$\bsm{x}_{\mathrm{rot}}^r$              & 
extrinsic rotations of radars           &
Section \ref{sect:init_sen_inertial_align}   &              
(\ref{equ:ls_one_shot_align})          \\
$\gravity{w}$                           & 
world-frame gravity vector              &
Section \ref{sect:init_sen_inertial_align}   &              
(\ref{equ:ls_one_shot_align})          \\ \midrule
$\mathcal{S}$                           & 
CPs of the scale B-spline             &
Section \ref{sect:init_lin_scale_spline_recovery}  &              
(\ref{equ:ls_lin_acce_spline_recovery}/\ref{equ:ls_lin_vel_spline_recovery}/\ref{equ:ls_lin_trans_spline_recovery})      \\
$\bsm{x}_{\mathrm{tm}}^r$               & 
time offsets of radars                  &
Section \ref{sect:init_lin_scale_spline_recovery}   &              
(\ref{equ:ls_lin_vel_spline_recovery})         \\ \bottomrule
\end{tabular}
\end{table}

\section{Continuous-Time Batch Optimization}
\label{sect:ct_ba}
Based on results from the multi-stage initialization procedure, a continuous-time-based batch optimization graph would be constructed and solved iteratively to refine all initialized parameters to better states.
While raw measurements from IMUs and radars can be tightly coupled into the optimization, scans from LiDARs and images from cameras require data association to construct correspondences for optimization.

\subsection{Data Association}
\label{sect:data_associate}
Point-to-surfel and visual reprojection association would be conducted for LiDARs and cameras respectively, based on NDT-derived rough LiDAR maps and SfM-derived visual structures.

\subsubsection{Point-to-Surfel Association for LiDARs}
Based on the recovered extrinsics and B-splines, the points from multiple LiDARs would be first transformed to the world frame $\coordframe{w}$ and merged as a unique global point cloud map.
This global map is then organized as an octree-based multi-level voxel tree using UFOMap \cite{duberg2020ufomap}, where each voxel stores the mean vector, covariance matrix, and covariance-derived planarity of contained points.

Based on the voxel tree, we first transform each raw LiDAR point from $\coordframe{l^k}$ to $\coordframe{w}$ and then find all candidate voxels that ($i$) contain this up-to-associated point, ($ii$) contain enough points (an empirical threshold) when being constructed, ($iii$) has depth between maximum and minimum depth thresholds, ($iv$) has planarity larger than the minimum planarity threshold, and ($v$) has point-to-surfel distance smaller than maximum distance threshold.
If such voxels exist, we would continue to select a voxel with the largest planarity and construct a corresponding surfel to associate it with this point, as a point-to-surfel correspondence.
To reduce computation complexity, all identified point-to-surfel correspondences would be downsampled, where each surfel is expected to associate a similar number of points.
The final total point-to-surfel correspondences of the $k$-th LiDAR can be expressed as:
\begin{equation}
\small
\begin{aligned}
\mathcal{D}_{l}^k\triangleq\left\lbrace
\left. \tau_n, \translationtilde{n}{l^k},\bsm{e}_s^w\right| 
n\in\mathbb{N},s\in\mathbb{N}
\right\rbrace,\quad
\bsm{e}_s^w\triangleq\begin{bmatrix}
\bsm{n}_s^w \\d_s^w
\end{bmatrix}
\end{aligned}
\end{equation}
where $ \translationtilde{n}{l^k}$ is the $n$-th raw LiDAR point stamped as $\tau_n$ and expressed in $\coordframe{l^k}$; $\bsm{e}_s^w$ denotes the $s$-th world-frame surfel plane defined by normal vector $\bsm{n}_s^w\in\mathbb{R}^3$ and distance $d_s^w\in\mathbb{R}$.

\subsubsection{Visual Reprojection Association for Cameras}
Based on the scaled SfM-derived visual landmarks, visual reprojection association is performed for each camera separately to construct visual reprojection correspondences.

Specifically, given a world-frame visual landmark captured by enough images (an empirical threshold), we first transform it to the camera coordinate frame that first captures it and then obtain its corresponding inverse depth in this image.
The inverse depth together with the raw 2D image observation would be associated with other image observations of this landmark, as visual reprojection correspondences.
Similar to point-to-surfel correspondences, the visual reprojection correspondences would be downsampled to reduce computation complexity.
The final total visual reprojection correspondences of the $m$-th camera can be expressed as:
\begin{equation}
\small
\begin{aligned}
\mathcal{D}_{c}^m\triangleq\left\lbrace
\left. 
\tau_{n},\tilde{\bsm{f}}_l^{c^m_n},\lambda_l^{c^m_n},
\tau_{s},\tilde{\bsm{f}}_l^{c^m_s}
\right| 
n\in\mathbb{N},s\in\mathbb{N}
\right\rbrace
\end{aligned}
\end{equation}
where $\tilde{\bsm{f}}_l^{c^m_n}$ is the first observation of the landmark captured by the $n$-th image stamped as $\tau_{n}$, the corresponding inverse depth is denoted as $\lambda_l^{c^m_n}$; $\tilde{\bsm{f}}_l^{c^m_s}$ is another observation captured by the $s$-th image stamped as $\tau_{s}$.

\subsection{Global Factor Graph Optimization}
\label{sect:batch_opt}
Finally, a global factor graph would be constructed for continuous-time-based batch optimization, where all initialized states would be refined to better ones.
Together five kinds of residuals are involved in the optimization, namely gyroscope residuals and accelerometer residuals for IMUs, Doppler velocity residuals for radars, point-to-surfel residuals for LiDARs, and visual reprojection residuals for cameras.

\subsubsection{Gyroscope and Accelerometer Residuals}
The gyroscope and accelerometer residuals have been defined in (\ref{equ:gyro_residual}) and (\ref{equ:acce_residual}) as $r_{\omega}\left(\tilde{\bsm{\omega}}^i_n\right)$ and $r_{a}\left( \tilde{\bsm{a}}^i_n\right)$ respectively, which introduce the optimization of IMU-related spatiotemporal parameters, gravity vector, and B-splines using raw inertial measurements.

\subsubsection{Radar Doppler Velocity Residual}
The radar Doppler velocity residual has been defined in (\ref{equ:radar_residual}) as $r_{r}\left(  \translationtilde{n}{r^j},\tilde{v}^{r^j}_{n}\right)$, which introduce the optimization of radar-related spatiotemporal parameters and B-splines using raw target measurements.

\subsubsection{LiDAR Point-to-Surfel Residual}
When a LiDAR point-to-surfel correspondence is introduced in optimization, a corresponding point-to-surfel residual can be constructed, which could be described as:
\begin{equation}
\small
r_l\left( \translationtilde{n}{l^k},\bsm{e}_s^w\right) \triangleq
\left( \bsm{n}_s^w\right) ^\top\cdot\left( \rotation{l^k}{w}(t_n)\cdot\translationtilde{n}{l^k}+\translation{l^k}{w}(t_n)\right) +d_s^w
\end{equation}
with
\begin{equation}
\small
\begin{aligned}
\rotation{l^k}{w}(\tau)&=\rotation{b^r}{w}(\tau)\cdot\rotation{l^k}{b^r}
,\quad
t_n\triangleq \tau_n+\timeoffsethat{l^k}{b^r}
\\
\translation{l^k}{w}(\tau)&=\rotation{b^r}{w}(\tau)\cdot\translation{l^k}{b^r}+\translation{b^r}{w}(\tau)
\end{aligned}
\end{equation}
where $\rotation{b^r}{w}(\tau)$ and $\translation{b^r}{w}(\tau)$ are quantities obtained from the B-splines.
Note that this residual introduces the optimization of LiDAR-related spatiotemporal parameters and B-splines.

\subsubsection{Visual Reprojection Residual}
When a visual reprojection correspondence is introduced in optimization, corresponding visual reprojection residuals could be constructed.
Given the reprojection correspondence between landmark observation $\tilde{\bsm{f}}_l^{c^m_n}$ and $\tilde{\bsm{f}}_l^{c^m_s}$, the visual reprojection residual could be described as:
\begin{equation}
\label{equ:visual_reproj_error}
\small
r_c\left( \tilde{\bsm{f}}_l^{c^m_n},\tilde{\bsm{f}}_l^{c^m_s}\right) \triangleq \pi\left( \rotation{c^m_n}{c^m_s}\cdot\translation{l}{c^m_n} +\translation{c^m_n}{c^m_s},\bsm{x}_{\mathrm{in}}^{c^m}\right) -\tilde{\bsm{f}}_l^{c^m_s}
\end{equation}
with
\begin{equation}
\small
\begin{aligned}
\translation{l}{c^m_n}&=\hat{\beta}^m\cdot\pi^{-1}\left( \tilde{\bsm{f}}_l^{c^m_n},\hat{\lambda}_l^{c^m_n},\bsm{x}_{\mathrm{in}}^{c^m}\right)
\\
\transform{c^m_n}{c^m_s}&\triangleq
\left( \transform{b^r}{w}(t_s)\cdot\transform{c^m}{b^r}\right) ^\top
\cdot\transform{b^r}{w}(t_n)\cdot\transform{c^m}{b^r}
\\
t_{(\cdot)}&\triangleq \tau_{(\cdot)}+\timeoffsethat{c^m}{b^r}+\big( \frac{v^{c^m_{(\cdot)}}_{l}}{h^c}-\frac{1}{2}\big) \times \hat{\tau}^{c^m}_\mathrm{red},\quad(\cdot)\simeq n/s
\end{aligned}
\end{equation}
where $\beta^m$ denotes the global scale of the $m$-th camera; $\pi^{-1}(\cdot)$ is the pinhole inverse projection function (the inverse operation of $\pi(\cdot)$, see (\ref{equ:pinhole_visual_proj})), which projects the 2D image observation to camera frame as a 3D landmark using the camera intrinsics $\bsm{x}_{\mathrm{in}}^{c^m}$ and inverse depth $\lambda_l^{c^m_n}$.
Note that this residual introduces the optimization of camera-related spatiotemporal parameters, visual scale factors, and B-splines.

\subsubsection{Continuous-Time Batch Optimization}
Using the residuals from multiple heterogeneous sensors, the final continuous-time batch optimization could be expressed as the following least-squares problem:
\begin{equation}
\small
\label{equ:batch_opt}
\begin{aligned}
\hat{\mathcal{X}}
=\arg\min
&\sum_{i}^{n_b}
\sum_{n}^{\mathcal{D}^i_{\omega}}
\rho_\omega\left(\left\| 
r_{\omega}\left( \tilde{\bsm{\omega}}^i_n\right) 
\right\| ^2_{\bsm{Q}_{\omega,n}^i}\right) 
\;\gets\mathrm{Gyro.\;Mes.}
\\
\mathrm{Acce.\;Mes.}\to\;+
&\sum_{i}^{n_b}
\sum_{n}^{\mathcal{D}^i_{a}}
\rho_a\left(\left\| 
r_{a}\left( \tilde{\bsm{a}}^i_n\right) 
\right\| ^2_{\bsm{Q}_{a,n}^i}\right) 
\\
\mathrm{Rad.\;Mes.}\to\;+
&\sum_{j}^{n_r}
\sum_{n}^{\mathcal{D}^j_{r}}
\rho_r\left( \left\|
r_{r}\left(  \translationtilde{n}{r^j},\tilde{v}^{r^j}_{n}\right)
\right\| ^2_{\bsm{Q}_{r,n}^j}\right) 
\\
\mathrm{Lidar.\;Mes.}\to\;+&\sum_{k}^{n_l}
\sum_{n,s}^{\mathcal{D}^k_{l}}
\rho_l\left(\left\|
r_l\left( \translationtilde{n}{l^k},\bsm{e}_s^w\right)
\right\| ^2_{\bsm{Q}_{l,n,s}^k}\right) 
\\
\mathrm{Cam.\;Mes.}\to\;+&\sum_{m}^{n_c}
\sum_{l,n,s}^{\mathcal{D}^m_{c}}
\rho_c\left(\left\|
r_c\left( \tilde{\bsm{f}}_l^{c^m_n},\tilde{\bsm{f}}_l^{c^m_s}\right)
\right\| ^2_{\bsm{Q}_{c,l,n,s}^m}\right)
\end{aligned}
\end{equation}
where $\bsm{Q}_{(\cdot)}$ denotes the information matrix of a residual; $\rho(\cdot)$ are the Cauchy loss functions to reduce outlier influence, especially for dynamic radar targets and inaccurate data association of LiDARs and cameras.
The \emph{Ceres solver} \cite{Agarwal_Ceres_Solver_2022} is used for solving this nonlinear problem.

\subsection{Multi-Batch Refinement}
\label{sect:refine}
Considering the high nonlinearity of the continuous-time-based batch optimization problem described in (\ref{equ:batch_opt}), we progressively select parameters to optimize in multiple batches, rather than optimize all parameters together in a batch, to ensure better convergence performance.

Specifically, we fix IMU-related spatiotemporal parameters and optimize other ones in the state vector in the first batch, since IMU-related spatiotemporal parameters recovered in the initialization procedure are accurate enough compared with other ones, due to the adequacy of inertial measurements and the rigorousness of initialization, see Table \ref{tab:m-clri-calib-init-final-diff}.
Considering the potential inaccuracy of visual scales recovered in initialization, we also optimize the inverse depth of visual features in the first batch.
In the second batch, IMU-related spatiotemporal parameters would be incrementally introduced in the factor graph, and optimized together with other parameters in the state vector.
If inertial intrinsics, especially the bias factors, have strong observability, they would also be estimated in this batch.
Otherwise, they are required to be statically pre-calibrated in a separate process (generally required in the multi-IMU calibration), see Appendix \hyperref[sect:app_inertial_intri_calib]{A}.

Note that when LiDARs are involved in calibration, the global LiDAR map and the point-to-surfel data association would be reconstructed before each batch optimization based on the estimates from the last optimization, i.e., the LiDAR-related spatiotemporal parameters and B-splines.
Considering the accuracy of the global LiDAR map is crucial for LiDAR-related spatiotemporal calibration, another batch optimization (the third one) would be performed when LiDARs are involved, where the parameters to be optimized remain the same as those in the second batch optimization, but are associated with more accurate global LiDAR map and point-to-surfel correspondences in the factor graph.

\section{Real-World Experiments}
\label{sect:experiments}
To validate the feasibility and effectiveness of the proposed calibration method, extensive real-world experiments were conducted, utilizing both self-collected and publicly available datasets.
To ensure the evaluation adequacy, we performed calibration experiments on four kinds of minimum primitives of multi-sensor integration, namely $(i)$ \textbf{M-I}: IMU-only multi-IMU, $(ii)$ \textbf{M-RI}: multi-radar multi-IMU, $(iii)$ \textbf{M-LI}: multi-LiDAR multi-IMU, $(iv)$ \textbf{M-CI}: multi-camera multi-IMU, and the maximum multi-sensor integration, namely $(v)$ \textbf{M-CLRI}: multi-camera multi-LiDAR multi-radar multi-IMU.
Note that the proposed resilient calibration method not only supports such five kinds of sensor suites, but also other ones, see Fig. \ref{fig:ikalibr}.
We only consider such five kinds of sensor suites in the following experiments, as they are the most descriptive suites on resilient calibration performance in \emph{iKalibr}.

\subsection{Dataset}
\subsubsection{Our Dataset}
\begin{figure}[t]
\centering
\includegraphics[width=\figscale\linewidth]{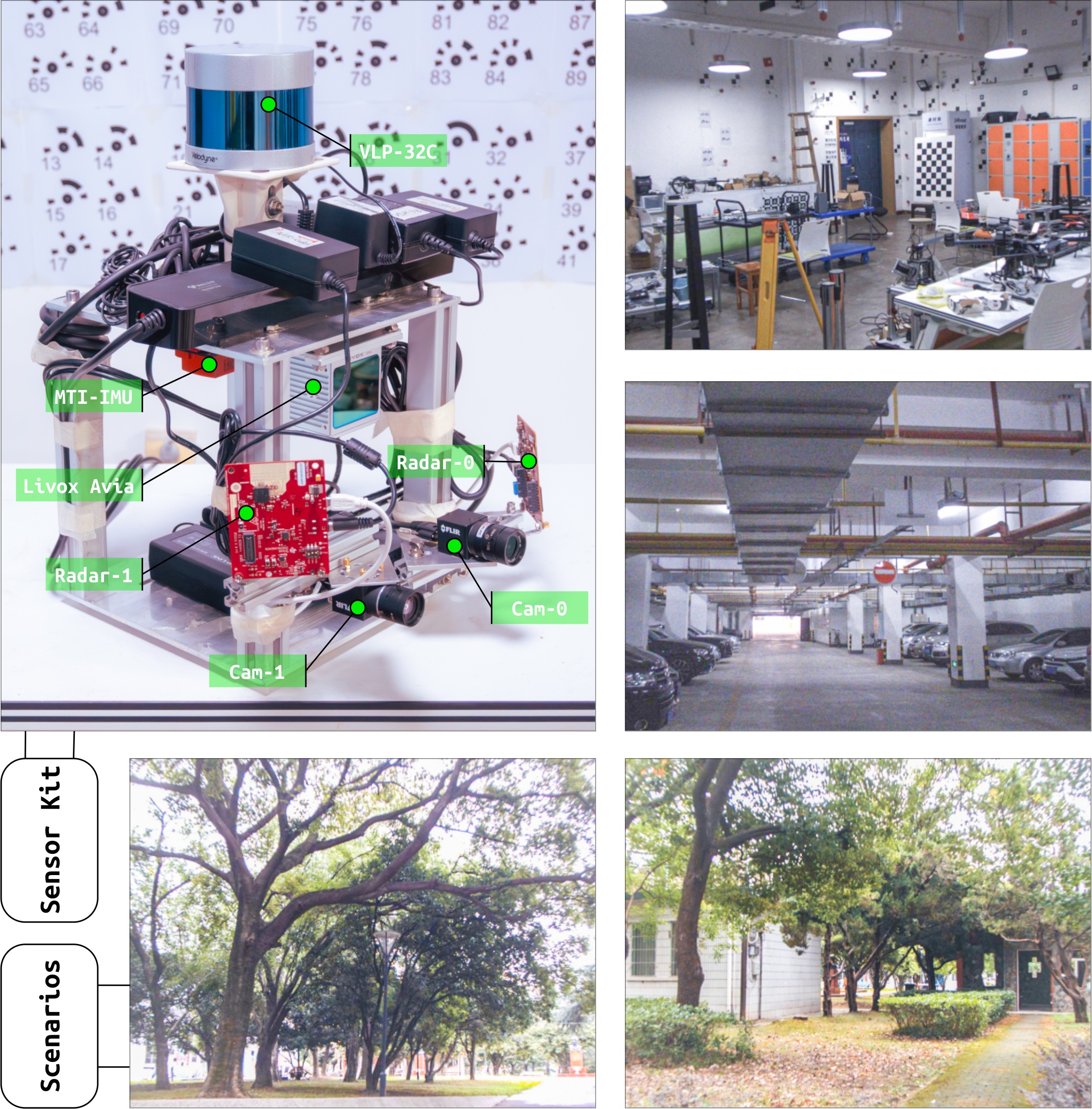}
\caption{The self-assembled sensor suite and typical scenarios in real-world experiments. Scenario images are picked from those that cameras captured.} 
\label{fig:kit}
\end{figure}

While a wealth of multi-sensor datasets for localization and mapping evaluation are publicly available and suitable for motion-based, targetless spatiotemporal calibration, most of them feature platforms with low-excitation stable motion, limiting their ability to support motion-based accurate spatiotemporal calibration.
This motivates us to undertake our own data collection for a reliable spatiotemporal evaluation on the proposed \emph{iKalibr}.
Specifically, we assembled a hardware platform shown in Fig. \ref{fig:kit}, which integrates two FLIR cameras (denoted as \emph{Cam-0} and \emph{Cam-1}), a Velodyne VLP-32C LiDAR (denoted as \emph{LiDAR-0}), a Livox Avia LiDAR (denoted as \emph{LiDAR-1}), two TI AWR1843 3D Radars (denoted as \emph{Rad-0} and \emph{Rad-1}), a MTI-G-710 MEMS IMU (denoted as \emph{IMU-0}), and a Livox Avia built-in MEMS IMU (denoted as \emph{IMU-1}).
The acquisition rate for cameras, LiDARs, and radars is set to 10 Hz, as for \emph{IMU-0} and \emph{IMU-1}, rates are set to 400 Hz and 200 Hz, respectively.
All sensors are only software-synchronized, thus unknown constant time offsets exist, whose determination would be considered in calibration.

To ensure the completeness of the evaluation, several scenarios were encompassed in experiments, covering indoor and outdoor, large and small scales, and structuredless and structured cases, see Fig. \ref{fig:kit}.
Data sequences lasting 30 to 60 seconds were collected in each scenario for experiment evaluation.

\subsubsection{Public Dataset}
While our datasets are open-sourced, to strengthen the evaluation reliability further, we also involved three public datasets that meet the demand for motion-based targetless calibration in experiments:
\begin{enumerate}
\item \emph{LI-Calib} dataset \cite{lv2020targetless}: contains ten data sequences collected handheld in both indoor (garage) and outdoor (court) scenarios.
The self-assembled sensor rig integrates three XSens-100 IMUs (denoted as \emph{IMU-0}, \emph{IMU-1}, and \emph{IMU-2}) sampled at
400 Hz and a Velodyne VLP-16 LiDAR sampled at 10 Hz.
This dataset would be utilized in the \textbf{M-I} and \textbf{M-LI} experiments.
Sensors except the LiDAR are hardware-synchronized.

\item \emph{TUM GS-RS} dataset \cite{schubert2019rolling}: contains ten data sequences collected handheld in the indoor scenarios.
The self-assembled sensor rig integrates two uEye UI-3241LE-M-GL cameras (denoted as \emph{Cam-0} and \emph{Cam-1}) sampled at 20 Hz and a Bosch BMI160 IMU sampled at 200 Hz.
The \emph{Cam-0} runs in GS mode and the \emph{Cam-1} in RS mode.
This dataset would be utilized in the \textbf{M-CI} experiment.
Sensors are hardware-synchronized.

\item \emph{River} dataset \cite{chen2024river}: contains three data sequences collected handheld in indoor scenario.
The self-assembled sensor rig integrates two AWR1843BOOST 3D
radars (denoted as \emph{Rad-0} and \emph{Rad-1}) sampled at 10 Hz and an XSens MTI-G-710 MESE IMU sampled at 400 Hz.
This dataset would be utilized in the \textbf{M-RI} experiment.
Sensors are only software-synchronized.
\end{enumerate}

\subsection{IMU-Only Multi-IMU (\textbf{M-I}) Calibration}
\label{sect:m-i-calib}

\begin{table}[t]
\renewcommand{\arraystretch}{\tabheight}
\setlength{\tabcolsep}{\tabwidth}
\centering
\caption{\textbf{Spatiotemporal Calibration Results in M-I Experiments}
\\iKalibr enables both spatial and temporal calibration for M-I suites with high repeatability
}
\label{tab:m-i-calib}
\begin{threeparttable}
\begin{tabular}{ccc|rrc}
\toprule
\multicolumn{3}{c|}{Method}                                                                                                                                                                       & \multicolumn{1}{c}{Ours}    & \multicolumn{1}{c}{Mix-Cal \cite{lee2022extrinsic}} & Pseudo GT                          \\ \midrule\midrule
\multicolumn{1}{c|}{\multirow{12}{*}{\rotatebox{90}{Our Dataset}}}                               & \multicolumn{1}{c|}{\multirow{12}{*}{\rotatebox{90}{IMU-1}}} & $\hat{p}_{x}$    & 2.901$\pm$\textbf{0.061}    & 2.913$\pm$0.097                                     & \ding{55}                          \\
\multicolumn{1}{c|}{}                                                                           & \multicolumn{1}{c|}{}                                                        & $\hat{p}_{y}$    & -12.280$\pm$\textbf{0.044}  & -12.251$\pm$0.080                                   & \ding{55}                          \\
\multicolumn{1}{c|}{}                                                                           & \multicolumn{1}{c|}{}                                                        & $\hat{p}_{z}$    & -3.199$\pm$\textbf{0.039}   & -3.148$\pm$0.055                                    & \ding{55}                          \\ \cmidrule{3-6} 
\multicolumn{1}{c|}{}                                                                           & \multicolumn{1}{c|}{}                                                        & $\hat{\theta}_r$ & -179.795$\pm$\textbf{0.013} & -179.787$\pm$0.047                                  & \ding{55}                          \\
\multicolumn{1}{c|}{}                                                                           & \multicolumn{1}{c|}{}                                                        & $\hat{\theta}_p$ & -0.462$\pm$\textbf{0.036}   & -0.474$\pm$0.079                                    & \ding{55}                          \\
\multicolumn{1}{c|}{}                                                                           & \multicolumn{1}{c|}{}                                                        & $\hat{\theta}_y$ & -87.599$\pm$0.149           & -87.615$\pm$\textbf{0.118}                          & \ding{55}                          \\ \cmidrule{3-6} 
\multicolumn{1}{c|}{}                                                                           & \multicolumn{1}{c|}{}                                                        & $\hat{\tau}$     & 1.007$\pm$0.196             & \multicolumn{1}{c}{\ding{55}}                       & \ding{55}                          \\ \midrule
\multicolumn{1}{c|}{\multirow{25}{*}{\rotatebox{90}{LI-Calib Dataset \cite{lv2020targetless}}}} & \multicolumn{1}{c|}{\multirow{12}{*}{\rotatebox{90}{IMU-1}}}                  & $\hat{p}_{x}$    & -9.063$\pm$\textbf{0.054}   & -9.088$\pm$0.085                                    & \multicolumn{1}{r}{-9.350\;\;\;}   \\
\multicolumn{1}{c|}{}                                                                           & \multicolumn{1}{c|}{}                                                        & $\hat{p}_{y}$    & 10.104$\pm$\textbf{0.055}   & 10.215$\pm$0.066                                    & \multicolumn{1}{r}{10.100\;\;\;}   \\
\multicolumn{1}{c|}{}                                                                           & \multicolumn{1}{c|}{}                                                        & $\hat{p}_{z}$    & 0.049$\pm$0.048             & 0.080$\pm$\textbf{0.047}                            & \multicolumn{1}{r}{0.000\;\;\;}    \\ \cmidrule{3-6} 
\multicolumn{1}{c|}{}                                                                           & \multicolumn{1}{c|}{}                                                        & $\hat{\theta}_r$ & -0.114$\pm$\textbf{0.021}   & -0.095$\pm$0.055                                    & \multicolumn{1}{r}{0.000\;\;\;}    \\
\multicolumn{1}{c|}{}                                                                           & \multicolumn{1}{c|}{}                                                        & $\hat{\theta}_p$ & 0.207$\pm$\textbf{0.034}    & 0.194$\pm$0.091                                     & \multicolumn{1}{r}{0.000\;\;\;}    \\
\multicolumn{1}{c|}{}                                                                           & \multicolumn{1}{c|}{}                                                        & $\hat{\theta}_y$ & -0.658$\pm$\textbf{0.023}   & -0.663$\pm$0.037                                    & \multicolumn{1}{r}{0.000\;\;\;}    \\ \cmidrule{3-6} 
\multicolumn{1}{c|}{}                                                                           & \multicolumn{1}{c|}{}                                                        & $\hat{\tau}$     & -0.192$\pm$0.742            & \multicolumn{1}{c}{\ding{55}}                       & \multicolumn{1}{r}{0.000\;\;\;}    \\ \cmidrule{2-6} 
\multicolumn{1}{c|}{}                                                                           & \multicolumn{1}{c|}{\multirow{12}{*}{\rotatebox{90}{IMU-2}}}                  & $\hat{p}_{x}$    & -17.576$\pm$\textbf{0.050}  & -17.458$\pm$0.081                                   & \ding{55}                          \\
\multicolumn{1}{c|}{}                                                                           & \multicolumn{1}{c|}{}                                                        & $\hat{p}_{y}$    & -7.633$\pm$0.052            & -7.557$\pm$\textbf{0.029}                           & \ding{55}                          \\
\multicolumn{1}{c|}{}                                                                           & \multicolumn{1}{c|}{}                                                        & $\hat{p}_{z}$    & -0.040$\pm$0.050            & 0.019$\pm$\textbf{0.037}                            & \multicolumn{1}{r}{0.000\;\;\;}    \\ \cmidrule{3-6} 
\multicolumn{1}{c|}{}                                                                           & \multicolumn{1}{c|}{}                                                        & $\hat{\theta}_r$ & -0.128$\pm$\textbf{0.024}   & -0.174$\pm$0.146                                    & \ding{55}                          \\
\multicolumn{1}{c|}{}                                                                           & \multicolumn{1}{c|}{}                                                        & $\hat{\theta}_p$ & 0.636$\pm$\textbf{0.057}    & 0.688$\pm$0.080                                     & \ding{55}                          \\
\multicolumn{1}{c|}{}                                                                           & \multicolumn{1}{c|}{}                                                        & $\hat{\theta}_y$ & -135.941$\pm$\textbf{0.020} & -135.965$\pm$0.085                                  & \multicolumn{1}{r}{-135.000\;\;\;} \\ \cmidrule{3-6} 
\multicolumn{1}{c|}{}                                                                           & \multicolumn{1}{c|}{}                                                        & $\hat{\tau}$     & -0.206$\pm$0.919            & \multicolumn{1}{c}{\ding{55}}                       & \multicolumn{1}{r}{0.000\;\;\;}    \\ \bottomrule
\end{tabular}
\begin{tablenotes} 
\item[*] Extrinsic translations in $(cm)$, extrinsic Euler angles in $(deg)$, and time offsets in $(ms)$. All parameters are with respect to IMU-0. Pseudo GTs are rough references from CAD provided by \cite{lv2020targetless}.
\end{tablenotes}
\end{threeparttable}
\end{table}

\begin{figure}[t]
\centering
\includegraphics[width=\figscale\linewidth]{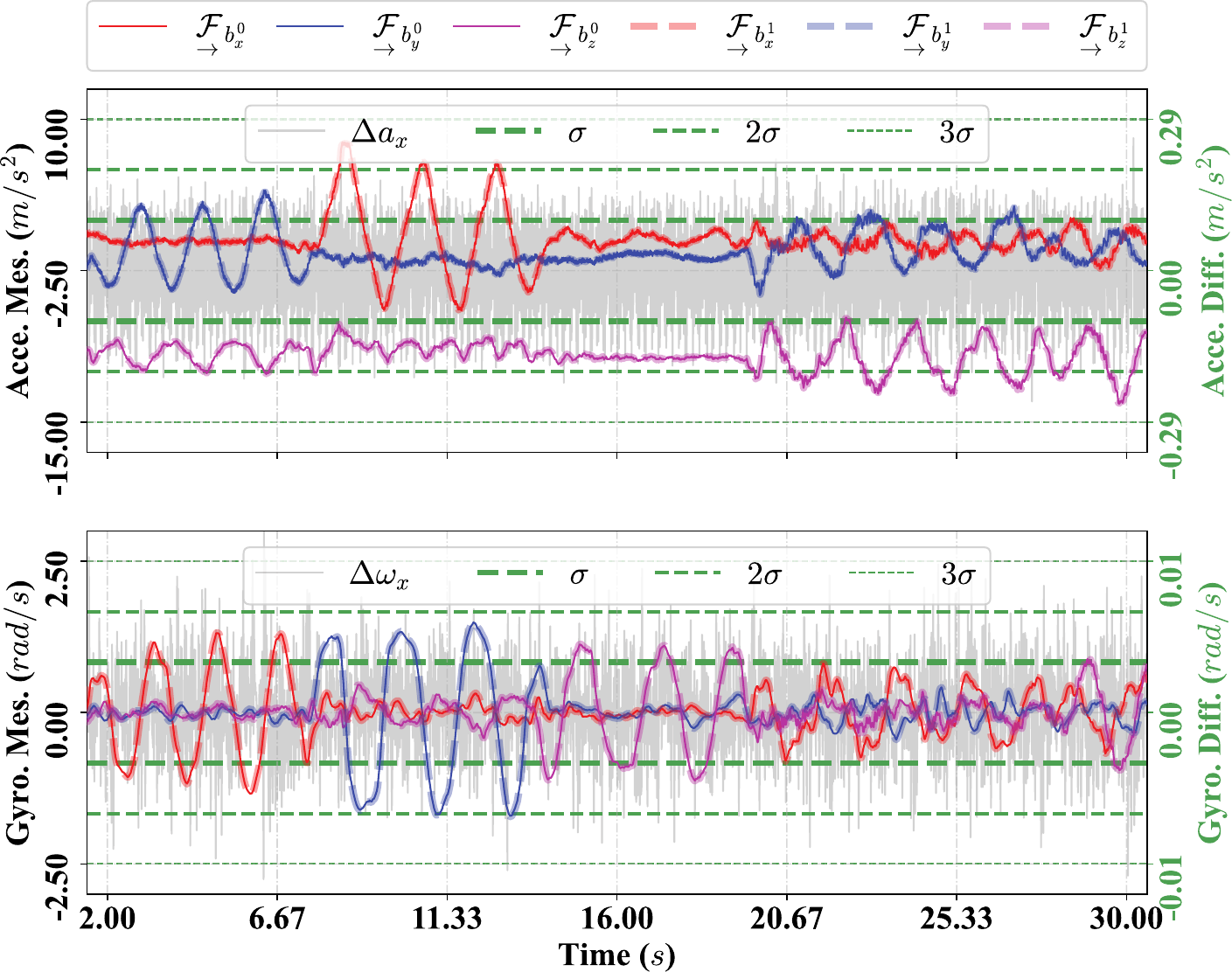}
\caption{
Reprojected inertial measurements (colored) of two IMUs (expressed in the coordinate frame of the reference IMU), and their difference (gray curves denote values while green lines denote corresponding STDs, only x-axis is plotted here). 
As multiple IMUs are rigidly connected, their inertial measurements aligned to a unique frame are expected to be the same.} 
\label{fig:inertial-unique-frame}
\end{figure}

\subsubsection{Quantitative Evaluation and Comparison}
In this experiment, two datasets, namely \emph{LI-Calib} dataset \cite{lv2020targetless} and ours, are utilized.
For both datasets, temporal estimation is incorporated\footnote{
Although three IMUs in the \emph{LI-Calib} dataset are hardware-synchronized, making temporal estimation unnecessary, it is nevertheless conducted here to evaluate the \emph{iKalibr} further, using the known ground truth of time offsets, i.e., 0.0 seconds.
}.
Due to the weak observability of inertial intrinsics in IMU-only multi-IMU calibration, intrinsics were pre-calibrated, see Appendix \hyperref[sect:app_inertial_intri_calib]{A}.
Considering the lack of intrinsic calibration data in \emph{LI-Calib} dataset, inertial intrinsics in \emph{LI-Calib} dataset are obtained from multi-LiDAR multi-IMU calibration, see Section \ref{sect:m-li-calib}.

The proposed \emph{iKalibr} was compared with \emph{Mix-Cal} \cite{lee2022extrinsic}, a state-of-the-art IMU-only dual-IMU extrinsic calibration method.
As inertial time offsets exist in our dataset, to make a fair comparison, inertial measurements given to spatial-only \emph{Mix-Cal} for extrinsic calibration are pre-synchronized based on the time offsets calibrated by \emph{iKalibr}.
Meanwhile, frequencies of two IMUs in our dataset are pre-processed to 200 Hz, to meet the requirement of \emph{Mix-Cal}.
Table \ref{tab:m-i-calib} provides the final calibration results, summarizing spatiotemporal estimates and corresponding standard deviations (STDs).
As can be seen, the difference of estimates from \emph{Mix-Cal} and ours is less than $0.1\;cm$ for translation and $0.1\;deg$ for rotation.
Regarding repeatability, \emph{iKalibr} achieves average STDs of $0.05\;cm$ for translation, $0.05\;deg$ for rotation, and $1.0\;ms$ for time offset, outperforming \emph{Mix-Cal} in general.
\subsubsection{Kinematic Consistency Evaluation}
To make a better presentation of calibration results, using the estimated spatiotemporal parameters and B-splines, we transformed raw inertial measurements of multiple IMUs to a unique coordinate frame, see Fig. \ref{fig:inertial-unique-frame}.
As can be seen, inertial measurements of two IMUs transformed to the same frame match each other well.
The STDs of inertial difference are about $0.1\;m/s^2$ for specific force and $0.03\;rad/s$ for angular velocity.
These results demonstrate the superior calibration accuracy and consistency of \emph{iKaibr}.

\subsection{Multi-Radar Multi-IMU (\textbf{M-RI}) Calibration}
\label{sect:m-ri-calib}

\begin{table}[t]
\renewcommand{\arraystretch}{\tabheight}
\setlength{\tabcolsep}{\tabwidth}
\centering
\caption{\textbf{Spatiotemporal Calibration Results in M-RI Experiments}
\\iKalibr enables both spatial and temporal calibration for M-RI suites with high repeatability
}
\label{tab:m-ri-calib}
\begin{threeparttable}
\begin{tabular}{ccc|rrc}
\toprule
\multicolumn{3}{c|}{Method}                                                                                                                                                        & \multicolumn{1}{c}{Ours}    & \multicolumn{1}{c}{X-RIO \cite{doer2021x}} & Pseudo GT                          \\ \midrule\midrule
\multicolumn{1}{c|}{\multirow{40}{*}{\rotatebox{90}{Our Dataset}}}                        & \multicolumn{1}{c|}{\multirow{12}{*}{\rotatebox{90}{IMU-1 (Livox)}}} & $\hat{p}_{x}$    & 3.193$\pm$\textbf{0.067}    & 5.301$\pm$2.377                            & \ding{55}                          \\
\multicolumn{1}{c|}{}                                                                     & \multicolumn{1}{c|}{}                                               & $\hat{p}_{y}$    & -12.132$\pm$\textbf{0.095}  & -13.525$\pm$1.728                          & \ding{55}                          \\
\multicolumn{1}{c|}{}                                                                     & \multicolumn{1}{c|}{}                                               & $\hat{p}_{z}$    & -3.166$\pm$\textbf{0.106}   & -3.802$\pm$1.611                           & \ding{55}                          \\ \cmidrule{3-6} 
\multicolumn{1}{c|}{}                                                                     & \multicolumn{1}{c|}{}                                               & $\hat{\theta}_r$ & -179.764$\pm$\textbf{0.019} & -178.851$\pm$0.900                         & \ding{55}                          \\
\multicolumn{1}{c|}{}                                                                     & \multicolumn{1}{c|}{}                                               & $\hat{\theta}_p$ & -0.453$\pm$\textbf{0.026}   & -2.047$\pm$0.815                           & \ding{55}                          \\
\multicolumn{1}{c|}{}                                                                     & \multicolumn{1}{c|}{}                                               & $\hat{\theta}_y$ & -87.535$\pm$\textbf{0.143}  & -88.673$\pm$1.018                          & \ding{55}                          \\ \cmidrule{3-6} 
\multicolumn{1}{c|}{}                                                                     & \multicolumn{1}{c|}{}                                               & $\hat{\tau}$     & 0.967$\pm$0.184             & \multicolumn{1}{c}{\ding{55}}              & \ding{55}                          \\ \cmidrule{2-6} 
\multicolumn{1}{c|}{}                                                                     & \multicolumn{1}{c|}{\multirow{12}{*}{\rotatebox{90}{Rad-0}}}         & $\hat{p}_{x}$    & -15.236$\pm$\textbf{0.785}  & -15.710$\pm$1.253                          & \ding{55}                          \\
\multicolumn{1}{c|}{}                                                                     & \multicolumn{1}{c|}{}                                               & $\hat{p}_{y}$    & -22.647$\pm$\textbf{0.710}  & -22.091$\pm$1.559                          & \ding{55}                          \\
\multicolumn{1}{c|}{}                                                                     & \multicolumn{1}{c|}{}                                               & $\hat{p}_{z}$    & 6.536$\pm$\textbf{0.702}    & 6.866$\pm$0.982                            & \ding{55}                          \\ \cmidrule{3-6} 
\multicolumn{1}{c|}{}                                                                     & \multicolumn{1}{c|}{}                                               & $\hat{\theta}_r$ & 174.599$\pm$1.054           & 174.597$\pm$\textbf{0.754}                 & \ding{55}                          \\
\multicolumn{1}{c|}{}                                                                     & \multicolumn{1}{c|}{}                                               & $\hat{\theta}_p$ & 8.563$\pm$0.641             & 8.701$\pm$\textbf{0.622}                   & \ding{55}                          \\
\multicolumn{1}{c|}{}                                                                     & \multicolumn{1}{c|}{}                                               & $\hat{\theta}_y$ & -133.296$\pm$\textbf{0.315} & -133.365$\pm$0.698                         & \ding{55}                          \\ \cmidrule{3-6} 
\multicolumn{1}{c|}{}                                                                     & \multicolumn{1}{c|}{}                                               & $\hat{\tau}$     & -118.777$\pm$1.139          & \multicolumn{1}{c}{\ding{55}}              & \ding{55}                          \\ \cmidrule{2-6} 
\multicolumn{1}{c|}{}                                                                     & \multicolumn{1}{c|}{\multirow{12}{*}{\rotatebox{90}{Rad-1}}}         & $\hat{p}_{x}$    & 15.551$\pm$\textbf{1.380}   & 15.477$\pm$1.821                           & \ding{55}                          \\
\multicolumn{1}{c|}{}                                                                     & \multicolumn{1}{c|}{}                                               & $\hat{p}_{y}$    & -20.631$\pm$\textbf{0.882}  & -20.194$\pm$0.910                          & \ding{55}                          \\
\multicolumn{1}{c|}{}                                                                     & \multicolumn{1}{c|}{}                                               & $\hat{p}_{z}$    & -0.056$\pm$\textbf{0.492}   & -0.165$\pm$1.315                           & \ding{55}                          \\ \cmidrule{3-6} 
\multicolumn{1}{c|}{}                                                                     & \multicolumn{1}{c|}{}                                               & $\hat{\theta}_r$ & -176.790$\pm$0.964          & -176.242$\pm$\textbf{0.872}                & \ding{55}                          \\
\multicolumn{1}{c|}{}                                                                     & \multicolumn{1}{c|}{}                                               & $\hat{\theta}_p$ & 4.533$\pm$\textbf{0.792}    & 4.702$\pm$0.833                            & \ding{55}                          \\
\multicolumn{1}{c|}{}                                                                     & \multicolumn{1}{c|}{}                                               & $\hat{\theta}_y$ & -42.018$\pm$\textbf{0.373}  & -41.829$\pm$0.774                          & \ding{55}                          \\ \cmidrule{3-6} 
\multicolumn{1}{c|}{}                                                                     & \multicolumn{1}{c|}{}                                               & $\hat{\tau}$     & -119.172$\pm$2.123          & \multicolumn{1}{c}{\ding{55}}              & \ding{55}                          \\ \midrule
\multicolumn{1}{c|}{\multirow{25}{*}{\rotatebox{90}{River Dataset \cite{chen2024river}}}} & \multicolumn{1}{c|}{\multirow{12}{*}{\rotatebox{90}{Rad-0}}}         & $\hat{p}_{x}$    & 12.912$\pm$\textbf{0.565}   & 13.022$\pm$1.365                           & \multicolumn{1}{r}{12.495\;\;\;}   \\
\multicolumn{1}{c|}{}                                                                     & \multicolumn{1}{c|}{}                                               & $\hat{p}_{y}$    & -26.172$\pm$\textbf{0.290}  & -26.021$\pm$2.190                          & \multicolumn{1}{r}{-25.812\;\;\;}  \\
\multicolumn{1}{c|}{}                                                                     & \multicolumn{1}{c|}{}                                               & $\hat{p}_{z}$    & 17.507$\pm$\textbf{0.363}   & 17.011$\pm$1.820                           & \multicolumn{1}{r}{18.146\;\;\;}   \\ \cmidrule{3-6} 
\multicolumn{1}{c|}{}                                                                     & \multicolumn{1}{c|}{}                                               & $\hat{\theta}_r$ & 1.551$\pm$1.264             & 2.191$\pm$\textbf{0.810}                   & \multicolumn{1}{r}{3.545\;\;\;}    \\
\multicolumn{1}{c|}{}                                                                     & \multicolumn{1}{c|}{}                                               & $\hat{\theta}_p$ & -7.113$\pm$\textbf{0.661}   & -7.018$\pm$0.721                           & \multicolumn{1}{r}{-7.909\;\;\;}   \\
\multicolumn{1}{c|}{}                                                                     & \multicolumn{1}{c|}{}                                               & $\hat{\theta}_y$ & -41.955$\pm$\textbf{0.670}  & -41.720$\pm$0.919                          & \multicolumn{1}{r}{-42.184\;\;\;}  \\ \cmidrule{3-6} 
\multicolumn{1}{c|}{}                                                                     & \multicolumn{1}{c|}{}                                               & $\hat{\tau}$     & -115.837$\pm$3.748          & \multicolumn{1}{c}{\ding{55}}              & \multicolumn{1}{r}{-115.909\;\;\;} \\ \cmidrule{2-6} 
\multicolumn{1}{c|}{}                                                                     & \multicolumn{1}{c|}{\multirow{12}{*}{\rotatebox{90}{Rad-1}}}         & $\hat{p}_{x}$    & 12.649$\pm$\textbf{0.865}   & 13.208$\pm$1.960                           & \multicolumn{1}{r}{12.890\;\;\;}   \\
\multicolumn{1}{c|}{}                                                                     & \multicolumn{1}{c|}{}                                               & $\hat{p}_{y}$    & 6.455$\pm$\textbf{0.407}    & 7.051$\pm$2.111                            & \multicolumn{1}{r}{6.533\;\;\;}    \\
\multicolumn{1}{c|}{}                                                                     & \multicolumn{1}{c|}{}                                               & $\hat{p}_{z}$    & 16.373$\pm$\textbf{1.010}   & 15.801$\pm$2.016                           & \multicolumn{1}{r}{16.622\;\;\;}   \\ \cmidrule{3-6} 
\multicolumn{1}{c|}{}                                                                     & \multicolumn{1}{c|}{}                                               & $\hat{\theta}_r$ & -0.398$\pm$\textbf{0.779}   & -0.299$\pm$0.810                           & \multicolumn{1}{r}{-0.818\;\;\;}   \\
\multicolumn{1}{c|}{}                                                                     & \multicolumn{1}{c|}{}                                               & $\hat{\theta}_p$ & -1.279$\pm$1.625            & -1.023$\pm$\textbf{1.117}                  & \multicolumn{1}{r}{-2.236\;\;\;}   \\
\multicolumn{1}{c|}{}                                                                     & \multicolumn{1}{c|}{}                                               & $\hat{\theta}_y$ & 49.051$\pm$\textbf{0.293}   & 49.602$\pm$0.612                           & \multicolumn{1}{r}{48.511\;\;\;}   \\ \cmidrule{3-6} 
\multicolumn{1}{c|}{}                                                                     & \multicolumn{1}{c|}{}                                               & $\hat{\tau}$     & -113.701$\pm$0.935          & \multicolumn{1}{c}{\ding{55}}              & \multicolumn{1}{r}{-117.536\;\;\;} \\ \bottomrule
\end{tabular}

\begin{tablenotes} 
\item[*] Extrinsic translations in $(cm)$, extrinsic Euler angles in $(deg)$, and time offsets in $(ms)$. All parameters are with respect to IMU-0. Pseudo GTs are rough references provided by \cite{chen2024river}.
\end{tablenotes}
\end{threeparttable}
\end{table}

\begin{figure}[t]
\centering
\includegraphics[width=\figscale\linewidth]{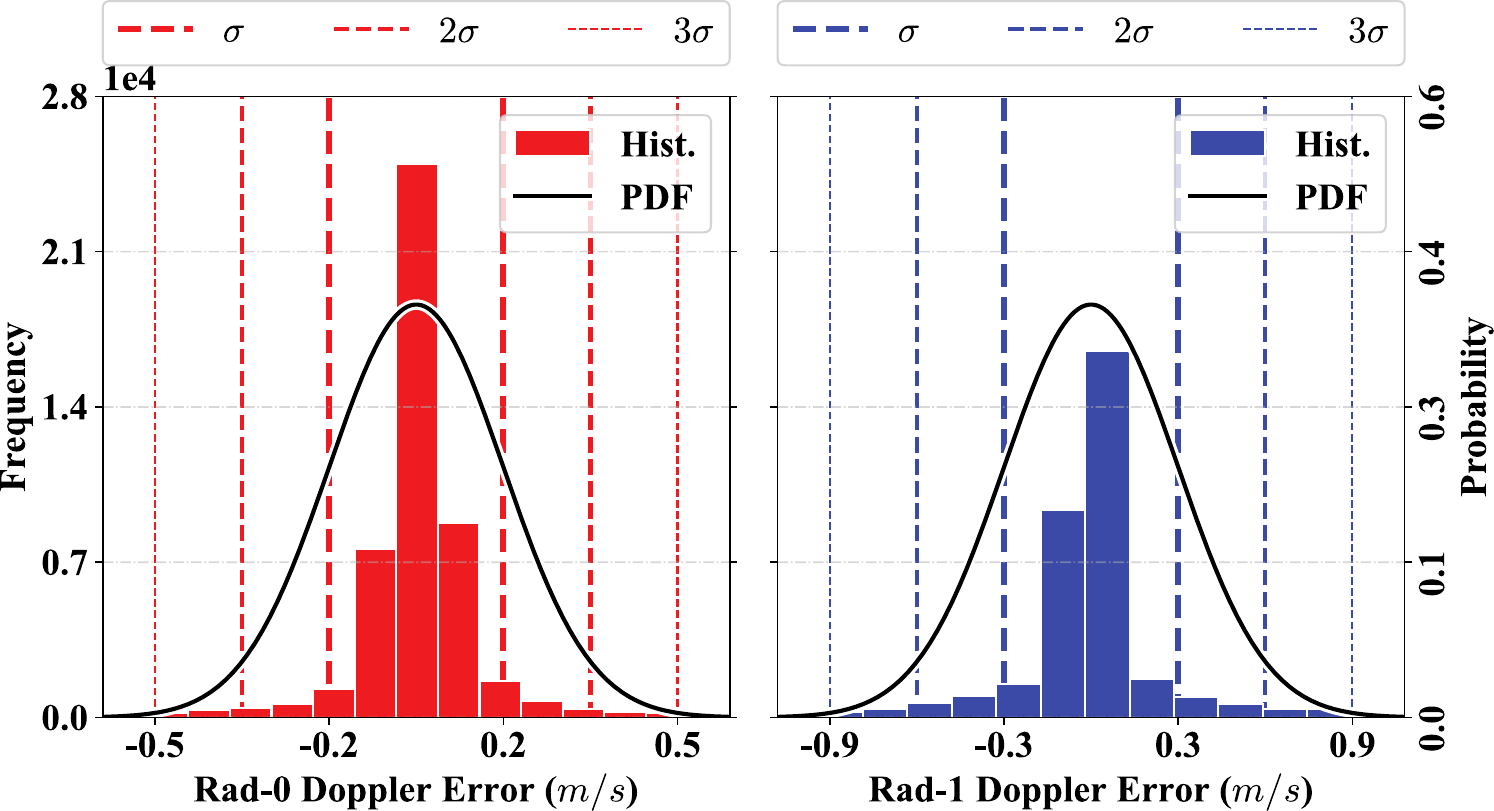}
\caption{The Histograms and probability density functions (PDFs) of Doppler velocity errors (i.e., Doppler velocity residuals defined in (\ref{equ:radar_residual})) for multiple radars.} 
\label{fig:radar-doppler}
\end{figure}

\subsubsection{Quantitative Evaluation and Comparison}
In this experiment, \emph{River} dataset \cite{chen2024river} and ours are utilized for calibration evaluation.
The state-of-the-art radar-inertial odometry with online extrinsic calibration, namely \emph{X-RIO} \cite{doer2021x}, is considered in comparison.
\emph{X-RIO} supports multi-radar single-IMU sensor configuration, which is employed in this evaluation.
In \emph{X-RIO} experiments, multiple radars are grouped with each IMU as a kit to perform the solving process.
Same as \emph{Mix-Cal}, measurements given to \emph{X-RIO} are temporally pre-synchronized.
Meanwhile, the initial spatial guesses given to \emph{X-RIO} are initialization results of \emph{iKalibr}.
Table \ref{tab:m-ri-calib} provides the final calibration results, summarizing spatiotemporal estimates and corresponding STDs.
As can be seen, estimates of two methods are close, while \emph{iKalibr} achieves better repeatability than \emph{X-RIO}.
The average STDs of estimates of \emph{iKalibr} are about $0.8\;cm$ for radar translation, $0.8\;deg$ for radar rotation, and $1.8\;ms$ for time offsets.
Additionally, compared with spatiotemporal parameters of IMUs, those of radars are poorer, as the noise level of radar target measurements is generally much larger than that of inertial measurements.

\subsubsection{Qualitative Evaluation via Radar Velocity Error}
Based on the estimated B-splines and raw radar target measurements, we computed the B-spline-derived Doppler velocities for targets and compared them with the radar-measured ones.
Fig. \ref{fig:radar-doppler} shows the distribution of Doppler velocity errors (as defined in (\ref{equ:radar_residual})) in one multi-radar multi-IMU calibration in \emph{iKalibr}.
Overall, the Doppler errors obey normal distribution, and the average STD is about $0.25\;m/s$.
In particular, the Doppler velocity error distribution of \emph{Rad-1} is poorer than that of \emph{Rad-0}, which matches the calibration results in Table \ref{tab:m-ri-calib}.
This may be due to the differing installation orientations of the two radars (as shown in Fig. \ref{fig:kit}) and their limited FoVs, which impact the quality of the radar data and may lead to variations in the calibration outcomes for the two radars (the configuration of two radars, such as weights and the Cauchy loss function factors, is identical in \emph{iKalibr} solving).

\subsection{Multi-LiDAR Multi-IMU (\textbf{M-LI}) Calibration}
\label{sect:m-li-calib}

\begin{table}[t]
\renewcommand{\arraystretch}{\tabheight}
\setlength{\tabcolsep}{\tabwidth}
\centering
\caption{\textbf{Spatiotemporal Calibration Results in M-LI Experiments}
\\iKalibr enables one-shot multi-IMU multi-LiDAR spatiotemporal calibration with high repeatability
}
\label{tab:m-li-calib}
\begin{threeparttable}
\begin{tabular}{ccc|rrc}
\toprule
\multicolumn{3}{c|}{Method}                                                                                                                                                                & \multicolumn{1}{c}{Ours}    & \multicolumn{1}{c}{OA-Calib \cite{lv2022observability}} & Pseudo GT                          \\ \midrule\midrule
\multicolumn{1}{c|}{\multirow{40}{*}{\rotatebox{90}{Our Dataset}}}                              & \multicolumn{1}{c|}{\multirow{12}{*}{\rotatebox{90}{IMU-1 (Livox)}}}   & $\hat{p}_{x}$    & 2.901$\pm$\textbf{0.060}    & 2.725$\pm$0.329                                         & \ding{55}                          \\
\multicolumn{1}{c|}{}                                                                           & \multicolumn{1}{c|}{}                                                 & $\hat{p}_{y}$    & -12.292$\pm$\textbf{0.048}  & -12.428$\pm$0.433                                       & \ding{55}                          \\
\multicolumn{1}{c|}{}                                                                           & \multicolumn{1}{c|}{}                                                 & $\hat{p}_{z}$    & -3.195$\pm$\textbf{0.037}   & -3.320$\pm$0.206                                        & \ding{55}                          \\ \cmidrule{3-6} 
\multicolumn{1}{c|}{}                                                                           & \multicolumn{1}{c|}{}                                                 & $\hat{\theta}_r$ & -179.800$\pm$\textbf{0.019} & -179.924$\pm$0.175                                      & \ding{55}                          \\
\multicolumn{1}{c|}{}                                                                           & \multicolumn{1}{c|}{}                                                 & $\hat{\theta}_p$ & -0.453$\pm$\textbf{0.029}   & -0.406$\pm$0.143                                        & \ding{55}                          \\
\multicolumn{1}{c|}{}                                                                           & \multicolumn{1}{c|}{}                                                 & $\hat{\theta}_y$ & -87.601$\pm$0.150           & -87.624$\pm$\textbf{0.145}                              & \ding{55}                          \\ \cmidrule{3-6} 
\multicolumn{1}{c|}{}                                                                           & \multicolumn{1}{c|}{}                                                 & $\hat{\tau}$     & 1.056$\pm$\textbf{0.235}    & 1.574$\pm$0.891                                         & \ding{55}                          \\ \cmidrule{2-6} 
\multicolumn{1}{c|}{}                                                                           & \multicolumn{1}{c|}{\multirow{12}{*}{\rotatebox{90}{LiDAR-0 (VLP)}}}   & $\hat{p}_{x}$    & -0.479$\pm$\textbf{0.274}   & -0.501$\pm$0.529                                        & \ding{55}                          \\
\multicolumn{1}{c|}{}                                                                           & \multicolumn{1}{c|}{}                                                 & $\hat{p}_{y}$    & -4.960$\pm$\textbf{0.115}   & -5.352$\pm$0.478                                        & \ding{55}                          \\
\multicolumn{1}{c|}{}                                                                           & \multicolumn{1}{c|}{}                                                 & $\hat{p}_{z}$    & -21.839$\pm$\textbf{0.581}  & -21.839$\pm$0.923                                       & \ding{55}                          \\ \cmidrule{3-6} 
\multicolumn{1}{c|}{}                                                                           & \multicolumn{1}{c|}{}                                                 & $\hat{\theta}_r$ & -179.785$\pm$0.041          & -179.792$\pm$\textbf{0.034}                             & \ding{55}                          \\
\multicolumn{1}{c|}{}                                                                           & \multicolumn{1}{c|}{}                                                 & $\hat{\theta}_p$ & 0.103$\pm$\textbf{0.025}    & 0.149$\pm$0.042                                         & \ding{55}                          \\
\multicolumn{1}{c|}{}                                                                           & \multicolumn{1}{c|}{}                                                 & $\hat{\theta}_y$ & 2.098$\pm$\textbf{0.138}    & 2.115$\pm$0.153                                         & \ding{55}                          \\ \cmidrule{3-6} 
\multicolumn{1}{c|}{}                                                                           & \multicolumn{1}{c|}{}                                                 & $\hat{\tau}$     & 7.190$\pm$\textbf{0.039}    & 7.927$\pm$0.892                                         & \ding{55}                          \\ \cmidrule{2-6} 
\multicolumn{1}{c|}{}                                                                           & \multicolumn{1}{c|}{\multirow{7}{*}{\rotatebox{90}{LiDAR-1 (Livox)}}} & $\hat{p}_{x}$    & \textcolor{red}{\bf{1.188$\pm$1.033}}             & \multicolumn{1}{c}{\ding{55}}                           & \multicolumn{1}{r}{4.165\;\;\;}    \\
\multicolumn{1}{c|}{}                                                                           & \multicolumn{1}{c|}{}                                                 & $\hat{p}_{y}$    & 2.592$\pm$0.310             & \multicolumn{1}{c}{\ding{55}}                           & \multicolumn{1}{r}{2.326\;\;\;}    \\
\multicolumn{1}{c|}{}                                                                           & \multicolumn{1}{c|}{}                                                 & $\hat{p}_{z}$    & -2.260$\pm$0.237            & \multicolumn{1}{c}{\ding{55}}                           & \multicolumn{1}{r}{-2.840\;\;\;}   \\ \cmidrule{3-6} 
\multicolumn{1}{c|}{}                                                                           & \multicolumn{1}{c|}{}                                                 & $\hat{\theta}_r$ & -1.203$\pm$0.025            & \multicolumn{1}{c}{\ding{55}}                           & \multicolumn{1}{r}{0.000\;\;\;}    \\
\multicolumn{1}{c|}{}                                                                           & \multicolumn{1}{c|}{}                                                 & $\hat{\theta}_p$ & -0.582$\pm$0.039            & \multicolumn{1}{c}{\ding{55}}                           & \multicolumn{1}{r}{0.000\;\;\;}    \\
\multicolumn{1}{c|}{}                                                                           & \multicolumn{1}{c|}{}                                                 & $\hat{\theta}_y$ & -0.324$\pm$0.023            & \multicolumn{1}{c}{\ding{55}}                           & \multicolumn{1}{r}{0.000\;\;\;}    \\ \cmidrule{3-6} 
\multicolumn{1}{c|}{}                                                                           & \multicolumn{1}{c|}{}                                                 & $\hat{\tau}$     & 6.792$\pm$0.235             & \multicolumn{1}{c}{\ding{55}}                           & \ding{55}                          \\ \midrule
\multicolumn{1}{c|}{\multirow{40}{*}{\rotatebox{90}{LI-Calib Dataset \cite{lv2020targetless}}}} & \multicolumn{1}{c|}{\multirow{12}{*}{\rotatebox{90}{IMU-1}}}           & $\hat{p}_{x}$    & -9.173$\pm$\textbf{0.056}   & -9.082$\pm$0.360                                        & \multicolumn{1}{r}{-9.350\;\;\;}   \\
\multicolumn{1}{c|}{}                                                                           & \multicolumn{1}{c|}{}                                                 & $\hat{p}_{y}$    & 10.109$\pm$\textbf{0.042}   & 10.237$\pm$0.310                                        & \multicolumn{1}{r}{10.100\;\;\;}   \\
\multicolumn{1}{c|}{}                                                                           & \multicolumn{1}{c|}{}                                                 & $\hat{p}_{z}$    & 0.059$\pm$\textbf{0.052}    & 0.652$\pm$0.542                                         & \multicolumn{1}{r}{0.000\;\;\;}    \\ \cmidrule{3-6} 
\multicolumn{1}{c|}{}                                                                           & \multicolumn{1}{c|}{}                                                 & $\hat{\theta}_r$ & -0.115$\pm$\textbf{0.022}   & -0.248$\pm$0.074                                        & \multicolumn{1}{r}{0.000\;\;\;}    \\
\multicolumn{1}{c|}{}                                                                           & \multicolumn{1}{c|}{}                                                 & $\hat{\theta}_p$ & 0.210$\pm$\textbf{0.040}    & -0.076$\pm$0.171                                        & \multicolumn{1}{r}{0.000\;\;\;}    \\
\multicolumn{1}{c|}{}                                                                           & \multicolumn{1}{c|}{}                                                 & $\hat{\theta}_y$ & -0.659$\pm$\textbf{0.017}   & -0.640$\pm$0.052                                        & \multicolumn{1}{r}{0.000\;\;\;}    \\ \cmidrule{3-6} 
\multicolumn{1}{c|}{}                                                                           & \multicolumn{1}{c|}{}                                                 & $\hat{\tau}$     & -0.208$\pm$\textbf{0.982}   & -0.324$\pm$1.276                                        & \multicolumn{1}{r}{0.000\;\;\;}    \\ \cmidrule{2-6} 
\multicolumn{1}{c|}{}                                                                           & \multicolumn{1}{c|}{\multirow{12}{*}{\rotatebox{90}{IMU-2}}}           & $\hat{p}_{x}$    & -17.673$\pm$\textbf{0.059}  & -17.271$\pm$0.531                                       & \ding{55}                          \\
\multicolumn{1}{c|}{}                                                                           & \multicolumn{1}{c|}{}                                                 & $\hat{p}_{y}$    & -7.619$\pm$\textbf{0.044}   & -7.241$\pm$0.340                                        & \ding{55}                          \\
\multicolumn{1}{c|}{}                                                                           & \multicolumn{1}{c|}{}                                                 & $\hat{p}_{z}$    & -0.028$\pm$\textbf{0.052}   & 0.494$\pm$0.613                                         & \multicolumn{1}{r}{0.000\;\;\;}   \\ \cmidrule{3-6} 
\multicolumn{1}{c|}{}                                                                           & \multicolumn{1}{c|}{}                                                 & $\hat{\theta}_r$ & -0.131$\pm$\textbf{0.031}   & 0.794$\pm$0.042                                         & \ding{55}                          \\
\multicolumn{1}{c|}{}                                                                           & \multicolumn{1}{c|}{}                                                 & $\hat{\theta}_p$ & 0.633$\pm$0.059             & 0.160$\pm$\textbf{0.055}                                & \ding{55}                          \\
\multicolumn{1}{c|}{}                                                                           & \multicolumn{1}{c|}{}                                                 & $\hat{\theta}_y$ & -135.947$\pm$\textbf{0.017} & -135.97$\pm$0.098                                       & \multicolumn{1}{r}{-135.000\;\;\;} \\ \cmidrule{3-6} 
\multicolumn{1}{c|}{}                                                                           & \multicolumn{1}{c|}{}                                                 & $\hat{\tau}$     & -0.352$\pm$1.111            & -0.335$\pm$\textbf{0.901}                               & \multicolumn{1}{r}{0.000\;\;\;}    \\ \cmidrule{2-6} 
\multicolumn{1}{c|}{}                                                                           & \multicolumn{1}{c|}{\multirow{12}{*}{\rotatebox{90}{LiDAR}}}           & $\hat{p}_{x}$    & -10.739$\pm$\textbf{0.153}  & -9.975$\pm$0.501                                        & \ding{55}                          \\
\multicolumn{1}{c|}{}                                                                           & \multicolumn{1}{c|}{}                                                 & $\hat{p}_{y}$    & 0.078$\pm$\textbf{0.099}    & 0.015$\pm$0.311                                         & \ding{55}                          \\
\multicolumn{1}{c|}{}                                                                           & \multicolumn{1}{c|}{}                                                 & $\hat{p}_{z}$    & 14.775$\pm$0.551            & 13.440$\pm$\textbf{0.442}                               & \ding{55}                          \\ \cmidrule{3-6} 
\multicolumn{1}{c|}{}                                                                           & \multicolumn{1}{c|}{}                                                 & $\hat{\theta}_r$ & -0.176$\pm$0.045            & -0.232$\pm$\textbf{0.041}                               & \ding{55}                          \\
\multicolumn{1}{c|}{}                                                                           & \multicolumn{1}{c|}{}                                                 & $\hat{\theta}_p$ & 0.291$\pm$\textbf{0.031}    & 0.213$\pm$0.062                                         & \ding{55}                          \\
\multicolumn{1}{c|}{}                                                                           & \multicolumn{1}{c|}{}                                                 & $\hat{\theta}_y$ & 90.839$\pm$\textbf{0.030}   & 90.896$\pm$0.073                                        & \ding{55}                          \\ \cmidrule{3-6} 
\multicolumn{1}{c|}{}                                                                           & \multicolumn{1}{c|}{}                                                 & $\hat{\tau}$     & 8.926$\pm$0.759             & 8.874$\pm$\textbf{0.753}                                & \ding{55}                          \\ \bottomrule
\end{tabular}

\begin{tablenotes} 
\item[*] Extrinsic translations in $(cm)$, extrinsic Euler angles in $(deg)$, and time offsets in $(ms)$. All parameters, except those of Livox LiDAR, are with respect to IMU-0. GTs of Livox LiDAR (with respect to its built-in IMU) are from the sensor manufacturer, and GTs of LI-Calib dataset are rough references from CAD provided by \cite{lv2020targetless}.
\end{tablenotes}
\end{threeparttable}
\end{table}

\begin{figure}[t]
\centering
\includegraphics[width=\figscale\linewidth]{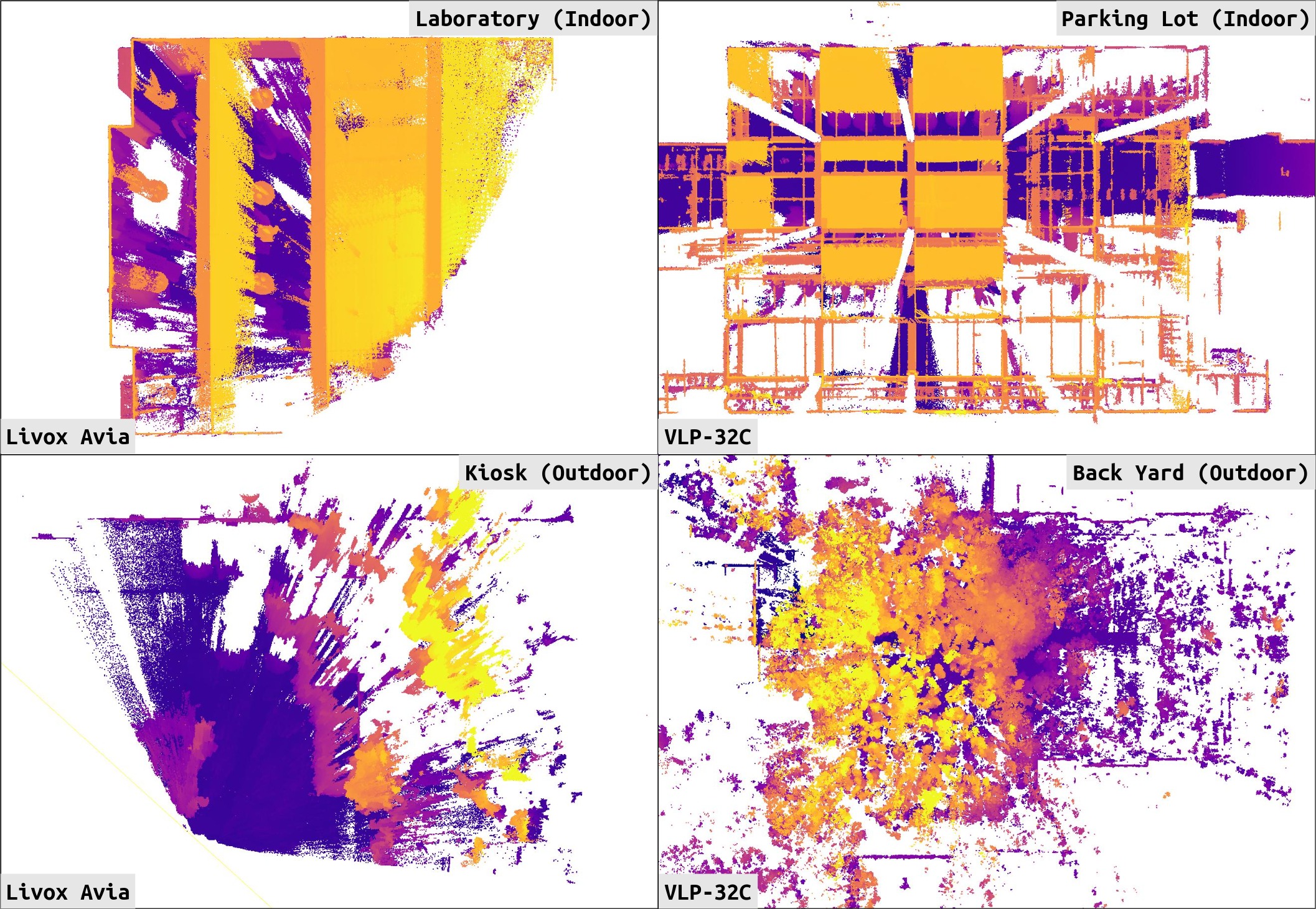}
\caption{The final LiDAR maps (bird eye view, parallel projection) from multi-LiDAR multi-IMU calibration in \emph{iKalibr}. Maps are colorized along the direction of the estimated gravity vector.} 
\label{fig:lidar_maps}
\end{figure}

\subsubsection{Quantitative Evaluation and Comparison}
In this experiment, the open-sourced \emph{LI-Calib} dataset \cite{lv2020targetless} is involved in calibration evaluation, which had been utilized in IMU-only multi-IMU evaluation (see Section \ref{sect:m-i-calib}).
The state-of-the-art single-LiDAR single-IMU spatiotemporal calibration method, namely \emph{OA-Calib} \cite{lv2022observability}, a follow-up work of \emph{LI-Calib} \cite{lv2020targetless} with additional observability-aware module, is considered in comparison.
Table \ref{tab:m-li-calib} provides the calibration results, summarizing spatiotemporal estimates and corresponding STDs.
It can be seen that for inertial spatiotemporal calibration, \emph{iKalibr} generally outperforms \emph{OA-Calib} in terms of repeatability (see STDs).
Such superiority comes from the one-shot calibration of \emph{iKalibr}, which avoids multiple separate single-LiDAR single-IMU calibration procedures, thus capable of better consistency.
Note that the small FoV solid-state Livox LiDARs are not yet supported in \emph{OA-Calib}, thus calibration results of Livox Avia are compared with manufacturer-provided reference.
Interestingly, it can be found that the y-axis and z-axis translation results of Livox Avia LiDAR are well calibrated, matching well with the reference and having small STDs, the x-axis translation is poorly calibrated with a large STD.
This is may related to ($i$) insufficient forward motion excitation and ($ii$) the lack of rear point-to-surfel constraints\footnote{
This point can be observed in Avia-built LiDAR maps in Fig. \ref{fig:lidar_maps}.
This situation arises since Avia adopts a data acquisition model inspired by \cite{zhu2022robust}, which is specifically tailored for small FoV LiDAR for stable LiDAR-only scan registration in spatiotemporal calibration, and is different from the data acquisition model of large FoV LiDAR, such as Velodyne.
The corresponding demonstration video can be viewed at the open-source repository for \emph{iKalibr}.
}.
As for Velodyne LiDARs, namely the VLP-16 in \emph{LI-Calib} dataset and the VLP-32C in our dataset, \emph{iKalibr} achieves comparable calibration results with \emph{OA-Calib}, and has a slight advantage in repeatability.

\subsubsection{Qualitative Evaluation via By-Products}
Based on estimated B-splines and gravity vector, calibrated spatiotemporal parameters, and raw LiDAR measurements, gravity-aligned LiDAR maps can be constructed, see Fig. \ref{fig:lidar_maps}.
It can be found that structural details of surroundings could be visible, and the color of objects coincides with corresponding elevations (the color gradient transitions from purple to yellow with increasing elevation).
These results demonstrate the superior consistency of \emph{iKaibr}.

\subsection{Multi-Camera Multi-IMU (\textbf{M-CI}) Calibration}
\label{sect:m-ci-calib}

\begin{table}[t]
\renewcommand{\arraystretch}{\tabheight}
\setlength{\tabcolsep}{\tabwidth}
\centering
\caption{\textbf{Spatiotemporal Calibration Results in M-CI Experiments}
\\iKalibr enables multi-IMU multi-camera spatiotemporal calibration with slightly lower repeatability than Kalibr
}
\label{tab:m-ci-calib}
\begin{threeparttable}
\begin{tabular}{ccc|rrc}
\toprule
\multicolumn{3}{c|}{Method}                                                                                                                                                                          & \multicolumn{1}{c}{Ours}   & \multicolumn{1}{c}{Kalibr \cite{furgale2013unified}} & Mean Diff. \\ \midrule\midrule
\multicolumn{1}{c|}{\multirow{40}{*}{\rotatebox{90}{Our Dataset}}}                                  & \multicolumn{1}{c|}{\multirow{12}{*}{\rotatebox{90}{IMU-1}}}      & $\hat{p}_{x}$               & 2.898$\pm$0.054            & 2.920$\pm$\textbf{0.003}                             & 0.022      \\
\multicolumn{1}{c|}{}                                                                               & \multicolumn{1}{c|}{}                                            & $\hat{p}_{y}$               & -12.286$\pm$\textbf{0.050} & -12.166$\pm$0.094                                    & 0.120      \\
\multicolumn{1}{c|}{}                                                                               & \multicolumn{1}{c|}{}                                            & $\hat{p}_{z}$               & -3.195$\pm$\textbf{0.042}  & -3.146$\pm$0.089                                     & 0.049      \\ \cmidrule{3-6} 
\multicolumn{1}{c|}{}                                                                               & \multicolumn{1}{c|}{}                                            & $\hat{\theta}_r$            & -179.797$\pm$0.021         & -179.820$\pm$\textbf{0.004}                          & 0.023      \\
\multicolumn{1}{c|}{}                                                                               & \multicolumn{1}{c|}{}                                            & $\hat{\theta}_p$            & -0.451$\pm$0.035           & -0.379$\pm$\textbf{0.008}                            & 0.072      \\
\multicolumn{1}{c|}{}                                                                               & \multicolumn{1}{c|}{}                                            & $\hat{\theta}_y$            & -87.601$\pm$0.148          & -87.515$\pm$\textbf{0.032}                           & 0.086      \\ \cmidrule{3-6} 
\multicolumn{1}{c|}{}                                                                               & \multicolumn{1}{c|}{}                                            & $\hat{\tau}$                & 1.032$\pm$0.351            & 0.750$\pm$\textbf{0.062}                             & 0.282      \\ \cmidrule{2-6} 
\multicolumn{1}{c|}{}                                                                               & \multicolumn{1}{c|}{\multirow{12}{*}{\rotatebox{90}{Cam-0}}}      & $\hat{p}_{x}$               & 7.495$\pm$0.207            & 7.941$\pm$\textbf{0.128}                             & 0.446      \\
\multicolumn{1}{c|}{}                                                                               & \multicolumn{1}{c|}{}                                            & $\hat{p}_{y}$               & -24.997$\pm$0.205          & -25.630$\pm$\textbf{0.016}                           & 0.633      \\
\multicolumn{1}{c|}{}                                                                               & \multicolumn{1}{c|}{}                                            & $\hat{p}_{z}$               & 11.956$\pm$0.165           & 12.219$\pm$\textbf{0.003}                            & 0.263      \\ \cmidrule{3-6} 
\multicolumn{1}{c|}{}                                                                               & \multicolumn{1}{c|}{}                                            & $\hat{\theta}_r$            & -93.227$\pm$0.043          & -93.245$\pm$\textbf{0.016}                           & 0.018      \\
\multicolumn{1}{c|}{}                                                                               & \multicolumn{1}{c|}{}                                            & $\hat{\theta}_p$            & -0.225$\pm$\textbf{0.007}  & -0.212$\pm$0.012                                     & 0.043      \\
\multicolumn{1}{c|}{}                                                                               & \multicolumn{1}{c|}{}                                            & $\hat{\theta}_y$            & -177.706$\pm$0.124         & -177.621$\pm$\textbf{0.065}                          & 0.085      \\ \cmidrule{3-6} 
\multicolumn{1}{c|}{}                                                                               & \multicolumn{1}{c|}{}                                            & $\hat{\tau}$                & -22.816$\pm$0.347          & -23.300$\pm$\textbf{0.141}                           & 0.484      \\ \cmidrule{2-6} 
\multicolumn{1}{c|}{}                                                                               & \multicolumn{1}{c|}{\multirow{12}{*}{\rotatebox{90}{Cam-1}}}      & $\hat{p}_{x}$               & -6.618$\pm$0.239           & -6.903$\pm$\textbf{0.051}                            & 0.285      \\
\multicolumn{1}{c|}{}                                                                               & \multicolumn{1}{c|}{}                                            & $\hat{p}_{y}$               & -25.501$\pm$0.176          & -26.467$\pm$\textbf{0.005}                           & 0.966      \\
\multicolumn{1}{c|}{}                                                                               & \multicolumn{1}{c|}{}                                            & $\hat{p}_{z}$               & 11.800$\pm$0.344           & 11.765$\pm$\textbf{0.075}                            & 0.035      \\ \cmidrule{3-6} 
\multicolumn{1}{c|}{}                                                                               & \multicolumn{1}{c|}{}                                            & $\hat{\theta}_r$            & 94.227$\pm$0.052           & 94.109$\pm$\textbf{0.001}                            & 0.118      \\
\multicolumn{1}{c|}{}                                                                               & \multicolumn{1}{c|}{}                                            & $\hat{\theta}_p$            & -0.490$\pm$0.012           & -0.519$\pm$\textbf{0.004}                            & 0.029      \\
\multicolumn{1}{c|}{}                                                                               & \multicolumn{1}{c|}{}                                            & $\hat{\theta}_y$            & 4.549$\pm$0.129            & 4.619$\pm$\textbf{0.045}                             & 0.070      \\ \cmidrule{3-6} 
\multicolumn{1}{c|}{}                                                                               & \multicolumn{1}{c|}{}                                            & $\hat{\tau}$                & -8.810$\pm$0.890           & -9.461$\pm$\textbf{0.134}                            & 0.651      \\ \midrule
\multicolumn{1}{c|}{\multirow{28}{*}{\rotatebox{90}{TUM GS-RS Dataset \cite{schubert2019rolling}}}} & \multicolumn{1}{c|}{\multirow{12}{*}{\rotatebox{90}{Cam-0 (GS)}}} & $\hat{p}_{x}$               & 0.324$\pm$0.300            & 0.780\hspace{4mm}                                    & 0.456      \\
\multicolumn{1}{c|}{}                                                                               & \multicolumn{1}{c|}{}                                            & $\hat{p}_{y}$               & 5.338$\pm$0.246            & 5.214\hspace{4mm}                                    & 0.123      \\
\multicolumn{1}{c|}{}                                                                               & \multicolumn{1}{c|}{}                                            & $\hat{p}_{z}$               & -4.394$\pm$0.611           & -4.284\hspace{4mm}                                   & 0.109      \\ \cmidrule{3-6} 
\multicolumn{1}{c|}{}                                                                               & \multicolumn{1}{c|}{}                                            & $\hat{\theta}_r$            & 179.630$\pm$0.088          & 179.645\hspace{4mm}                                  & 0.015      \\
\multicolumn{1}{c|}{}                                                                               & \multicolumn{1}{c|}{}                                            & $\hat{\theta}_p$            & -0.339$\pm$0.040           & -0.424\hspace{4mm}                                   & 0.085      \\
\multicolumn{1}{c|}{}                                                                               & \multicolumn{1}{c|}{}                                            & $\hat{\theta}_y$            & -90.334$\pm$0.012          & -90.341\hspace{4mm}                                  & 0.007      \\ \cmidrule{3-6} 
\multicolumn{1}{c|}{}                                                                               & \multicolumn{1}{c|}{}                                            & $\hat{\tau}$                & -0.113$\pm$0.107           & -0.059\hspace{4mm}                                   & 0.054      \\ \cmidrule{2-6} 
\multicolumn{1}{c|}{}                                                                               & \multicolumn{1}{c|}{\multirow{12}{*}{\rotatebox{90}{Cam-1 (RS)}}} & $\hat{p}_{x}$               & 0.464$\pm$0.254            & 0.699\hspace{4mm}                                    & 0.235      \\
\multicolumn{1}{c|}{}                                                                               & \multicolumn{1}{c|}{}                                            & $\hat{p}_{y}$               & -5.753$\pm$0.157           & -5.708\hspace{4mm}                                   & 0.044      \\
\multicolumn{1}{c|}{}                                                                               & \multicolumn{1}{c|}{}                                            & $\hat{p}_{z}$               & -4.228$\pm$0.533           & -4.227\hspace{4mm}                                   & 0.001      \\ \cmidrule{3-6} 
\multicolumn{1}{c|}{}                                                                               & \multicolumn{1}{c|}{}                                            & $\hat{\theta}_r$            & 179.430$\pm$0.077          & 179.370\hspace{4mm}                                  & 0.059      \\
\multicolumn{1}{c|}{}                                                                               & \multicolumn{1}{c|}{}                                            & $\hat{\theta}_p$            & -0.272$\pm$0.030           & -0.241\hspace{4mm}                                   & 0.030      \\
\multicolumn{1}{c|}{}                                                                               & \multicolumn{1}{c|}{}                                            & $\hat{\theta}_y$            & -90.162$\pm$0.011          & -90.158\hspace{4mm}                                  & 0.003      \\ \cmidrule{3-6} 
\multicolumn{1}{c|}{}                                                                               & \multicolumn{1}{c|}{}                                            & $\hat{\tau}$                & 0.386$\pm$0.389            & 0.197\hspace{4mm}                                    & 0.189      \\
\multicolumn{1}{c|}{}                                                                               & \multicolumn{1}{c|}{}                                            & $\hat{\tau}_{\mathrm{red}}$ & 30.554$\pm$0.116           & 30.177\hspace{4mm}                                   & 0.376      \\ \bottomrule
\end{tabular}

\begin{tablenotes} 
\item[*] Extrinsic translations in $(cm)$, extrinsic Euler angles in $(deg)$, and time offsets in $(ms)$. All parameters are with respect to IMU-0. \item[*] The extrinsic parameters and readout time calibration results of Kalibr for the TUM dataset are obtained from \cite{schubert2019rolling}. However, the time offsets for hardware-synchronized cameras in TUM dataset are not provided (calibration datasets are provided). Therefore, we performed time offset calibration for cameras using Kalibr for comparison.
\end{tablenotes}
\end{threeparttable}
\end{table}

\begin{figure}[t]
\centering
\includegraphics[width=\figscale\linewidth]{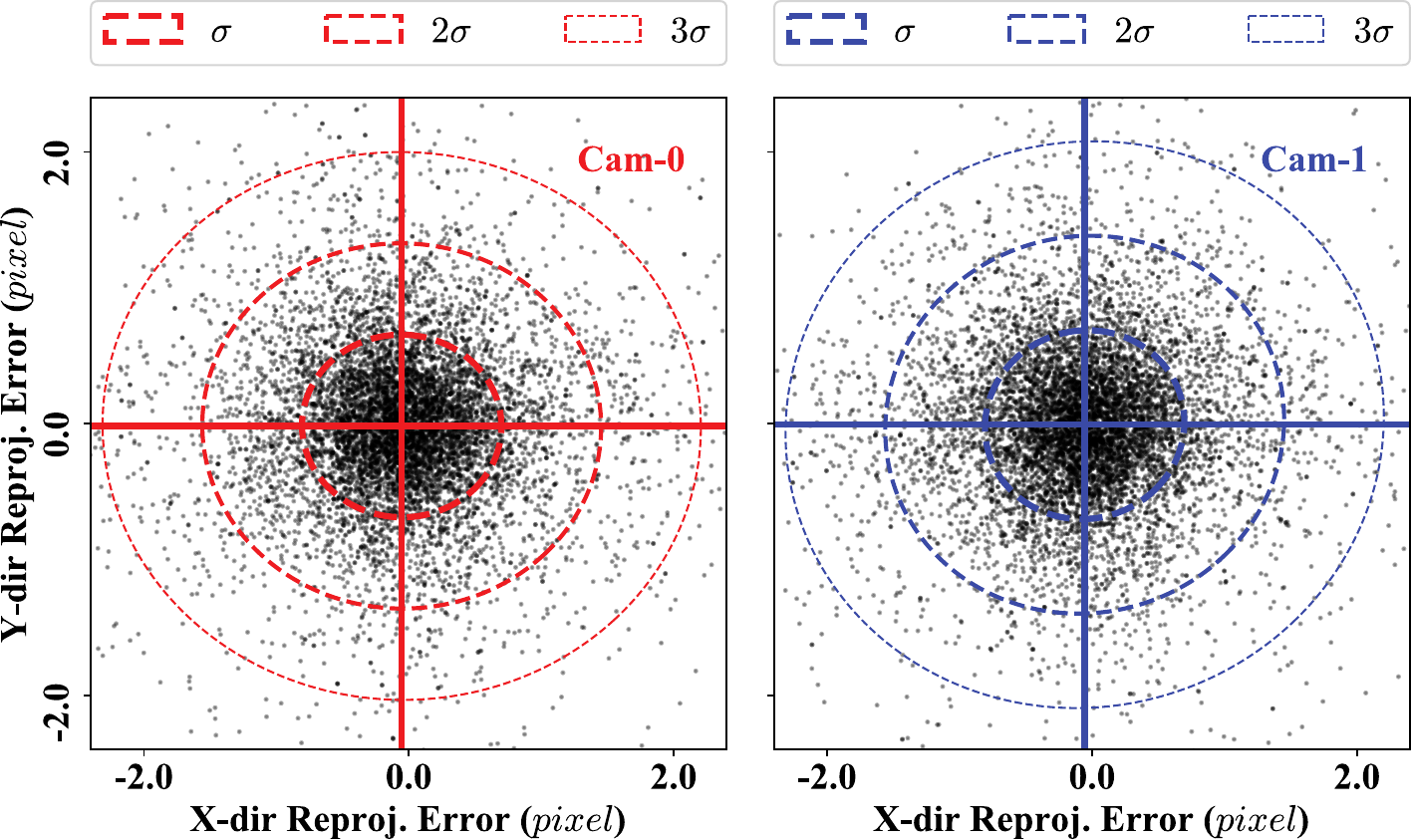}
\caption{The distribution of visual reprojection errors for multiple cameras. Solid lines represent means of reprojection errors.}
\label{fig:reproj_error}
\end{figure}

\begin{figure}[t]
\centering
\includegraphics[width=\figscale\linewidth]{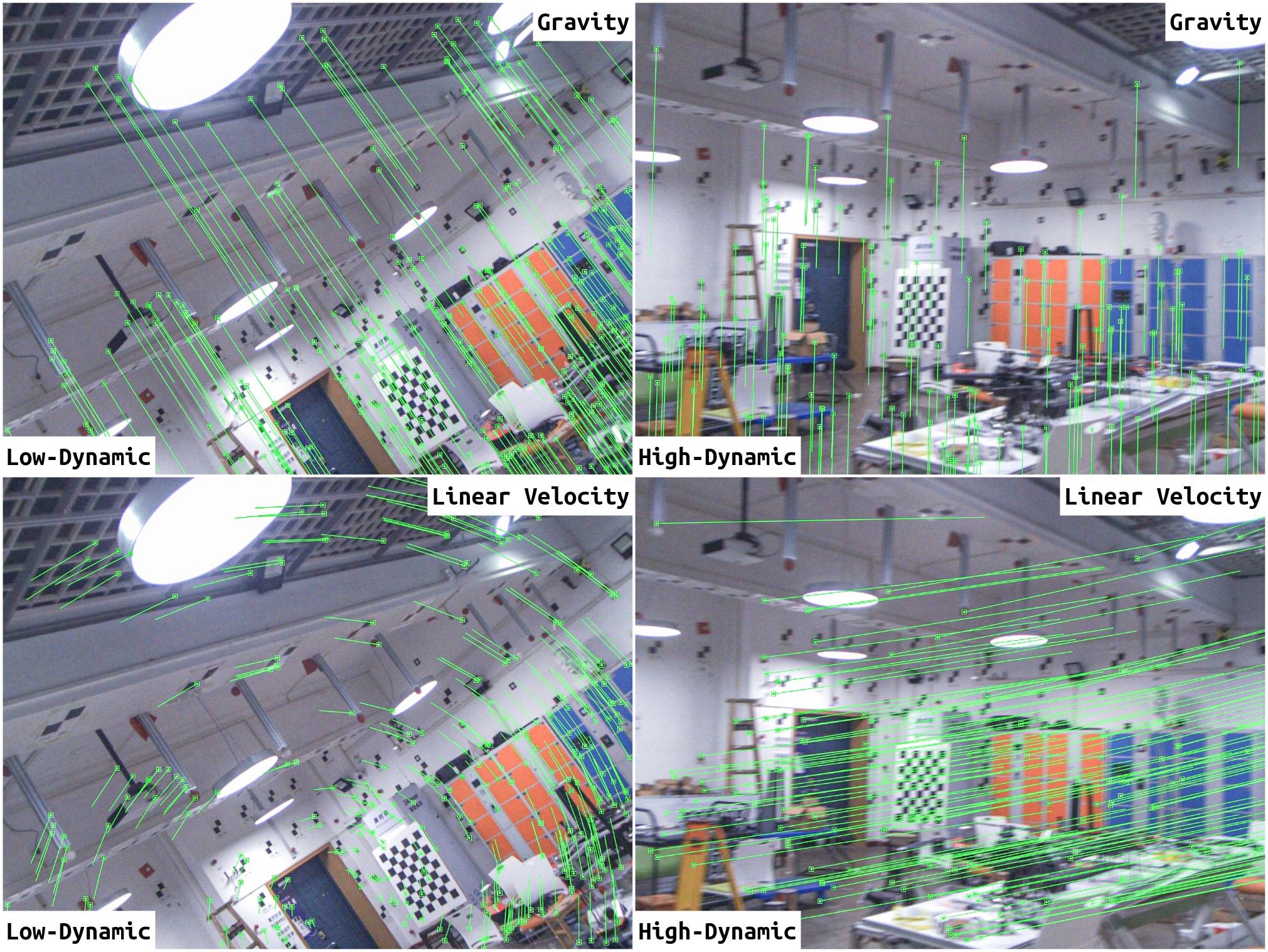}
\caption{The visual kinematics of high and low dynamic moments in our \emph{Laboratory} dataset. Kinematics coincides with the structural details and motion blur (see chessboard). Tails of features in images represent gravity vectors or linear velocities.}
\label{fig:visual_kinematics}
\end{figure}

\subsubsection{Quantitative Evaluation and Comparison}
In this experiment, the TUM GS-RS dataset \cite{schubert2019rolling} is involved in the evaluation, together with our dataset, mainly as a supplement to RS camera (the readout time) calibration evaluation.
To make a reliable comparison, the well-known chessboard-based visual-inertial spatiotemporal calibration toolkit, namely \emph{Kalibr} \cite{furgale2013unified}, is considered in the experiment.
Table \ref{tab:m-ci-calib} provides the calibration results, summarizing spatiotemporal estimates and corresponding STDs.
It can be found that the proposed \emph{iKalibr} can generate comparable calibration results with \emph{Kalibr}.
The estimate differences between \emph{Kalibr} and ours are about $1\;cm$ for translation, $0.1\;deg$ for rotation, and $0.8\;ms$ for time offset.
More specifically, \emph{Kalibr} is superior to \emph{iKalibr} in translation calibration (see STDs).
This is reasonable as the target-based \emph{Kalibr} could explicitly obtain accurate scale priori from the chessboard, while the target-free \emph{iKalibr} estimates visual scale using linear acceleration information from IMU.
In terms of readout time determination of RS camera (TUM GS-RS dataset), the difference between the reference and our estimated one is less than $0.5\;ms$, with a STD of $0.15\;ms$.

\subsubsection{Qualitative Evaluation via By-Products}
Based on the estimated B-splines, inverse depths of features, and feature matching correspondences, we computed the visual reprojection errors (as defined in (\ref{equ:visual_reproj_error})) for each camera (using the sequence 1 in TUM GS-RS dataset), see Fig. \ref{fig:reproj_error}. It can be found that errors obey zero-mean Gaussian distribution, and STDs for both GS and RS cameras are within $1\;pixel$.
We also projected the visual kinematics (gravity vectors and linear velocities of landmarks) to each image to validate estimation consistency, see Fig. \ref{fig:visual_kinematics}.
As can be seen, the direction of estimated gravity matches well with the structures of the surroundings.
The motion blur of images coincides with different dynamic motions (see the chessboard).

\subsection{One-Shot Multi-Sensor (\textbf{M-CLRI}) Calibration}
\label{sect:m-clri-calib}

\begin{figure}[t]
\centering
\includegraphics[width=\figscale\linewidth]{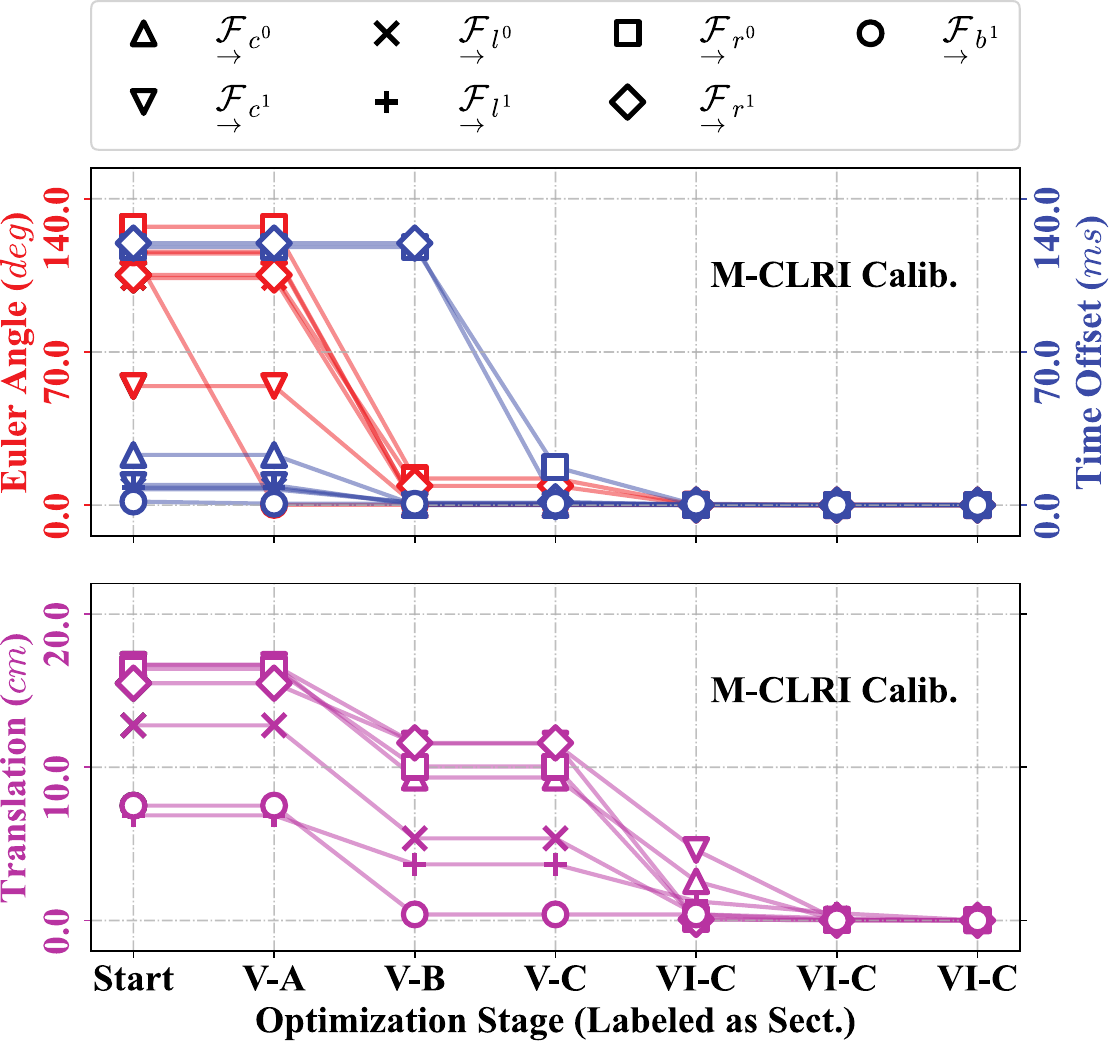}
\caption{The convergence performance of spatiotemporal optimization in \textbf{M-CLRI} calibration, where all parameters are with respect to \emph{IMU-0}. Quantities are the RMSEs of estimates with respect to the final ones. For better readers' understanding, each stage is labeled by its corresponding section index.} 
\label{fig:convergence}
\end{figure}

\begin{table*}[t]
\setlength{\tabcolsep}{\tabwidth}
\centering
\caption{\textbf{Spatiotemporal Calibration Differences Before and After Batch Optimization in M-CLRI Experiments}
\\In iKalibr's dynamic spatiotemporal initialization, radar shows significantly lower accuracy compared to the other three sensors, particularly for extrinsic translation initialization
}
\label{tab:m-clri-calib-init-final-diff}
\begin{threeparttable}

\begin{tabular}{r|rrr|rrr|r}
\toprule
\multirow{2.5}{*}{Sensor} & \multicolumn{3}{c|}{Extrinsic Translation Diff. ($cm$)}                                                                      & \multicolumn{3}{c|}{Extrinsic Rotation Diff. ($deg$)}                                                                                 & \multicolumn{1}{c}{Temporal Diff. ($ms$)} \\ \cmidrule{2-8} 
                        & \multicolumn{1}{c}{$\delta\hat{p}_{x}$} & \multicolumn{1}{c}{$\delta\hat{p}_{y}$} & \multicolumn{1}{c|}{$\delta\hat{p}_{z}$} & \multicolumn{1}{c}{$\delta\hat{\theta}_r$} & \multicolumn{1}{c}{$\delta\hat{\theta}_p$} & \multicolumn{1}{c|}{$\delta\hat{\theta}_y$} & \multicolumn{1}{c}{$\delta\hat{\tau}$}    \\ \midrule\midrule
IMU-1                   & 0.083$\pm$0.131                         & 0.230$\pm$0.250                         & -0.442$\pm$0.080                         & -0.010$\pm$0.021                           & 0.051$\pm$0.020                            & 0.180$\pm$0.025                             & -0.246$\pm$0.231                          \\ \midrule
Rad-0                   & 4.255$\pm$4.722                         & 12.220$\pm$3.251                        & 4.945$\pm$4.350                          & -5.975$\pm$7.209                           & -2.413$\pm$5.814                           & 1.711$\pm$7.472                             & 26.413$\pm$20.926                         \\
Rad-1                   & -8.249$\pm$2.998                        & 10.626$\pm$4.941                        & 5.504$\pm$4.121                          & -7.640$\pm$6.790                           & -3.679$\pm$8.658                           & -1.640$\pm$5.201                            & 27.160$\pm$23.872                         \\ \midrule
LiDAR-0                 & -4.870$\pm$1.324                        & -1.492$\pm$2.064                        & 4.982$\pm$1.301                          & -0.076$\pm$0.474                           & -0.128$\pm$0.185                           & 0.468$\pm$0.579                             & -1.628$\pm$0.607                          \\
LiDAR-1                 & -4.658$\pm$3.167                        & 8.467$\pm$3.603                         & 1.775$\pm$2.542                          & -0.058$\pm$0.223                           & -0.288$\pm$0.497                           & -0.073$\pm$0.176                            & -1.068$\pm$0.449                          \\ \midrule
Cam-0                   & -6.391$\pm$3.358                        & 3.072$\pm$2.525                         & -2.073$\pm$1.930                         & -0.079$\pm$0.028                           & 0.031$\pm$0.022                            & -0.086$\pm$0.034                            & -0.021$\pm$0.058                          \\
Cam-1                   & -6.878$\pm$2.891                        & 4.552$\pm$2.581                         & -0.947$\pm$2.296                         & -0.177$\pm$0.048                           & -0.020$\pm$0.031                           & -0.200$\pm$0.046                            & 0.075$\pm$0.105                           \\ \bottomrule
\end{tabular}

\begin{tablenotes} 
\item[*] Extrinsic translations in $(cm)$, extrinsic Euler angles in $(deg)$, and time offsets in $(ms)$.
\item[*] The values represent \textbf{differences} between spatiotemporal calibration results obtained from \textbf{initialization} and those after \textbf{batch optimizations}.
\end{tablenotes}
\end{threeparttable}
\end{table*}

\begin{table*}[t]
\setlength{\tabcolsep}{\tabwidth}
\centering
\caption{\textbf{Spatiotemporal Calibration Results in M-CLRI Experiments}
\\iKalibr enables one-shot resilient calibration, with lower forward extrinsic translation accuracy for Livox LiDAR
}
\label{tab:m-clri-calib}
\begin{threeparttable}

\begin{tabular}{r|rrr|rrr|r}
\toprule
\multirow{2.5}{*}{Sensor} & \multicolumn{3}{c|}{Extrinsic Translation ($cm$)}                                                          & \multicolumn{3}{c|}{Extrinsic Rotation ($deg$)}                                                                     & \multicolumn{1}{c}{Temporal Param. ($ms$)} \\ \cmidrule{2-8} 
                        & \multicolumn{1}{c}{$\hat{p}_{x}$} & \multicolumn{1}{c}{$\hat{p}_{y}$} & \multicolumn{1}{c|}{$\hat{p}_{z}$} & \multicolumn{1}{c}{$\hat{\theta}_r$} & \multicolumn{1}{c}{$\hat{\theta}_p$} & \multicolumn{1}{c|}{$\hat{\theta}_y$} & \multicolumn{1}{c}{$\hat{\tau}$}           \\ \midrule\midrule
IMU-1                   & 2.908$\pm$0.061                   & -12.297$\pm$0.049                 & -3.185$\pm$0.038                   & -179.793$\pm$0.014                   & -0.462$\pm$0.039                     & -87.599$\pm$0.151                     & 1.128$\pm$0.270                            \\ \midrule
Rad-0                   & -15.016$\pm$1.587                 & -22.891$\pm$0.810                 & 6.524$\pm$0.709                    & 174.333$\pm$1.120                    & 8.643$\pm$1.134                      & -133.398$\pm$0.280                    & -118.798$\pm$1.176                         \\
Rad-1                   & 15.376$\pm$1.643                  & -20.846$\pm$0.866                 & -0.057$\pm$0.429                   & -176.786$\pm$1.499                   & 4.397$\pm$0.607                      & -41.529$\pm$0.392                     & -119.171$\pm$1.651                         \\ \midrule
LiDAR-0                 & -0.571$\pm$0.174                  & -4.925$\pm$0.093                  & -21.693$\pm$0.579                  & -179.787$\pm$0.040                   & 0.105$\pm$0.027                      & 2.093$\pm$0.141                       & 7.193$\pm$0.034                            \\
LiDAR-1                 & 0.378$\pm$0.303                   & \textcolor{red}{\bf{-13.753$\pm$1.085}}                 & -0.858$\pm$0.300                   & 178.994$\pm$0.011                    & 0.120$\pm$0.026                      & -87.273$\pm$0.141                     & 7.844$\pm$0.065                            \\ \midrule
Cam-0                   & 7.386$\pm$0.214                   & -25.287$\pm$0.274                 & 11.828$\pm$0.285                   & -93.182$\pm$0.046                    & -0.221$\pm$0.008                     & -177.706$\pm$0.138                    & -22.787$\pm$0.327                          \\
Cam-1                   & -6.640$\pm$0.182                  & -25.902$\pm$0.284                 & 11.732$\pm$0.416                   & 94.188$\pm$0.054                     & -0.497$\pm$0.012                     & 4.556$\pm$0.139                       & -8.784$\pm$0.850                           \\ \bottomrule
\end{tabular}

\begin{tablenotes} 
\item[*] Extrinsic translations in $(cm)$, extrinsic Euler angles in $(deg)$, and time offsets in $(ms)$. All parameters are with respect to IMU-0.
\end{tablenotes}
\end{threeparttable}
\end{table*}

\subsubsection{Quantitative Evaluation and Convergence Performance}
While the previous experiments evaluated \emph{iKalibr} on four kinds of minimum primitives of multi-sensor inertial integration, the most notable feature of \emph{iKalibr} is supporting one-shot multi-sensor spatiotemporal determination, as long as the sensor kit satisfies (\ref{equ:support_kit_type}).
Table \ref{tab:m-clri-calib-init-final-diff} provides differences between spatiotemporal parameters obtained from initialization and those after batch optimizations.
It can be seen that, among the various sensors, the radar's initialization results are relatively poorer compared to the other sensors, which holds for the extrinsics and time delays to be calibrated.
The IMU's initialization results are relatively accurate, aligning most closely with the final batch optimization results.
Table \ref{tab:m-clri-calib} provides the final calibration results, including the estimates and their STDs.
It can be found that among four kinds of sensors, the IMUs achieve the highest repeatability, followed by cameras and LiDARs, and radars have the lowest repeatability.
This is largely related to the measuring noise levels of the sensors themselves.
The convergence performance of spatiotemporal parameters in one-shot calibration is shown in Fig. \ref{fig:convergence}.
Due to the designed initialization procedure and multi-stage refinement, spatiotemporal parameters can steadily converge to final states.

\subsubsection{Qualitative Evaluation via By-Products}
\begin{figure}[t]
\centering
\includegraphics[width=\linewidth]{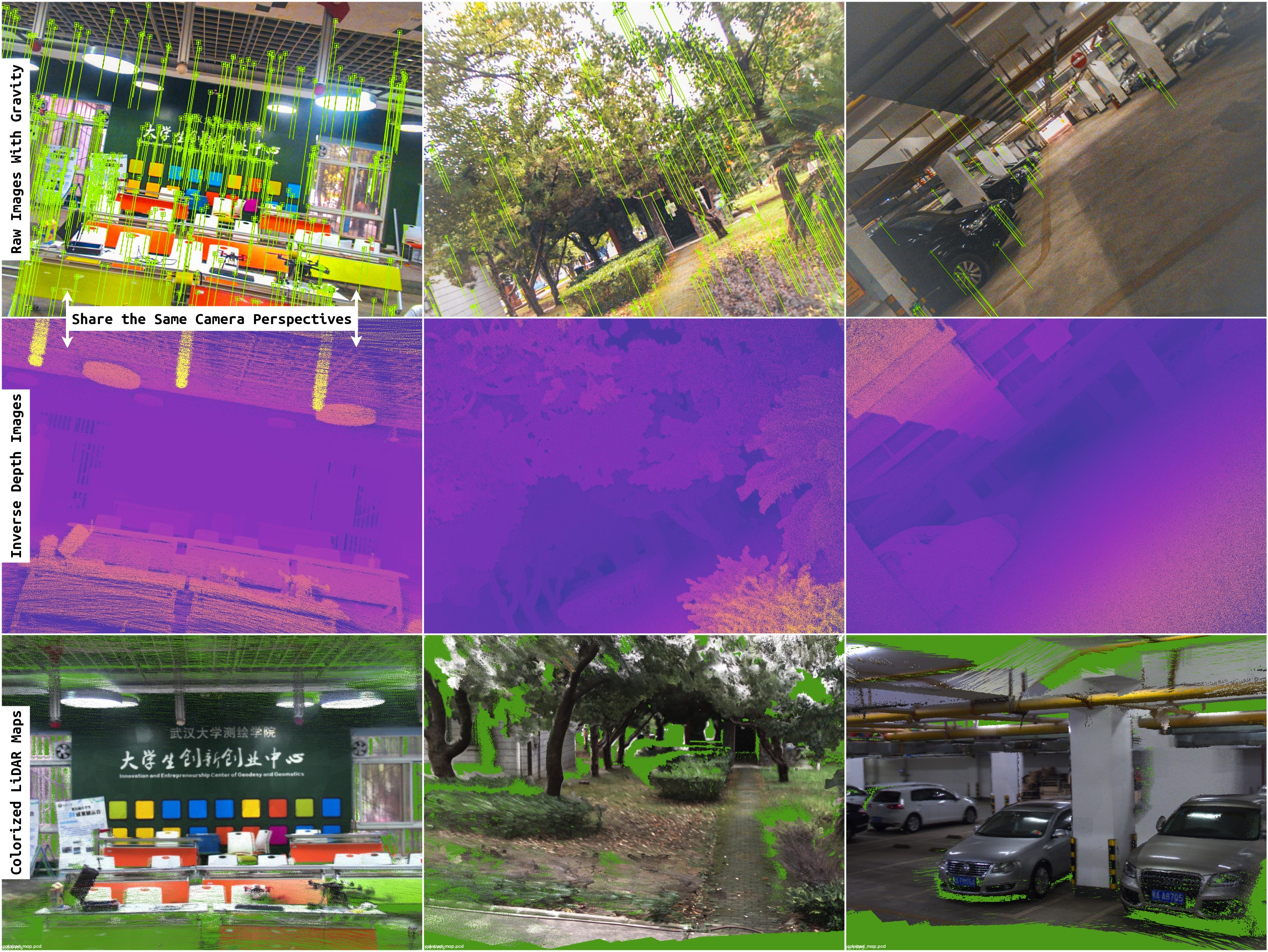}
\caption{Raw images (top row) and their covisible inverse depth images (middle row) sharing the same camera perspectives. Inverse Depth images are created by projecting LiDAR points to image planes based on camera intrinsics, the estimated spatiotemporal parameters, and global LiDAR map.
The bottom row of images demonstrates the colorized LiDAR point cloud maps, where green indicates LiDAR blind area.}
\label{fig:inverse_depth}
\end{figure}

When camera and LiDAR are involved in calibration, covisible inverse depth images can be created using the LiDAR map, B-splines, and camera intrinsics, see Fig. \ref{fig:inverse_depth}, for a qualitative calibration consistency evaluation.
It’s clear that the depth information matches well with the raw images, and the colors and structures within colorized LiDAR maps correspond well,  demonstrating the consistency of the proposed \emph{iKalibr}.

\subsection{Statistics of Computation Time Consumption}
\begin{table}[t]
\centering
\caption{\textbf{Computation Consumption Statistics in Experiments}
\\iKalibr has considerable room for improvement in camera-involved calibration, particularly in the SfM module
}
\label{tab:comp_consumption}
\begin{threeparttable}

\begin{tabular}{l|clcc}
\toprule
\multicolumn{1}{c|}{\multirow{4}{*}{\begin{tabular}[c]{@{}c@{}}Hardware\\ Configuration\end{tabular}}}     & OS Name                                    & \multicolumn{3}{l}{Ubuntu 20.04.6 LTS 64-Bit}                                     \\ \cmidrule{2-5} 
\multicolumn{1}{c|}{}                                                                                      & Processor                                  & \multicolumn{3}{l}{12th Gen Intel® Core™ i9}                                      \\ \cmidrule{2-5} 
\multicolumn{1}{c|}{}                                                                                      & Graphics                                   & \multicolumn{3}{l}{Mesa Intel® Graphics}                                          \\ \midrule\midrule
\multicolumn{1}{c|}{\multirow{2.5}{*}{\begin{tabular}[c]{@{}c@{}}Sensor Suite\\ Configuration\end{tabular}}} & \multicolumn{4}{c}{Computation Consumption (\textbf{unit: minute})}                                                            \\ \cmidrule{2-5} 
\multicolumn{1}{c|}{}                                                                                      & \multicolumn{2}{c}{Init. (\ref{sect:init}) + \bf{SfM}\cite{schoenberger2016sfm}} & BOs (\ref{sect:batch_opt})     & Total      \\ \midrule
M-I\hspace{15pt} (\ref{sect:m-i-calib})                                                                                 & \multicolumn{2}{c}{00.031 + 00.000}                                              & 00.041                         & 00.072     \\
M-RI\hspace{10pt} (\ref{sect:m-ri-calib})                                                                               & \multicolumn{2}{c}{00.079 + 00.000}                                              & 00.205                         & 00.284     \\
M-LI\hspace{10pt} (\ref{sect:m-li-calib})                                                                               & \multicolumn{2}{c}{03.002 + 00.000}                                              & 03.323                         & 06.325     \\
M-CI\hspace{10pt} (\ref{sect:m-ci-calib})                                                                               & \multicolumn{2}{c}{00.209 + \bf{28.305}}                                         & 00.818                         & 29.332     \\
M-CLRI (\ref{sect:m-clri-calib})                                                                           & \multicolumn{2}{c}{03.522 + \bf{28.305}}                                         & 04.655                         & 36.482     \\ \bottomrule
\end{tabular}

\begin{tablenotes} 
\item[*] Time consumed by iKalibr in the initialization (see Section \ref{sect:init}) and batch optimizations (BOs, see Section \ref{sect:batch_opt}) are separately recorded.
\end{tablenotes}
\end{threeparttable}
\vspace{-10pt}
\end{table}
Table \ref{tab:comp_consumption} presents the average computation time required by \emph{iKalibr} for five sensor configurations targeted in the real-world experiments.
The time consumption represents the average calculated over several 40-second sequences in our self-collected dataset.
In the experiments, the time distances for the rotation B-spline and linear scale B-spline were set to 0.05 seconds.
In the optimization process, 10 threads were used for parallel execution.
It can be observed that calibration for systems integrating only IMU or radar requires less computation time and is more efficient. In contrast, calibration for systems integrating LiDAR, particularly cameras, demands significantly more computation time. Specifically, when cameras are involved in the calibration, performing time-intensive SfM using COLMAP introduces an additional computational burden, which is the issue we are currently addressing, with promising progress toward an elegant solution, see \cite{chen2024ikalibr}.

\section{Conclusion and Future Work}
\label{sect:conclusion}
In this work, we propose a unified target-free spatiotemporal calibration framework orienting to resilient integrated inertial systems, termed as \emph{iKalibr}, which supports both spatial and temporal determination, enables one-shot resilient and compact calibration, and requires no additional artificial target or aiding sensor.
Considering the high nonlinearity of continuous-time-based batch spatiotemporal optimization, a rigorous initialization is first conducted to recover accurate initial guesses of all parameters in estimator. Following that, a continuous-time-based graph optimization is organized based on raw measurements, data correspondences, and initials of states.
The graph optimization would be performed several times until the final convergence, to steadily refine all states to better ones.
Sufficient real-world experiments are performed, and the results show that \emph{iKalibr} is capable of accurate and consistent resilient spatiotemporal calibration.
The IMU, radar, LiDAR, and optical camera are supported currently in \emph{iKalibr}.

It is also worth noting that \emph{iKalibr} currently has its limitations.
First, due to the lack of spatiotemporal ground truth in experiments, the evaluation of calibration results may be somewhat coarse, leaving room for potential improvement. Second, as a motion-based spatiotemporal calibrator, \emph{iKalibr} requires six-DoF motion excitation to ensure spatiotemporal observability, which may be challenging to implement for certain ground vehicles or large, heavy platforms. The issue of poor forward calibration for small FoV LiDAR also requires improvement. Finally, \emph{iKalibr} has so far been validated with a limited set of sensor types, and adaptations for other types of sensors are still required.
In future work, we will focus on addressing the aforementioned limitations while also improving computational efficiency and extending \emph{iKalibr} to support more types of sensors, such as event cameras.

\section*{Appendix A}
\section*{Stationary Inertial Intrinsic Calibration}
\label{sect:app_inertial_intri_calib}
When performing IMU-only multi-IMU spatiotemporal calibration in \emph{iKalibr}, the intrinsics of the IMU, namely $\bsm{x}_{\mathrm{in}}^b$, are with weak observability, which could lead to the problem rank deficient and further adversely affects the spatiotemporal determination.
Considering this, the intrinsics are expected to be pre-calibrated using a separate process, and would be set to constants and not optimized in spatiotemporal optimization.
Specifically, given the inertial measurement of $\coordframe{b}$, we can associate them with the world-frame angular velocity and linear acceleration as:
\begin{equation}
\small
\begin{aligned}
\bsm{a}(\tau)&=\left(\rotation{b}{w}(\tau) \right) ^\top\cdot\left(\linacce{b}{w}(\tau)-\gravity{w}\right) 
\\
\bsm{\omega}(\tau)&=\left(\rotation{b}{w}(\tau) \right) ^\top\cdot\angvel{b}{w}(\tau)
\end{aligned}.
\end{equation}
When the body is stationary, we have the following approximation:
\begin{equation}
\small
\label{equ:stationary}
\rotation{b}{w}(\tau)\equiv\bsm{I}_{3\times 3},\;
\linacce{b}{w}(\tau)\equiv\bsm{0}_{3\times 1},\;
\angvel{b}{w}(\tau)\equiv\bsm{0}_{3\times 1}.
\end{equation}
Although the body is not strictly stationary with respect to the inertial space and would lead to trace angular velocity and linear acceleration due to the rotation of the earth, (\ref{equ:stationary}) still holds for MEMS IMUs this work focuses on, since their high noise level would drown out such perception.
Based on the stationary inertial measurements collected under several poses (generally six symmetric poses), the following constraint would be constructed for each stationary data piece for intrinsic determination:
\begin{equation}
\small
\label{equ:intri_calib}
\hat{\bsm{x}}_{\mathrm{in}}^b
=\arg\min
\sum_{i}^{\mathcal{S}_{\mathrm{sta}}}\sum_{n}^{\mathcal{D}_{i}}
\left( \left\| 
r_{\omega}\left( \tilde{\bsm{\omega}}_{i,n}\right) 
\right\| ^2_{\bsm{Q}_{i,n}}
+
\left\| 
r_{a}\left( \tilde{\bsm{a}}_{i,n}\right) 
\right\| ^2_{\bsm{Q}_{i,n}}\right) 
\end{equation}
with
\begin{equation}
\small
\begin{aligned}
r_{\omega}\left( \tilde{\bsm{\omega}}_{i,n}\right) &\triangleq
h_\omega\left( \bsm{0}_{3\times 1},\hat{\bsm{x}}_{\mathrm{in}}^b\right) 
-\tilde{\bsm{\omega}}_{i,n}
\\
r_{a}\left( \tilde{\bsm{a}}_{i,n}\right) &\triangleq
h_a\left( -\hat{\bsm{g}}^{w_i},\hat{\bsm{x}}_{\mathrm{in}}^b\right) 
-\tilde{\bsm{a}}_{i,n}
\end{aligned}
\end{equation}
where $\mathcal{D}_{i}$ denotes the $i$-th data piece in the stationary data sequence $\mathcal{S}_{\mathrm{sta}}$; $\bsm{g}^{w_i}$ is the world-frame gravity vector of the $i$-th data piece, which would also be estimated in this problem.
Note that while all intrinsics of accelerometer can be determined by (\ref{equ:intri_calib}), only the bias of intrinsics for gyroscope can be calibrated in this problem. Other intrinsic parameters of gyroscope, such as scale and non-orthogonal factors, are without observability.
In fact, these unobservable factors are generally calibrated by IMU providers and have been compensated in inertial outputs.

\section*{Appendix B}
\section*{Sensor-Inertial Alignment Constraints}
\label{sect:app_alignment}

The inertial measurement of $\coordframe{b}$, i.e., body-frame angular velocity and specific force, can be associated with the B-spline-derived world-frame angular velocity and linear acceleration, which is described as follows:
\begin{equation}
\small
\label{equ:single_imu_kinematics}
\begin{aligned}
\bsm{a}(\tau)&=\left(\rotation{b}{w}(\tau) \right) ^\top\cdot\left(\linacce{b}{w}(\tau)-\gravity{w}\right) 
\\
\bsm{\omega}(\tau)&=\left(\rotation{b}{w}(\tau) \right) ^\top\cdot\angvel{b}{w}(\tau)
\end{aligned}
\end{equation}
where $\rotation{b}{w}(\tau)$, $\angvel{b}{w}(\tau)$, and $\linacce{b}{w}(\tau)$ are kinematics from the B-splines.
By introducing inertial extrinsics, (\ref{equ:single_imu_kinematics}) can be extended to multiple IMUs:
\begin{equation}
\small
\label{equ:multi_imu_kinematics}
\begin{aligned}
\bsm{a}^i(\tau)&=\left(\rotation{b^i}{w}(\tau) \right) ^\top\cdot\left(\linacce{b^i}{w}(\tau)-\gravity{w}\right) 
\\
\bsm{\omega}^i(\tau)&=\left(\rotation{b^i}{w}(\tau) \right) ^\top\cdot\angvel{b^i}{w}(\tau)
\end{aligned}
\end{equation}
with
\begin{equation}
\small
\begin{aligned}
\rotation{b^i}{w}(\tau)&=\rotation{b^r}{w}(\tau)\cdot\rotation{b^i}{b^r},
\qquad
\angvel{b^i}{w}(\tau)=\angvel{b^r}{w}(\tau),
\\
\linacce{b^i}{w}(\tau)&=
\linacce{b^r}{w}(\tau)
+\left( \liehat{\angacce{b^r}{w}(\tau)}+\liehat{\angvel{b^r}{w}(\tau)}^2
\right) 
\cdot\rotation{b^r}{w}(\tau)\cdot\translation{b^i}{b^r}
\end{aligned}
\end{equation}
where $\rotation{b^r}{w}(\tau)$, $\angvel{b^r}{w}(\tau)$, $\angacce{b^r}{w}(\tau)$, and $\linacce{b^r}{w}(\tau)$ are kinematics from the B-splines of the reference IMU $\coordframe{b^r}$.
By performing time integration on (\ref{equ:multi_imu_kinematics}), the linear velocity variation in timepiece $[\tau_n,\tau_{n\smallplus 1})$ can be obtained as:
\begin{equation}
\small
\label{equ:velocity_variation}
\begin{aligned}
\linvel{b^r}{w}(\tau_{n\smallplus 1})-\linvel{b^r}{w}(\tau_n)
=\bsm{c}^i_{n,n\smallplus 1}-\bsm{A}^i_{n,n\smallplus 1}
\cdot\translation{b^i}{b^r}+\gravity{w}\cdot\Delta\tau_{n,n\smallplus 1}
\end{aligned}
\end{equation}
with
\begin{equation}
\small
\begin{aligned}
\bsm{c}^i_{n,n\smallplus 1}&\triangleq
\int_{\tau_n}^{\tau_{n\smallplus 1}}\rotation{b^r}{w}(t)\cdot\rotation{b^i}{b^r} \cdot\bsm{a}^{i}(t) \cdot \mathrm{d}t
\\
\bsm{A}^i_{n,n\smallplus 1}&\triangleq
\int_{\tau_n}^{\tau_{n\smallplus 1}}\left(\liehat{\angacce{b^r}{w}(t)}+ \liehat{\angvel{b^r}{w}(t)}^2\right) \cdot\rotation{b^r}{w}(t)\cdot \mathrm{d}t
\end{aligned}
\end{equation}
where $\Delta\tau_{n,n\smallplus 1}=\tau_{n\smallplus 1}- \tau_{n}$.
Continuing to perform time integration on (\ref{equ:velocity_variation}), the position variation in timepiece $[\tau_n,\tau_{n\smallplus 1})$ can be obtained as:
\begin{equation}
\small
\label{equ:position_variation}
\begin{aligned}
\translation{b^r}{w}(\tau_{n\smallplus 1})-\translation{b^r}{w}(\tau_n)
&=
\bsm{d}^i_{n,n\smallplus 1}-\bsm{B}^i_{n,n\smallplus 1}
\cdot\translation{b^i}{b^r}
\\
+&\linvel{b^r}{w}(\tau_n)\cdot\Delta\tau_{n,n\smallplus 1}+
\frac{1}{2}\cdot\gravity{w}\cdot\Delta^2\tau_{n,n\smallplus 1}
\end{aligned}
\end{equation}
with
\begin{equation}
\small
\begin{aligned}
\bsm{d}^i_{n,n\smallplus 1}&\triangleq
\iint_{\tau_n}^{\tau_{n\smallplus 1}}\rotation{b^r}{w}(t)\cdot\rotation{b^i}{b^r} \cdot\bsm{a}^{i}(t) \cdot \mathrm{d}t^2
\\
\bsm{B}^i_{n,n\smallplus 1}&\triangleq
\iint_{\tau_n}^{\tau_{n\smallplus 1}}\left(\liehat{\angacce{b^r}{w}(t)}+ \liehat{\angvel{b^r}{w}(t)}^2\right) \cdot\rotation{b^r}{w}(t)\cdot \mathrm{d}t^2
\end{aligned}.
\end{equation}
Note that the right parts in both (\ref{equ:velocity_variation}) and (\ref{equ:position_variation}) could be computed independently for each IMU based on the fitted rotation B-spline, inertial extrinsic rotations and time offsets, and raw specific force measurements in timepiece $[\tau_n,\tau_{n\smallplus 1})$.
Integration items, namely $\bsm{c}^i_{n,n\smallplus 1}$, $\bsm{A}^i_{n,n\smallplus 1}$, $\bsm{d}^i_{n,n\smallplus 1}$, and $\bsm{B}^i_{n,n\smallplus 1}$, can be obtained by numerical integration methods, such as midpoint rule, trapezoidal rule, or Simpson’s rule \cite{suli2003introduction}.

By performing an equivalent transformation on the variation items (left parts) in (\ref{equ:velocity_variation}) and (\ref{equ:position_variation}), additional sensors can be involved to construct sensor-inertial alignment constraints.
For example, regarding the radar, $\linvel{b^r}{w}(\cdot)$ can be organized based on radar-derived radar-frame linear velocities and extrinsics between the radar and the reference IMU.
In terms of the LiDAR and camera, $\translation{b^r}{w}(\cdot)$ can be organized based on odometry-derived positions and extrinsics.
As for the IMU, $\linvel{b^r}{w}(\cdot)$ in (\ref{equ:velocity_variation}) would be treated as estimated quantities explicitly.
For more details, see Section \ref{sect:init_sen_inertial_align}.

\section*{Acknowledgments}
The calibration data acquisition is performed on the GREAT (GNSS+ REsearch, Application and Teaching) software, developed by the GREAT Group, School of Geodesy and Geomatics (SGG), Wuhan University.

\section*{CRediT Authorship Contribution Statement}
\textbf{Shuolong Chen}: Conceptualisation, Methodology, Software, Validation, Original Draft, Revision.
\textbf{Xingxing Li}: Supervision, Funding Acquisition, Review and Editing.
\textbf{Shengyu Li} and \textbf{Yuxuan Zhou}: Review and Editing.
\textbf{Xiaoteng Yang}: Data Curation.
		
\bibliographystyle{IEEEtran}
\bibliography{reference}
\newcommand{\vspacebio}{\vspace{-4cm}}
\vspacebio
\begin{IEEEbiography}[{\includegraphics[width=1in,height=1.25in,clip,keepaspectratio,cframe={black!8!white}]{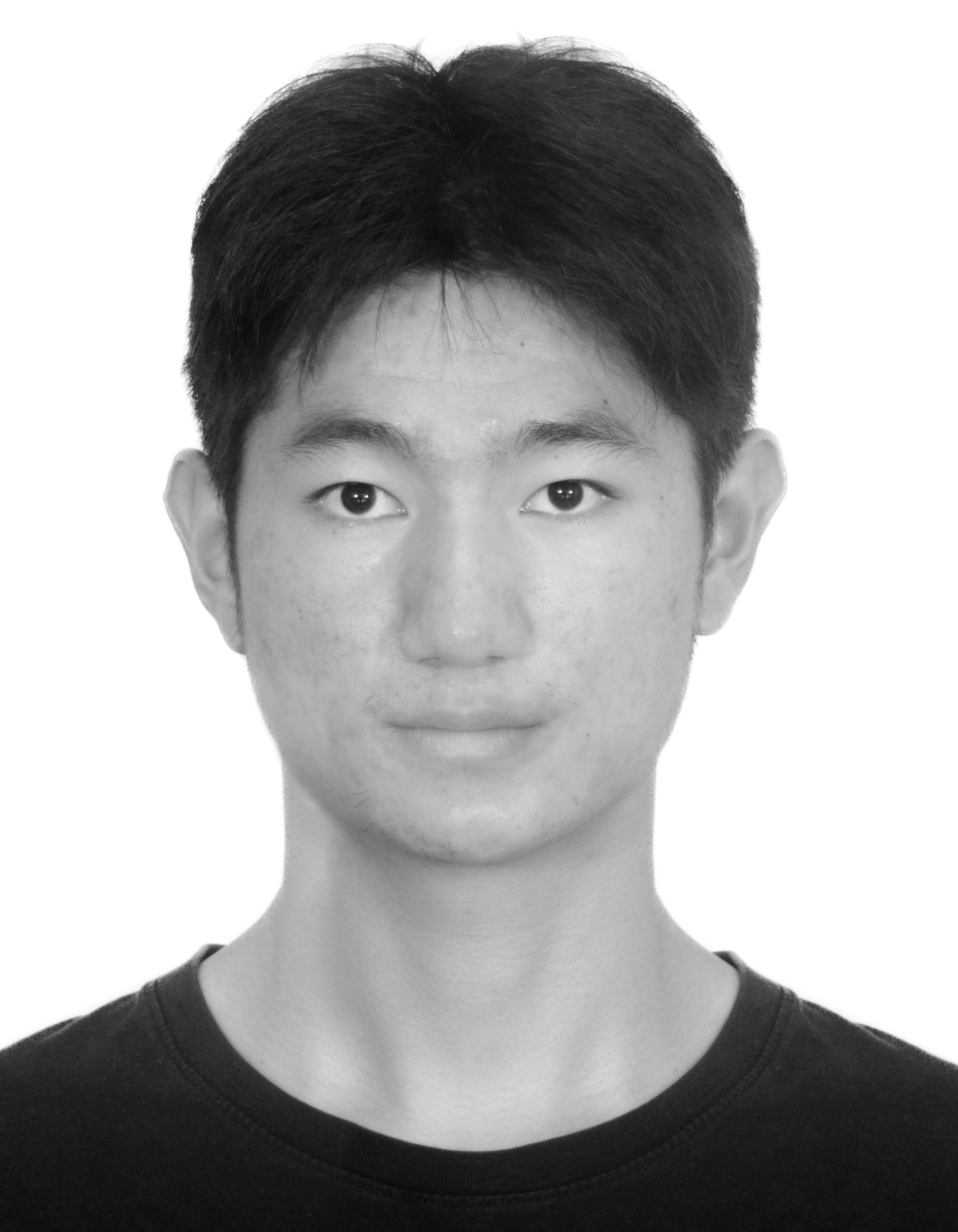}}]{Shuolong Chen}
	received the B.S. degree in geodesy and geomatics engineering from Wuhan University, Wuhan China, in 2023.
	
	He is currently a master candidate at the School of Geodesy and Geomatics (SGG), Wuhan University. His area of research currently focuses on integrated navigation systems and multi-sensor fusion, mainly on spatiotemporal calibration.
	
	Contact him via	e-mail: \emph{shlchen@whu.edu.cn}
\end{IEEEbiography}
\vspacebio
\begin{IEEEbiography}[{\includegraphics[width=1in,height=1.25in,clip,keepaspectratio,cframe={black!8!white}]{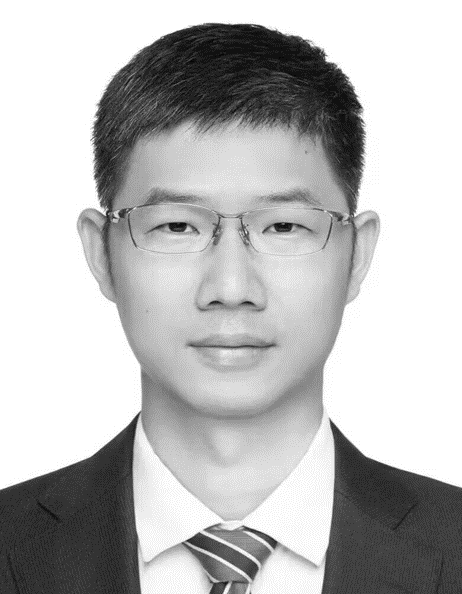}}]{Xingxing Li}
	received the B.S. degree from Wuhan University, Wuhan China, in 2008, and the Ph.D. degree from the Department of Geodesy and Remote Sensing, German Research Centre for Geosciences (GFZ), Potsdam, Germany, in 2015.
	
	He is currently a professor at the Wuhan University. His current research mainly involves GNSS precise data processing and multi-sensor fusion.
	
	Contact him via	e-mail: \emph{xxli@sgg.whu.edu.cn}
\end{IEEEbiography}
\vspacebio
\begin{IEEEbiography}[{\includegraphics[width=1in,height=1.25in,clip,keepaspectratio,cframe={black!8!white}]{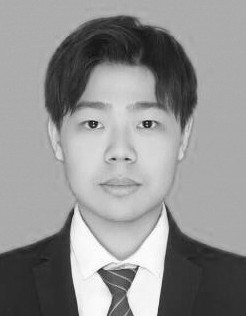}}]{Shengyu Li}
	received the M.S. degree in	geodesy and survey engineering from Wuhan University, Wuhan, China, in 2022.
	
	He is currently a doctor candidate at the School of Geodesy and Geomatics (SGG), Wuhan University, China. His current research focuses on multi-sensor fusion and integrated navigation system.
	
	Contact him via	e-mail: \emph{lishengyu@whu.edu.cn}
\end{IEEEbiography}
\vspacebio
\begin{IEEEbiography}[{\includegraphics[width=1in,height=1.25in,clip,keepaspectratio,cframe={black!8!white}]{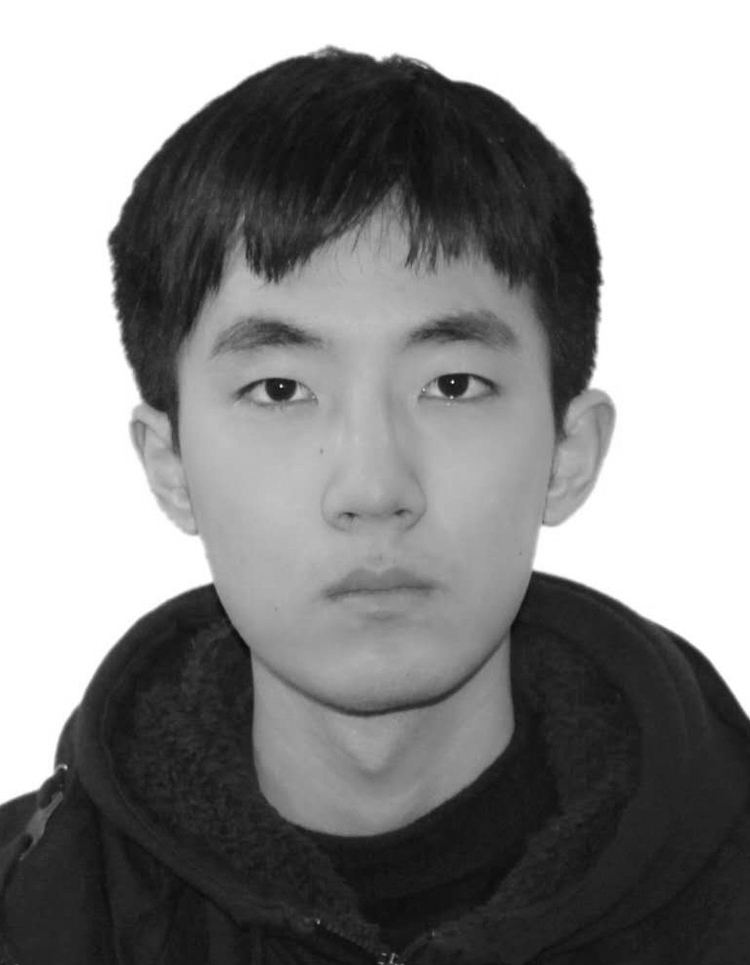}}]{Yuxuan Zhou}
	received the M.S. degree in geodesy and survey engineering from Wuhan University, Wuhan, China, in 2022.
	
	He is currently a doctor candidate at the School of Geodesy and Geomatics (SGG), Wuhan University, China. His current research areas include integrated navigation systems, simultaneous localization and	mapping, and multi-sensor fusion.
	
	Contact him via	e-mail: \emph{yuxuanzhou@whu.edu.cn}
\end{IEEEbiography}
\vspacebio
\begin{IEEEbiography}[{\includegraphics[width=1in,height=1.25in,clip,keepaspectratio,cframe={black!8!white}]{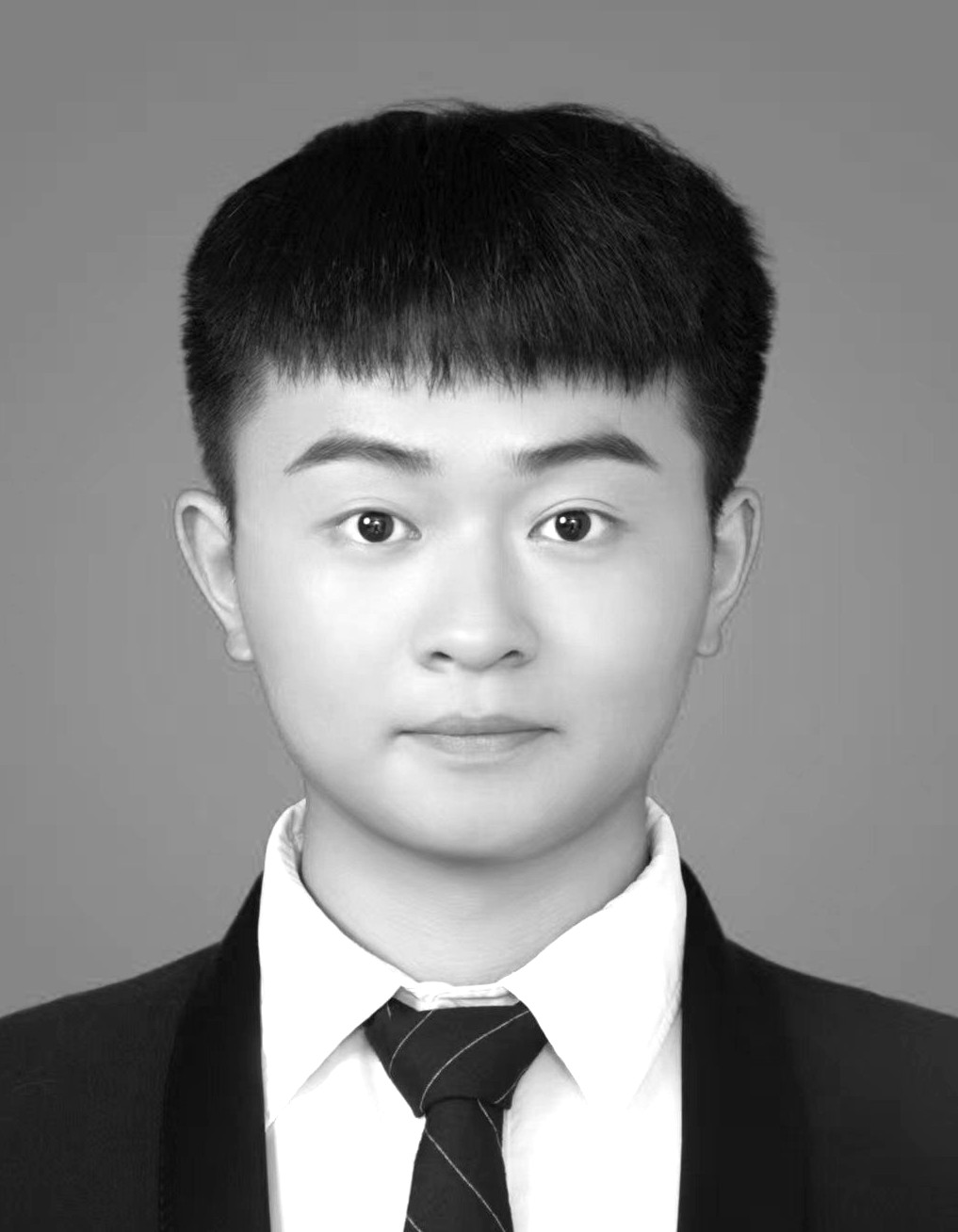}}]{Xiaoteng Yang}
    received the B.S. degree in geodesy and geomatics engineering from Wuhan University, Wuhan China, in 2023.
    
    He is currently a master candidate at the School of Geodesy and Geomatics (SGG), Wuhan University, also a member of the Mobile Autonomous Sensing And Surveying LAB. His current research mainly involves UAV-based multi-sensor fusion and integrated navigation system.
    
    Contact him via    e-mail: \emph{xtyang@whu.edu.cn}
\end{IEEEbiography}
\end{document}